\documentclass[11pt, a4paper, onecolumn, copyright, goog]{google}
\usepackage{subcaption}
\usepackage{comment}
\usepackage{multirow}
\usepackage{hyperref}
\hypersetup{
    colorlinks=true,
    linkcolor=blue,
    urlcolor=blue,
}

\usepackage[utf8]{inputenc}
\usepackage{xcolor}
\uselogo{}

\usepackage{dm-colors}
\usepackage{amsmath}
\usepackage{pstricks, pst-node}
\usepackage{verbatim}
\usepackage{scalerel}
\usepackage{longtable}
\usepackage{array}  
\usepackage{booktabs}
\usepackage{enumitem}
\usepackage{xspace}
\usepackage{bm}
\usepackage{bbm}
\usepackage{mathtools}
\usepackage{soul}
\usepackage{epsfig}
\usepackage{graphicx}
\usepackage{tcolorbox}
\tcbuselibrary{breakable}
\usepackage{amssymb}
\usepackage{colortbl}
\usepackage{csquotes}
\usepackage{etex}
\usepackage{setspace}
\usepackage{svg}
\usepackage{tabularx,ragged2e}
\usepackage{placeins}
\usepackage{afterpage}
\usepackage[symbol]{footmisc}
\usepackage[bibstyle=nature,citestyle=numeric-comp,%
            natbib=true,backend=biber,maxbibnames=5,minbibnames=1,%
            giveninits=false,sorting=none]{biblatex}
\usepackage{nameref}
\usepackage{varioref}
\usepackage[noabbrev,capitalize]{cleveref}
\usepackage{etoc}
\graphicspath{{figures/}}

\def\numpcps{9\xspace}

\title{Scalable Evaluation of Medical AI using Evidence-based Practice Recommendations}

\author[1]{Teya Bergamaschi}
\author[2]{Bijal Rajput}
\author[1]{Akshay Goel}
\author[1]{Matthew Thompson}
\author[1]{Jacqueline Shreibati}
\author[1]{Dave Steiner}
\author[1]{Ashley Carrick}
\author[1]{Mahvish Nagda}
\author[1]{Dale R. Webster}
\author[1]{Avinatan Hassidim}
\author[1]{Katherine Chou}
\author[1]{Yossi Matias}
\author[1]{Po-Hsuan Cameron Chen}
\author[1]{Christina Chen}
\author[$\dagger$,1]{Mike Schaekermann}
\author[$\dagger$,1]{Yun Liu}
\author[$\dagger$,1]{Anil Palepu}

\affil[$\dagger$]{Equal leadership}

\affil[1]{Google Research}
\affil[2]{NYU Grossman School of Medicine, NYU Langone Health}

\correspondingauthor{$\{$mikeshake$,\,$liuyun$,\,$apalepu$\}$@google.com}
\begin{document}

\begin{abstract}
Rigorously evaluating medical artificial intelligence (AI) requires testing open-ended clinical management across diverse presentations and stages of care. However, existing benchmarks rely on manual expert authoring, limiting coverage across conditions, conditional care pathways, and evolving guidelines. We develop an automated pipeline that extracts and verifies structured clinical criteria and branching decision trees from practice recommendations to generate scenarios spanning the care continuum, including nuanced presentations and  practice variations across regions and guideline editions. Applied to 436 \emph{BMJ Best Practice} clinical monographs (354 US, 82 International), this yields 3,200 scenarios spanning differential diagnosis, investigations, treatment, and complications, plus targeted suites for regional specificity (731 US/International pairs) and temporal freshness (630 scenarios). To grade responses, we develop a clinician-calibrated autorater that scores management plans on a 4-point rubric stratifying deviations by clinical severity. Across Gemini models, scaling improves adherence, while web search ($+0.8$ percentage points [pp] on average) and multi-sample synthesis ($+2.2\text{ pp}$) combine additively ($+2.8\text{ pp}$). In context, full monographs consistently trail condensed excerpts. On targeted suites, web search narrows international performance gaps ($+4.2\text{ pp}$) and improves adherence to newly updated guidance ($+4.6\text{ pp}$). In a study with Primary Care Physicians ($n = \numpcps$) across 200 scenarios, AI assistance raises adherence from $50.5\%$ to $70.0\%$, driven largely by fewer clinical omissions. Overall, our results demonstrate a scalable evaluation setup using
an evidence-based corpus originally written to guide human clinical decision-making, and demonstrate its viability to assess medical AI and its assistance on clinician decision-making.

\end{abstract}

\maketitle

\begin{refsegment}
\section{Introduction}
\label{sec:introduction}

Clinical care spans a sequence of decisions, from initial sufficiency of history and data gathering to differential diagnosis, workup, treatment selection, dosing, monitoring, and management of complications. Rigorously evaluating artificial intelligence (AI) systems for clinical use requires testing open-ended management across these stages and diverse patient presentations, rather than isolated clinical questions. While clinical practice guidelines and point-of-care references such as \emph{BMJ Best Practice}~\cite{bmj_best_practice} do not capture every nuance of bedside care, they systematically outline how conditions present and how management branches across patient subgroups. Because such resources also maintain regional editions and continuous evidence updates, they provide a living reference standard for evaluating clinical AI across care stages, health systems, and evolving medical evidence.

Translating this clinical complexity into scalable evaluations, however, remains a challenge. Large-scale multiple-choice and task-specific benchmarks~\cite{jin2021disease,khandekar2024medcalc,pal2023med} achieve broad coverage by reducing clinical reasoning to constrained option selection or narrow sub-skills. By contrast, open-ended rubric evaluations~\cite{arora2025healthbench,hicks2026healthbench}, specialist safety and management studies~\cite{wu2025first,palepu2025exploring,o2026large}, and simulated consultations~\cite{tu2025towards,lievin2026towards} capture richer management decisions, but rely on manual expert authoring for each case or rubric (\cref{sec:related_work}). This manual bottleneck limits coverage across conditional branches of care and makes it difficult to update benchmarks when practice standards differ by region or change over time.

\begin{figure}[t!]
  \centering
  \includegraphics[width=\textwidth]{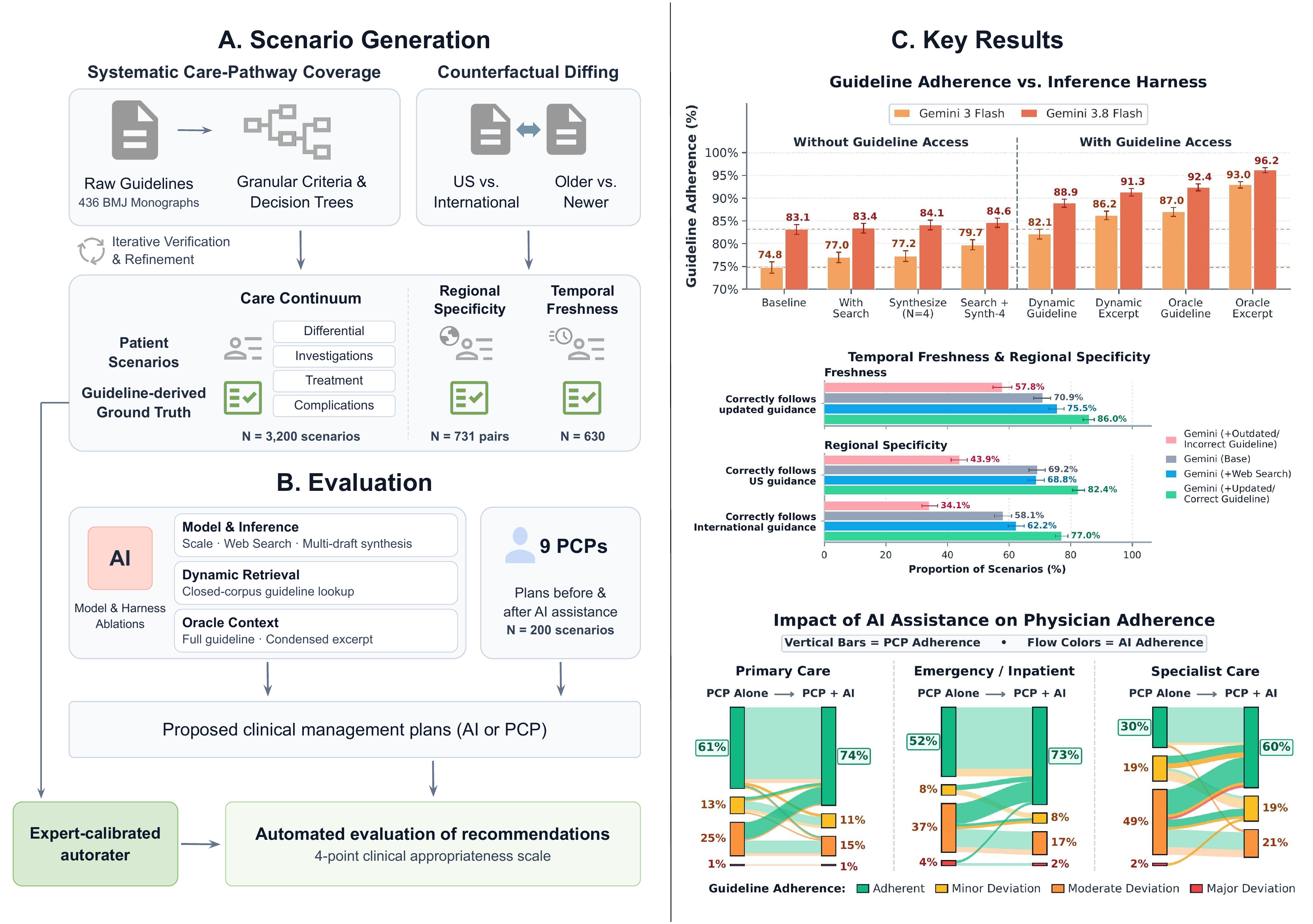}
  \vspace{0.1cm}
  \caption{\textbf{Overview of the Scenario Generation Methodology, Evaluation Framework, and Key Findings.} 
  \textbf{a}. Automated pipeline parsing practice guidance into structured clinical criteria and branching treatment decision trees, with iterative verification to generate scenarios across four stages of care (Differential Diagnosis, Diagnostic Investigations, Treatment Selection, and Complication Management), alongside targeted scenarios isolating regional and temporal guideline differences from \emph{BMJ Best Practice} clinical monographs. 
  \textbf{b}. Multi-dimensional evaluation comparing Gemini models across model sizes, inference strategies, search configurations, and context settings, as well as primary care physicians with and without AI assistance. 
  \textbf{c}. Summary of primary findings from top to bottom: comparison of Gemini models and inference methods, the effect of web search (Grounding with Google Search) and guideline context on temporal freshness and regional specificity, and the assistive impact of AI recommendations on clinician management quality.}
  \label{fig:contributions_overview}
\end{figure}

To address this bottleneck, we develop an automated pipeline with built-in verification that converts clinical practice recommendations into open-ended patient scenarios and reference ground truth criteria (\cref{fig:contributions_overview}a). The pipeline parses each guideline into structured diagnostic, workup, and complication criteria alongside branching treatment decision trees that capture how management depends on severity, comorbidities, contraindications, and prior treatment response. From each extracted item or decision-tree path, it generates patient vignettes and validates them through rule-based checks and iterative faithfulness review against the source guidance. Because this traverses each monograph's criteria and conditional branches systematically, it enables targeting of specific, nuanced presentations within every condition (such as particular comorbidities, contraindications, severity tiers, or atypical presentations). Applying a similar automated workflow to paired regional editions and longitudinal revisions isolates concrete clinical divergences (such as shifting diagnostic thresholds, preferred therapies, or escalation rules) and automatically synthesizes scenarios built to stress-test those exact differences. Applied across 436 \emph{BMJ Best Practice} clinical monographs (354 US, 82 International), this approach yields 4,561 scenarios: a 3,200-scenario core benchmark spanning differential diagnosis, investigations, treatment, and complications, plus targeted suites for regional specificity (731 scenarios with paired US/International criteria) and temporal freshness (630 scenarios across 140 recently updated topics).

To grade responses at scale, we calibrate a large language model (LLM) \emph{autorater} against physician ratings on a 4-point ordinal clinical appropriateness scale (\cref{tab:likert_rubric}). Using this framework, we evaluate Gemini across model sizes, multi-sample synthesis, web search, and guideline context formats, alongside a reader study measuring AI assistance for practicing clinicians (\cref{fig:contributions_overview}b,c). We find that on the core benchmark, model scaling and multi-sample synthesis substantially improve guideline adherence and combine additively with web search. When guidelines are provided in context, condensed structured excerpts consistently outperform full monographs; under dynamic retrieval from the guideline corpus, locating the right guidance remains harder when presenting symptoms are undifferentiated. On the targeted suites, web search improves adherence on recently updated and International guidance, though large gaps remain relative to providing the target guideline itself. Finally, in a study with Primary Care Physicians ($n = \numpcps$) across 200 primary care, emergency/inpatient, and specialist scenarios, reviewing AI recommendations raises adherence and reduces clinical omissions across all three settings, especially on specialist cases (\cref{fig:contributions_overview}c). Together, these results demonstrate how evidence-based clinical references can support scalable, targeted, and continuously updatable evaluation of medical AI across the care continuum and evolving regional practice standards.

\section{Related Work}
\label{sec:related_work}

\subsection{Benchmarks and Evaluation of Medical AI}
Early evaluations of medical LLMs focused on multiple-choice licensing exams such as MedQA~\cite{jin2021disease,singhal2023large,nori2023can,Lievin2024-qd}, alongside targeted sub-task benchmarks for clinical calculations (MedCalc-Bench~\cite{khandekar2024medcalc}), medical hallucination (Med-HALT~\cite{pal2023med}), and medication reasoning (RxQA~\cite{lievin2026towards}). While many of these are easy to scale, these benchmarks test constrained option selection or isolated skills rather than integrated patient management.

As frontier models have approached or surpassed $90\%$ accuracy on multiple-choice exams~\cite{nori2023can,singhal2025toward,saab2024capabilities}, recent work has shifted toward open-ended management and conversational evaluations. These benchmarks span multi-task clinical suites (MedHELM~\cite{bedi2025medhelm}), open-ended responses graded with physician-authored rubrics (HealthBench and HealthBench Professional~\cite{arora2025healthbench,hicks2026healthbench}), electronic consultation safety audits (NOHARM~\cite{wu2025first}), and expert-graded diagnostic and subspecialty management studies~\cite{mcduff2025towards,doi:10.1126/science.adz4433,palepu2025exploring,o2026large}. Complementary work evaluates interactive simulated consultations collectively spanning diagnostic history-taking, multimodal data collection, and multi-visit care (CRAFT-MD~\cite{johri2025evaluation}, AgentClinic~\cite{Schmidgall2024-zf}, and AMIE~\cite{tu2025towards,saab2026advancing,lievin2026towards}), as well as sequential diagnosis (SDBench~\cite{nori2026sequential}) and virtual electronic health record agent workflows (MedAgentBench~\cite{jiang2025medagentbench}). Because these evaluations rely on static case banks or manual expert authoring, however, they remain costly to expand across conditions and conditional branches of care or to update as clinical standards change. Recent work has also begun translating narrative guidelines into executable decision support (CPGPrompt~\cite{deng2026cpgprompt}) and conversational AI evaluations (CPGBench~\cite{tan2026cpgbench}), though extensive and systematic evaluation across branching care pathways and paired temporal revisions and regional editions remains limited.

Because clinical decision support systems increasingly use retrieval-augmented generation (RAG)~\cite{lewis2020retrieval} to ground responses in medical reference texts~\cite{zakka2024almanac,xiong2024benchmarking}, benchmark design must also account for how models consume retrieved evidence. Broader studies show that LLMs often underuse details buried in long inputs~\cite{liu2024lost} and are distracted by topically related but irrelevant text~\cite{shi2023large}, motivating systematic evaluation of how model scale, retrieval, and context format interact across clinical tasks.

\subsection{Clinical Practice Guidelines, Temporal Freshness, and Regional Specificity}
Clinical practice guidelines (CPGs) are developed by specialty societies and health authorities through systematic evidence synthesis, grading, and structured recommendation development~\cite{woolf2012developing}. Because primary CPGs are lengthy narrative documents that can become outdated between multi-year revision cycles~\cite{shekelle2001validity}, point-of-care resources such as \emph{BMJ Best Practice} and UpToDate continuously synthesize emerging evidence into structured, condition-specific recommendations. Although clinicians often have sound reasons to depart from standardized pathways due to multimorbidity, patient preferences, or local constraints~\cite{tinetti2004potential,boyd2005clinical,cabana1999don}, these syntheses establish a shared baseline for evidence-based care.

In clinical practice, guideline volume, time constraints, and frequent updates create substantial awareness and familiarity barriers for physicians~\cite{cabana1999don}. Evolving and region-specific guidance poses distinct challenges for LLMs. Over time, models without external retrieval rely on static training data that may encode superseded standards~\cite{vu2024freshllms}. Across health systems, diagnostic thresholds, approved formularies, and care pathways often vary by country, yet models trained predominantly on US medical text may default to US conventions. Moreover, even when updated or local guidance is supplied in context, conflicts between retrieved text and parametric priors can degrade adherence in either direction---persisting with outdated training knowledge or over-deferring to flawed context~\cite{longpre2021entity,xie2024adaptive}. Comparing longitudinal revisions and regional editions of practice guidance offers a direct way to evaluate both challenges.

\subsection{Patient Safety and Clinical Error Severity Taxonomies}
Prior clinical quality and safety research has developed complementary frameworks to assess clinical appropriateness (such as the RAND/UCLA Appropriateness Method~\cite{fitch2001rand}, which evaluates whether expected benefits outweigh harms across clinical options) and to grade error severity. Patient-safety harm scales---such as the World Health Organization (WHO) International Classification for Patient Safety degree-of-harm scale~\cite{runciman2009towards} and the Agency for Healthcare Research and Quality (AHRQ) Common Formats Harm Scale~\cite{williams2015reliability}---stratify observed outcomes from no harm through mild, moderate, or severe harm to death, while methods such as the Dean \& Barber medication error severity scale~\cite{dean1999validated} score potential harm when actual patient outcomes are unknown. Following these approaches, we score plans on a 4-point ordinal clinical appropriateness rubric (\cref{tab:likert_rubric}) adapted for clinician-calibrated LLM-based evaluation~\cite{zheng2023judging}.

\subsection{Clinician--AI Collaboration in Diagnostic and Management Reasoning}
Although many studies compare standalone LLMs directly against clinicians on diagnostic and management tasks~\cite{doi:10.1126/science.adz4433,palepu2025exploring}, strong standalone accuracy does not guarantee effective clinician--AI collaboration: in diagnostic reasoning, LLM assistance improved physicians' differential diagnoses on complex clinical cases~\cite{mcduff2025towards}, whereas in other cases, access to a general-purpose LLM yielded no significant gain over conventional resources alone~\cite{goh2024large}. In clinical management, randomized controlled studies have shown that LLM assistance improves physician management plans in general medicine~\cite{goh2025gpt}, subspecialty cardiology~\cite{o2026large}, and multi-specialty e-consultations~\cite{wu2025first}. Medical AI has also been evaluated in a randomized study of asynchronous physician oversight of conversational AI~\cite{vedadi2025towards}, retrospective comparisons of AI and physician recommendations in virtual urgent care~\cite{zeltzer2025comparison}, and real-world primary care deployments: a quality-improvement study reported fewer diagnostic and treatment errors~\cite{korom2025ai}, while a subsequent pragmatic cluster-randomized trial of essentially the same system similarly improved documentation and treatment-plan quality yet found no statistically significant reduction in 14-day treatment failure~\cite{agweyu2026generative}.

At the same time, clinician--AI collaboration carries important trade-offs. Assistive gains can be tempered by automation bias and broader susceptibility to flawed recommendations~\cite{goddard2012automation,gaube2021ai,jabbour2023measuring}. Conversely, even when AI recommendations are comprehensive, LLMs tend to favor more exhaustive workups, testing, and referrals than treating physicians, substantially increasing projected care costs~\cite{aran2026comparative}, and in a prospective ambulatory primary care study, physicians outperformed conversational AI on the practicality and cost-effectiveness of management plans~\cite{brodeur2026conversational}. Evaluating clinician--AI collaboration therefore requires carefully measuring not only helpful revisions and detrimental plan changes, but also how clinicians selectively filter AI recommendations across care settings.

\section{Methods}
\label{sec:methods}

We first describe our automated pipeline for converting practice recommendations into structured clinical criteria, branching treatment decision trees, and verified patient scenarios across stages of care, regions, and guideline updates (\cref{sec:scenarios}; \cref{fig:methodology_and_materialization}). Using the scenarios generated from \emph{BMJ Best Practice} alongside a clinician-calibrated 4-point autorater, we then evaluate standalone models, retrieval and context strategies, and physician--AI collaboration (\cref{sec:autoeval}). Full schemas, validation rules, and prompt templates are provided in \cref{appendix:corpus_and_schemas,appendix:extraction_pipeline_details}.

\subsection{Scenario Extraction and Generation}
\label{sec:scenarios}

\emph{BMJ Best Practice} is a clinical decision support tool that supports healthcare professionals in making evidence-based decisions on diagnosis, treatment, and prevention. Where evidence is scarce or equivocal, expert opinion is provided in \emph{BMJ Best Practice}. We use a corpus of 436 structured \emph{BMJ Best Practice} clinical monographs: 354 \emph{US} editions and 82 paired \emph{International} editions covering topics where regional recommendations differ. In the US editions, the evidence approach focuses on US clinical guidelines and practice, using US drug nomenclature (USANs) and appropriate units for test results and clinical parameters. In the International editions, the general evidence approach includes US clinical guidelines in addition to those from major UK, European, Oceania, and other global guideline producers, using recommended International Non-proprietary Names (rINNs) for drug nomenclature and appropriate units for test results and clinical parameters; a subset of topics is anchored primarily in UK and European clinical guideline sources. Core and physician evaluations use a July 2026 corpus snapshot, while targeted regional and temporal evaluations use a September 2026 snapshot (comparing updated US editions against a December 2025 snapshot for temporal freshness; \cref{methods:targeted_diff_scenarios}).

\subsubsection{Structured Clinical Item and Decision-Tree Extraction}
\label{sec:methods:structured_items}
We parse each clinical monograph with Gemini 3 Flash into structured records across four clinical categories (\cref{fig:methodology_and_materialization}a; detailed schemas and prompts in \cref{appendix:corpus_and_schemas}):
\begin{itemize}
    \item \textbf{Differential Diagnosis, Investigations, and Complications:} From narrative text and diagnostic tables, we extract (i) candidate differential diagnoses with distinguishing features and diagnostic tests; (ii) diagnostic investigations with expected findings and priority tiers (first-line versus adjunct/second-line); and (iii) disease- or treatment-related complications with onset timeframe, likelihood, and clinical description.
    \item \textbf{Hierarchical Treatment Decision Trees:} Because treatment depends on patient risk factors, severity, comorbidities, contraindications, and prior response, we parse each monograph's treatment section into a branching decision tree (\cref{fig:methodology_and_materialization}b). Branch nodes represent patient states or decision points, treatment entries record prioritized interventions (first-line, second-line, or adjunct) with alternative options, dosing details, and clinical warnings or contraindications, and child nodes capture follow-up steps based on treatment response. Following individual paths through the tree yields distinct scenarios across presentations and stages of care.
\end{itemize}

Before scenario synthesis, each extracted clinical item and decision tree undergoes automated verification and targeted refinement against the source monograph XML (\cref{fig:methodology_and_materialization}a; details and prompts in \cref{appendix:extraction_verification}). Deterministic rules check structural integrity (tree acyclicity, non-empty required fields, populated leaf-node treatment plans, and duplicate entries), while category-specific LLM critique and an independent direct XML audit check clinical accuracy, safety, omissions, and strict groundedness (flagging any external clinical logic absent from the source XML). Flagged records undergo refinement using the structured issue feedback and source XML to produce verified records.

\begin{figure}[ht!]
  \centering
  \includegraphics[width=\textwidth]{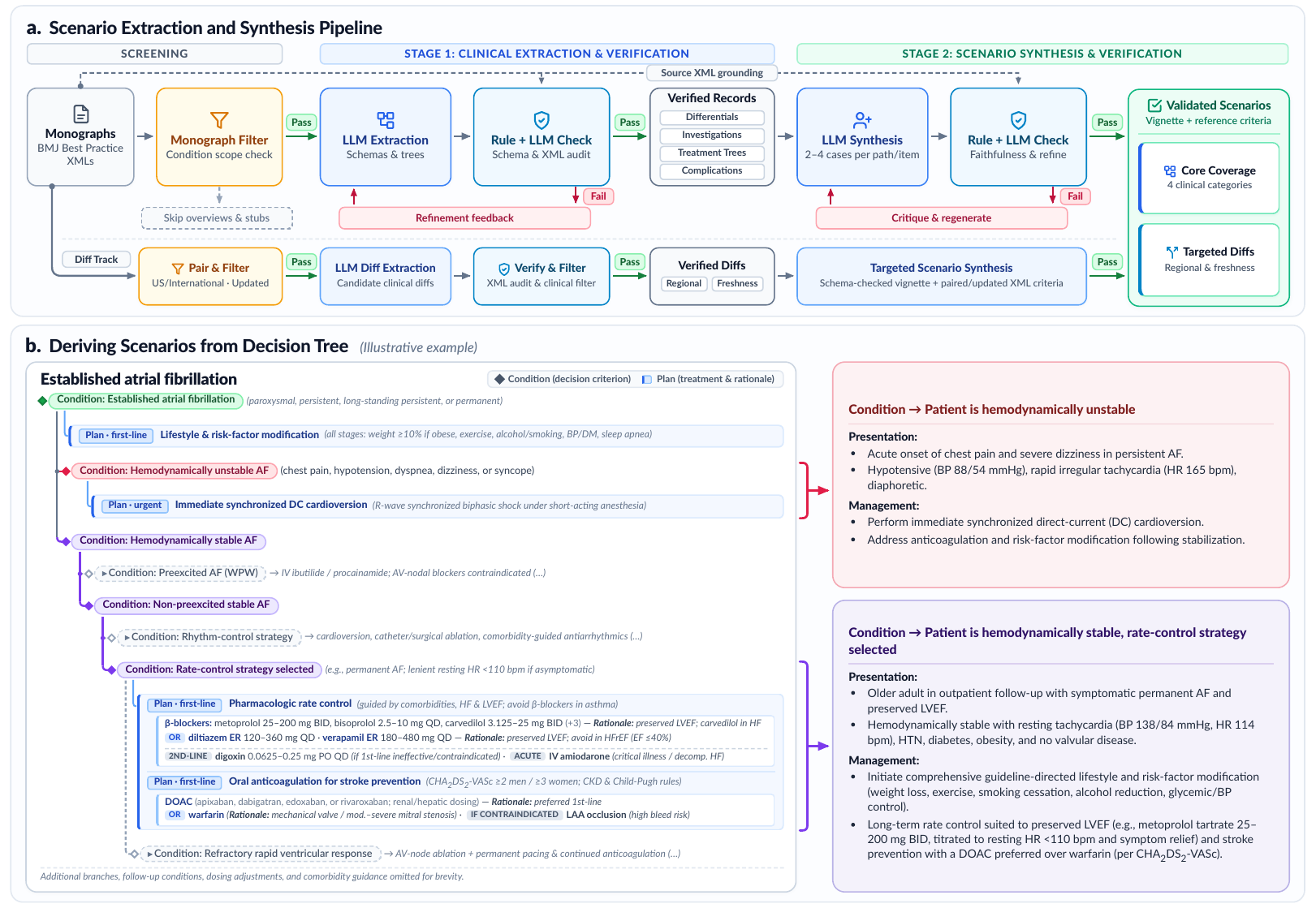}
  \vspace{0.1cm}
  \caption{\textbf{Automated guideline extraction pipeline and decision-tree scenario generation.} 
  \textbf{a}. Scenario extraction and synthesis pipeline: raw guidelines undergo LLM-based structured extraction, and rule- plus LLM-based faithfulness checks grounded against the source XML to produce verified clinical records, followed by vignette synthesis and iterative verification. Separately, pairs of clinical monographs (paired regional and updated editions) are processed with automated XML verification and clinical filtering to contribute additional targeted scenarios alongside the core set of scenarios. 
  \textbf{b}. Summarized illustrative example of deriving clinical scenarios from a branching treatment decision tree for \emph{Established atrial fibrillation} (abridged for visual clarity; full schemas and prompts in \cref{appendix:corpus_and_schemas}), showing how different conditional branches (e.g., an unstable presentation requiring synchronized direct-current cardioversion versus a stable outpatient presentation managed with rate control, oral anticoagulation, and risk-factor modification) map to distinct patient presentations and management criteria. Note: this example decision tree was condensed to illustrate the branching and was extracted from an earlier timepoint's clinical monograph. The actual contents may not necessarily reflect accurate, up-to-date medical information.}
  \label{fig:methodology_and_materialization}
\end{figure}

\subsubsection{Iterative Scenario Synthesis, Validation, and Sampling}
\label{sec:methods:scenario_generation}
From each extracted clinical item or treatment-tree path, we use Gemini 3 Flash to generate 2--4 candidate scenarios including standard, atypical, borderline, and high-acuity presentations, pairing each patient vignette with an open-ended clinical prompt and reference grading criteria grounded in the source clinical monograph. Automated rule-based checks confirm structural completeness and strip explicit guideline mentions, after which an LLM verifier evaluates faithfulness to the source guidance and regenerates candidates (up to 5 retries) when discrepancies arise. Passing candidates undergo a final refinement pass to ensure realistic vital signs and laboratory values and clear grading criteria (\cref{appendix:extraction_pipeline_details}).

From the resulting pool of verified US scenarios, we sample a balanced 3,200-scenario core benchmark (800 per clinical category) and a balanced 200-scenario evaluation set (50 per category) for physician evaluation and autorater calibration (\cref{appendix:extraction_pipeline_details}).

\subsubsection{Regional and Temporal Targeted Scenarios}
\label{methods:targeted_diff_scenarios}
To isolate cases where guidance differs by region or changes over time, we use Gemini 3.8 Flash to compare paired monograph editions, extracting and verifying substantive clinical differences across diagnostic thresholds, first-line drugs, dosing limits, and surgical indications while ignoring stylistic differences (\cref{fig:methodology_and_materialization}a; \cref{appendix:targeted_diff_prompts,appendix:illustrative_regional_freshness_examples}). For Regional Specificity, we compare the paired US and International editions and convert each verified divergence into a patient vignette with paired US and International reference criteria. For Temporal Freshness, we compare earlier and updated US editions revised between corpus snapshots to synthesize scenarios anchored to the updated guidance.

\subsection{Evaluation}
\label{sec:autoeval}

\subsubsection{Clinical Appropriateness Scale and Autorater}
Building on rubric- and severity-based clinical LLM evaluation frameworks~\cite{arora2025healthbench,wu2025first}, we grade each model or clinician management plan against the scenario's structured reference criteria using an LLM autorater (Gemini 3.8 Flash; prompt in \cref{fig:lst_autorater_prompt}) on a 4-point ordinal appropriateness scale (\cref{tab:likert_rubric}): \emph{Adherent / Acceptable} (score 4), \emph{Minor Deviation} (score 3), \emph{Moderate Deviation} (score 2), and \emph{Major Deviation} (score 1). Before assigning a score, the autorater reconciles key actions in the candidate plan against the structured reference ground truth, writes a brief clinical rationale summarizing substantive omissions or discrepancies, and evaluates severity sequentially from \emph{Major Deviation} to \emph{Adherent / Acceptable}. We calibrated the autorater prompt against physician ratings of standalone AI management plans on the 200-scenario evaluation set and tested its repeatability and sensitivity to prompt ordering, list structure, and response length (\cref{appendix:autorater_calibration}). On the core benchmark and physician study, we report both the full four-tier distribution and \emph{guideline adherence}, defined as the proportion of responses rated as 4 (``adherent / acceptable''); for the targeted temporal and regional suites, the same autorater prompt is appended with the verified divergence and targeted checklist to assess adherence specifically to the updated or regionally matched recommendation.
\begin{table}[h]
\centering
\caption{\textbf{Clinical appropriateness rubric.} Ordinal 4-point scale used by expert clinicians and the calibrated LLM autorater to evaluate proposed management plans against ground-truth practice recommendations.}
\label{tab:likert_rubric}
\footnotesize
\renewcommand{\arraystretch}{1.12}
\begin{tabularx}{\textwidth}{cl>{\RaggedRight\arraybackslash}X}
\toprule
\textbf{Score (1--4)} & \textbf{Category} & \textbf{Description} \\
\midrule
4 & Adherent / Acceptable & Matches the evidence-based practice recommendation, or an alternative but clinically appropriate choice (e.g., equivalent drug from the same class). \\
3 & Minor Deviation & Suboptimal but safe (e.g., omitting routine monitoring; ordering a test too early). \\
2 & Moderate Deviation & Clinically significant omission or suboptimal choice without immediate risk of severe harm. \\
1 & Major Deviation & Contraindicated treatment, unsafe dose, or missed life-threatening red flag. \\
\bottomrule
\end{tabularx}
\end{table}

\subsubsection{Experimental Evaluations}
\label{sec:methods:experimental_evaluations}
We conduct three primary sets of experiments:

\paragraph{Model, Inference Harness, and Context Representation Ablations.}
On the 3,200-scenario core benchmark, we evaluate seven Gemini models across two size tiers (Gemini Flash Lite versions 3.1 and 3.5; Gemini Flash versions 3, 3.5, 3.6, 3.7, and 3.8) under eight inference and context configurations (\cref{fig:system_scaling_context}; harness details and full rating distributions in \cref{appendix:inference_harnesses,appendix:extended_results}):
\begin{itemize}
    \item \emph{Baseline:} Generating a management plan directly from the patient vignette without external tools or guideline text.
    \item \emph{With Search:} Generating search queries from the vignette, retrieving top Google Search results in a single batch, and conditioning the response on those results.
    \item \emph{Synthesize-4} and \emph{Search + Synth-4:} Following inference-time sampling and aggregation approaches~\cite{wang2022self,snell2024scaling}, sampling $N=4$ independent candidate plans and synthesizing them into a single consensus recommendation, conditioned on either the vignette alone (\emph{Synthesize-4}) or the vignette plus a fixed set of retrieved search results (\emph{Search + Synth-4}).
    \item \emph{Dynamic Guideline} and \emph{Dynamic Excerpt:} Using the evaluated model to select up to $K$ (default $5$) relevant titles from the 354-title US \emph{BMJ Best Practice} catalog based on the vignette, then conditioning on either the full cleaned monographs (\emph{Dynamic Guideline}, retrieval analyses in \cref{appendix:dynamic_retrieval}) or the model-selected structured JSON sections extracted in \cref{sec:methods:structured_items} (\emph{Dynamic Excerpt}).
    \item \emph{Oracle Guideline} and \emph{Oracle Excerpt:} Providing the exact (``ground truth'') source clinical monograph directly in context as an upper-bound reference, formatted as either the full cleaned monograph (\emph{Oracle Guideline}) or its single category-matched structured JSON section (\emph{Oracle Excerpt}).
\end{itemize}

\paragraph{Targeted Evaluations of Temporal Freshness and Regional Specificity.}
We evaluate the same seven models on the Temporal Freshness and Regional Specificity suites across four configurations (\cref{fig:targeted_diffs}):
\begin{itemize}
    \item \emph{Base:} Prompting with the patient vignette (and target region for regional scenarios) without search or guideline text.
    \item \emph{+Web Search:} Using the same single-pass query generation and Google Search retrieval harness as \emph{With Search} above to retrieve newly published information or regional guidance.
    \item \emph{+Updated/Correct Guideline:} Providing the revised monograph (Temporal Freshness) or regionally matched edition (Regional Specificity) in context as an upper-bound reference.
    \item \emph{+Outdated/Incorrect Guideline:} Providing the superseded historical monograph or conflicting regional edition (e.g., the US edition for an International scenario) in context to test susceptibility to outdated or mismatched guidance.
\end{itemize}

\paragraph{Primary Care Physician Baseline and AI Assistance Study.}
To measure the impact of AI assistance, we recruited $n = \numpcps$ board-certified Primary Care Physicians (PCPs) with a median of 11 years of post-residency clinical experience (IQR: 11--12; range: 5--26; mean $12.3 \pm 5.6$ years) to evaluate the 200-scenario evaluation set (50 per clinical category; 1 physician per scenario; further details in \cref{appendix:pcp_study_protocol}). In a two-stage sequential design, each physician first wrote a free-text management plan covering differential diagnosis, diagnostic investigations, and treatment without AI support (\emph{PCP Alone}, with standard non-AI clinical references permitted). Immediately after submitting, the physician viewed an AI recommendation (Gemini 3 Flash with \emph{Search + Synth-4}) and could revise or keep their initial plan (\emph{PCP + AI}). Where physicians entered shorthand post-AI revision notes (e.g., ``no change'' or brief incremental additions), an automated normalization step converted them into standalone clinical plans while preserving original wording and removing references to AI assistance. Both pre- and post-AI plans were scored by the calibrated autorater blinded to the plan source (\cref{tab:likert_rubric}), and paired pre- and post-AI responses were compared using Gemini 3.8 Flash to categorize unchanged responses, specific clinical additions, and detrimental edits (\cref{fig:clinician_baseline}). For analysis by scope of practice, the 200 scenarios were stratified post-hoc into \emph{Primary Care (Clinic)} ($N=95$), \emph{Emergency / Inpatient Generalist} ($N=52$), and \emph{Specialist Care} ($N=53$) using Gemini 3.8 Flash (\cref{fig:lst_pcp_stratification_prompt}) with clinician spot-checks confirming placement.

\subsubsection{Statistical Analysis}
\label{sec:methods:statistical_analysis}
We evaluated standalone model configurations across five independent model runs per scenario (each graded once by the calibrated autorater), averaging runs to a scenario-level adherence rate and computing $95\%$ confidence intervals ($2.5\text{th}$ and $97.5\text{th}$ percentiles) via 1,000 non-parametric bootstrap resamples clustered by scenario (\cref{appendix:extended_results}). For the unassisted, AI-alone, and AI-assisted management plans in the paired physician study (\cref{fig:clinician_baseline}), each management plan was graded across five independent autorater passes and assigned its median ordinal rating. Pre- versus post-AI clinician changes were summarized via tier-transition flows (\cref{fig:clinician_baseline}a), upgrade, unchanged, and downgrade proportions, and $95\%$ confidence intervals on plan modification rates across 1,000 care-setting-stratified scenario bootstrap resamples (\cref{appendix:pcp_study_protocol}).

\section{Results}
\label{sec:results}

\subsection{Scenario Generation and Curation}
\label{results:scenario_curation}

Across the US corpus, our extraction and validation pipeline yielded 32,621 verified clinical scenarios, from which we sampled the balanced 3,200-scenario core benchmark and the 200-scenario physician and calibration set (\cref{appendix:extraction_pipeline_details}). Together with 630 Temporal Freshness scenarios (from the 140 of 283 revised US topics that contained substantive clinical updates) and 731 paired Regional Specificity scenarios (each with localized US and International variants across the 82 topic pairs), the three evaluation suites comprise 4,561 scenarios (\cref{results:targeted_diffs}).

\subsection{Clinician-Calibrated Autorater Validation}
\label{results:autorater_validation}

Against physician ratings of standalone AI management plans on the 200-scenario evaluation set (\cref{appendix:autorater_calibration}), the calibrated 4-point autorater achieved $89.0\%$ ($178/200$) exact four-tier agreement and $96.0\%$ ($192/200$) adjacent-tier agreement ($|\Delta\text{tier}| \le 1$; five-run median autorater rating; \cref{tab:supp_autorater_confusion}). Perturbation checks showed high test--retest repeatability ($98.5\%$ exact agreement; $100.0\%$ adjacent-tier agreement) and low sensitivity to prompt section order, list and field ordering, and response length ($95.5\%\text{--}97.0\%$ exact agreement; $99.5\%\text{--}100.0\%$ adjacent-tier agreement; \cref{appendix:autorater_consistency}, \cref{tab:supp_autorater_unit_tests}).

\subsection{Performance Across Inference Harnesses, Scenario Categories, and Models}
\label{results:models_and_inference}
\begin{figure}[ht!]
  \centering
  \includegraphics[width=\textwidth]{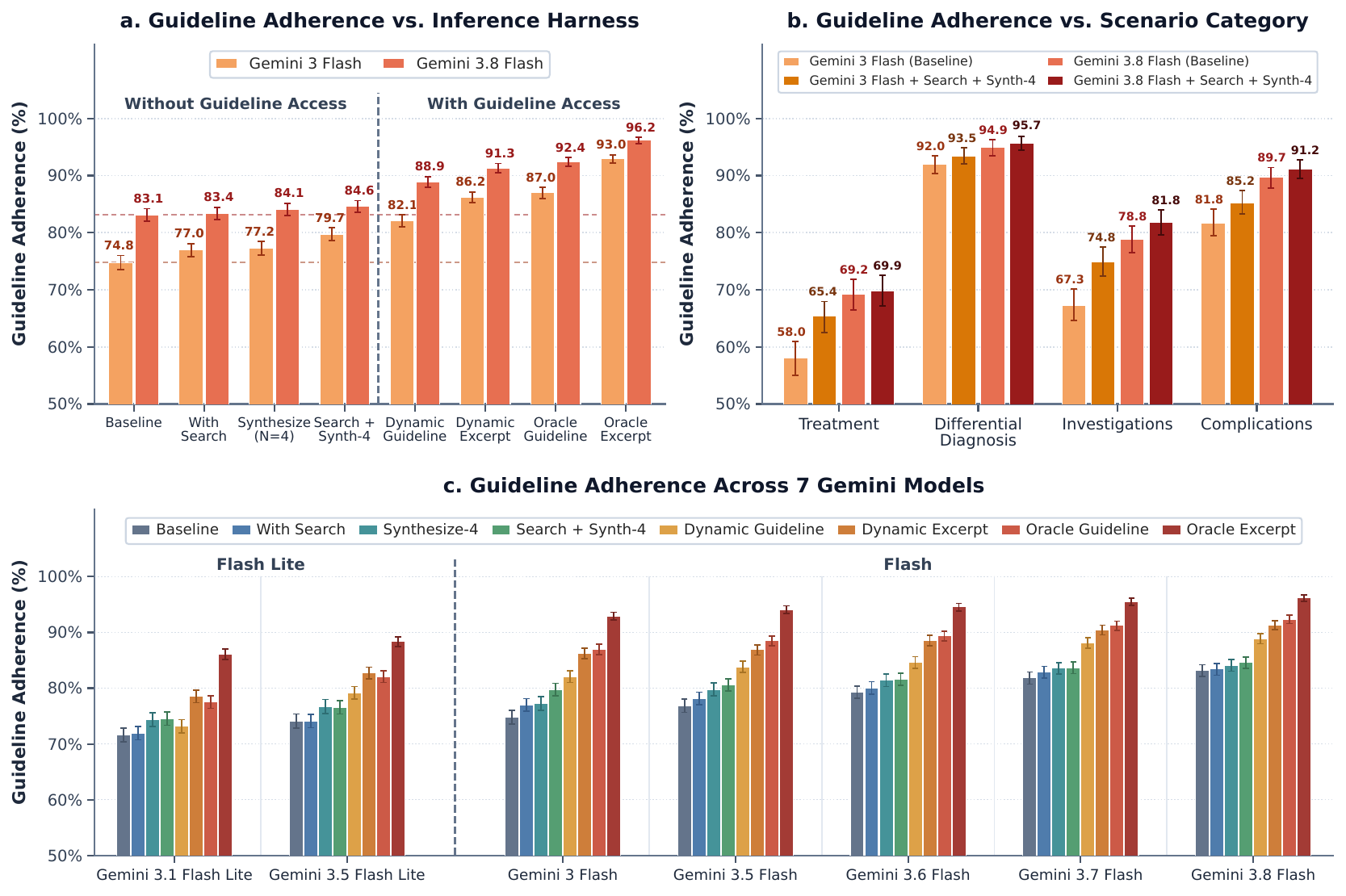}
  \vspace{0.1cm}
  \caption{\textbf{Guideline adherence across models, inference harnesses, and context representations.} \textbf{a}. Adherence across all eight inference and context settings for Gemini 3 Flash and Gemini 3.8 Flash. \textbf{b}. Adherence by clinical scenario category (Treatment, Differential Diagnosis, Investigations, and Complications) comparing \emph{Baseline} and \emph{Search + Synth-4} on Gemini 3 Flash and Gemini 3.8 Flash. \textbf{c}. Adherence across all seven \emph{Flash Lite} and \emph{Flash} models and eight configurations ($N = 3{,}200$ scenarios; error bars indicate $95\%$ CIs).}
  \label{fig:system_scaling_context}
\end{figure}

On the 3,200-scenario core benchmark (\cref{fig:system_scaling_context}a,c), web search (\emph{With Search}; $+0.8$ percentage points [pp] on average across models) and multi-sample synthesis (\emph{Synthesize-4}; $+2.2\text{ pp}$ on average) each improve guideline adherence over \emph{Baseline}, and combining them (\emph{Search + Synth-4}; $+2.8\text{ pp}$ on average) yields additive gains, reaching $84.6\%$ on Gemini 3.8 Flash. Conditioning on corpus guidelines (\cref{fig:system_scaling_context}a, right) raises adherence further, though dynamic retrieval trails the corresponding oracle configuration across every model due to imperfect recall and competing retrieved monographs, with retrieval recall lowest on undifferentiated \emph{Differential Diagnosis} presentations (\cref{appendix:dynamic_retrieval}). Under both dynamic and oracle access, condensed structured excerpts outperform full-length monographs across every model, reaching up to $96.2\%$ with \emph{Oracle Excerpt} on Gemini 3.8 Flash (residual error analysis in \cref{appendix:oracle_residual}).

By clinical category (\cref{fig:system_scaling_context}b), baseline adherence is substantially higher on \emph{Differential Diagnosis} and \emph{Complications}, whereas \emph{Investigations} and especially \emph{Treatment} planning remain harder. Combining search and synthesis (\emph{Search + Synth-4}) improves performance across all four categories.

Comparing all seven models (\cref{fig:system_scaling_context}c; \cref{appendix:extended_results}, \cref{tab:supp_full_model_matrix}), adherence increases steadily with model scale (\emph{Flash Lite} vs.\ \emph{Flash}) and generation (from $71.6\%$ baseline on Gemini 3.1 Flash Lite to $83.1\%$ on Gemini 3.8 Flash). Remaining errors are mostly \emph{Moderate} or \emph{Minor Deviations}, while \emph{Major Deviations} remain below $1\%$ across all configurations ($0.18\%\text{--}0.87\%$; \cref{tab:supp_full_model_matrix}). Whereas standard \emph{Flash} models benefit from every inference scaffold, smaller \emph{Flash Lite} models gain little from web search and show a wider average gap between full-monograph and condensed-excerpt conditioning (particularly under oracle access).

\subsection{Temporal Freshness and Regional Specificity}
\label{results:targeted_diffs}

\paragraph{Incorporating Newly Revised Guidelines (Temporal Freshness).}
On the Temporal Freshness suite (\cref{fig:targeted_diffs}a; examples in \cref{appendix:illustrative_regional_freshness_examples}), ungrounded adherence (\emph{Base}) rises with model scale and across \emph{Flash} generations ($70.9\%$ cross-model average), though models frequently default to superseded recommendations. Web search (\emph{+Web Search}) improves adherence across all seven models ($+4.6\text{ pp}$ on average), but in-context guidelines drive much larger shifts in both directions: supplying the superseded edition (\emph{+Outdated/Incorrect Guideline}) pulls adherence well below baseline on every model, whereas supplying the revised edition (\emph{+Updated/Correct Guideline}) raises adherence by $+15.0\text{ pp}$ on average.

\begin{figure}[ht!]
  \centering
  \includegraphics[width=\textwidth]{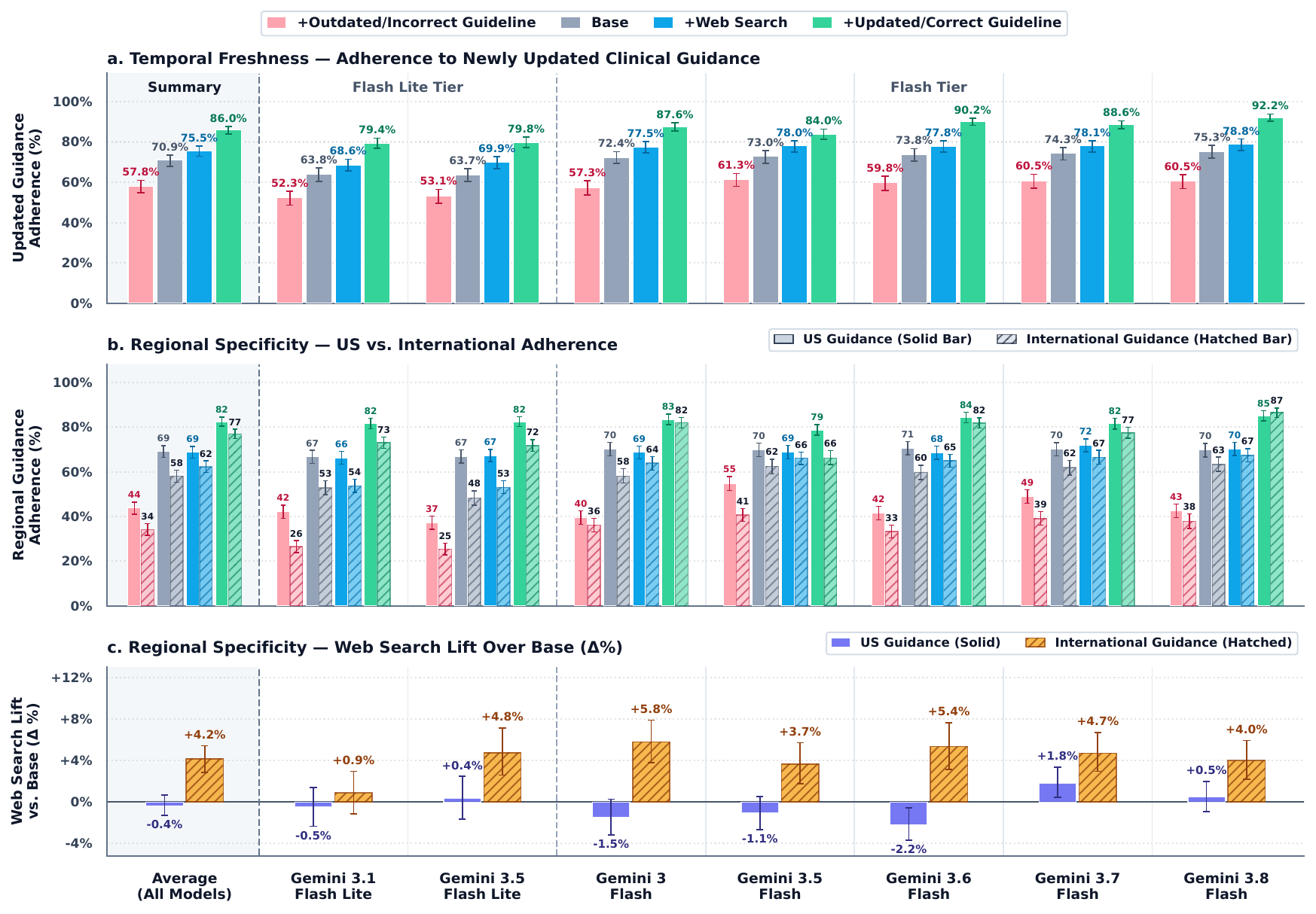}
  \vspace{0.1cm}
  \caption{\textbf{Performance on scenarios targeting temporal and regional guideline differences.} 
  \textbf{a}. Temporal freshness adherence ($N = 630$ scenarios) for the cross-model average and seven individual models under outdated guideline context, baseline, web search, and updated guideline context. 
  \textbf{b}. Regional specificity adherence ($N = 731$ US/International scenario pairs) comparing US (solid bars) and International (hatched bars) variants under mismatched regional guidance, baseline, web search, and matched regional guidance. 
  \textbf{c}. Adherence gain from web search (\emph{+Web Search} minus \emph{Base}) on US ($-0.4\text{ pp}$ average) versus International ($+4.2\text{ pp}$ average) scenarios. Error bars indicate $95\%$ CIs.}
  \label{fig:targeted_diffs}
\end{figure}

\paragraph{Resolving Regional Discordance (US vs.\ International).}
On the Regional Specificity suite (\cref{fig:targeted_diffs}b), ungrounded models score substantially higher on US scenarios than on paired International variants across all seven models ($+11.1\text{ pp}$ on average). Enabling web search (\emph{+Web Search}; \cref{fig:targeted_diffs}b,c) selectively lifts International adherence ($+4.2\text{ pp}$ on average, versus $-0.4\text{ pp}$ on US scenarios), partially narrowing this gap. Direct guideline conditioning again performs best: providing the conflicting regional edition (\emph{+Outdated/Incorrect Guideline}) sharply degrades adherence on both US and International cases, whereas providing the matched regional edition (\emph{+Updated/Correct Guideline}) boosts International adherence by $+18.9\text{ pp}$ on average and cuts the cross-model US--International gap roughly in half (from $11.1\text{ pp}$ to $5.4\text{ pp}$), largely closing it on most \emph{Flash} models (\cref{fig:targeted_diffs}b).

\subsection{Primary Care Physician Reader Study and the Impact of AI Assistance}
\label{results:pcp_results}

Across the 200-scenario physician reader study (\cref{fig:clinician_baseline}; case examples in \cref{appendix:representative_pcp_examples}), unassisted primary care physicians (\emph{PCP Alone}) achieved $50.5\%$ guideline adherence, with remaining plans mostly rated as Moderate or Minor Deviations and rarely as Major Deviations ($2.0\%$). Lower unassisted scores largely reflected omitted condition-specific diagnostic workups, first-line or subspecialty therapies, medication-safety checks, or acute triage steps, particularly on hospital and specialist cases outside routine outpatient scope.

After reviewing the AI recommendation (\emph{PCP + AI}), overall adherence rose from $50.5\%$ to $70.0\%$, with tier upgrades ($25.5\%$) far outnumbering downgrades ($3.0\%$) and Major Deviations declining to $1.0\%$. Adherence improved across all three care settings (\cref{fig:clinician_baseline}a; \cref{appendix:pcp_study_protocol}), most notably in \emph{Specialist Care} ($30.2\%$ to $60.4\%$). Consistent with this pattern, physicians kept their initial plans unchanged more often on familiar \emph{Primary Care (Clinic)} cases ($22.1\%$) than on \emph{Emergency / Inpatient Generalist} ($13.5\%$) or \emph{Specialist Care} cases ($7.5\%$; \cref{fig:clinician_baseline}b). Most revisions added differential diagnoses, diagnostic workups, counseling, medications, or specialist referrals, whereas detrimental edits such as dropping a required test or adding an inappropriate action were rare (\cref{fig:clinician_baseline}c).

\begin{figure}[htbp!]
  \centering
  \includegraphics[width=\textwidth,height=0.73\textheight,keepaspectratio]{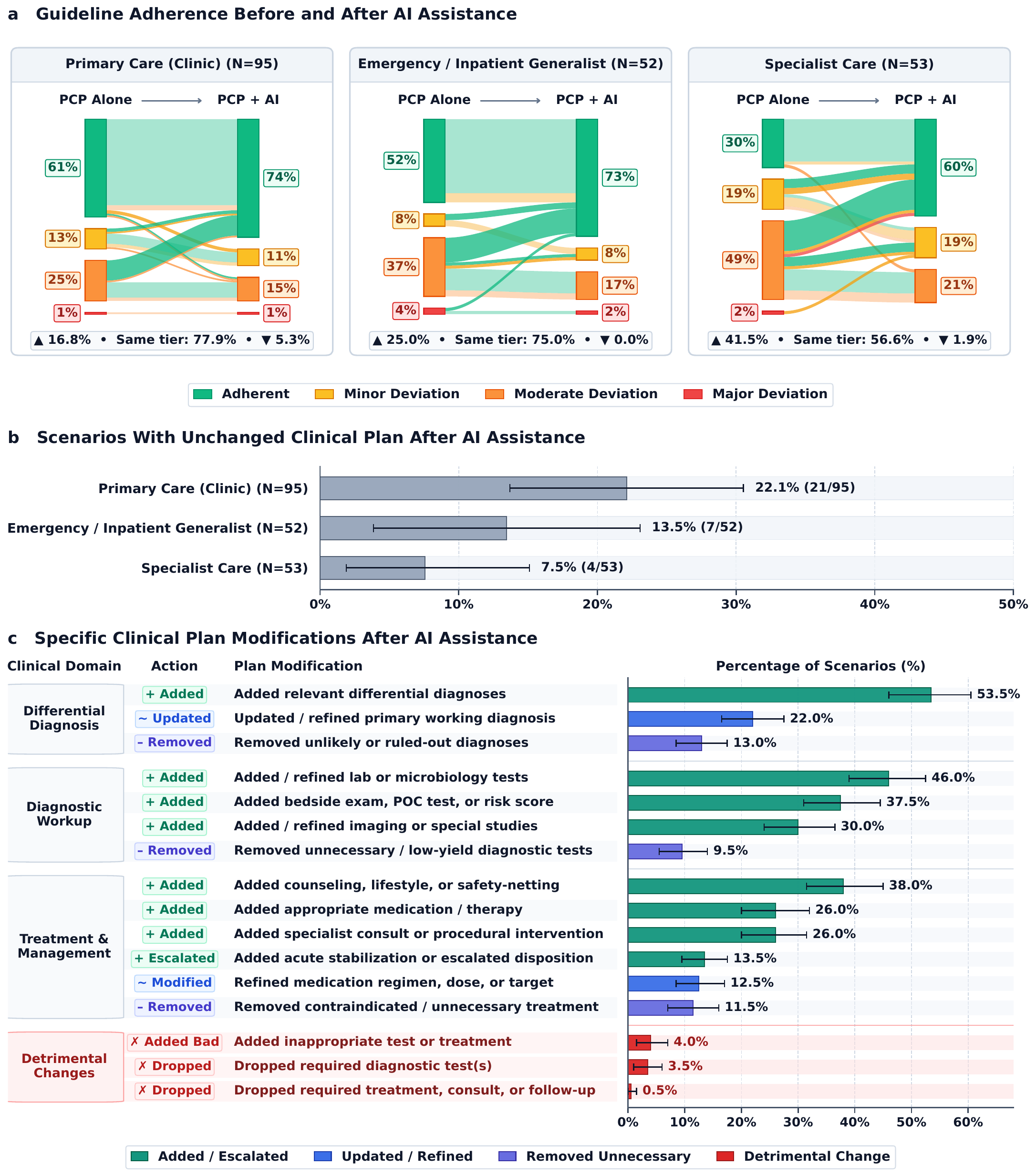}
  \vspace{-0.1cm}
  \caption{\textbf{Primary Care Physician Baseline and Impact of AI Assistance Across the Care Continuum.} 
  \textbf{a}. Ordinal rating distributions (from \emph{Adherent} [Score 4] through \emph{Major Deviation} [Score 1]) and tier-transition summaries ($\blacktriangle$ upgrades, unchanged tier, $\blacktriangledown$ downgrades) before (\emph{PCP Alone}) and after (\emph{PCP + AI}) AI assistance across the three post-hoc care settings (Primary Care (Clinic), $N=95$; Emergency / Inpatient Generalist, $N=52$; Specialist Care, $N=53$), graded by the clinician-calibrated autorater; vertical bars show physician adherence, while connecting ribbons are colored by the standalone AI recommendation's appropriateness tier (darker ribbons indicate tier changes). 
  \textbf{b}. Proportion of scenarios with an unchanged clinical plan after AI assistance across care settings, and \textbf{c}. specific plan modifications across Differential Diagnosis, Diagnostic Workup, Treatment \& Management, and Detrimental Changes, both quantified using Gemini 3.8 Flash to compare paired pre- and post-AI plans (error bars indicate $95\%$ CIs across 1,000 care-setting-stratified scenario bootstrap resamples).}
  \label{fig:clinician_baseline}
\end{figure}

\section{Discussion}
\label{sec:discussion}

\subsection{Scaling Open-Ended Clinical Evaluation Across the Care Continuum}
\label{subsec:discussion_precision}

Our evaluation approach enables automatic, scalable, and comprehensive generation of verified open-ended clinical scenarios from evidence-based practice recommendations. By programmatically traversing the extracted structured criteria and treatment trees, this method delivers not only broad coverage across conditions, but also precise targeting of nuanced, branch-specific clinical presentations (such as specific comorbidities, contraindications, severity cutoffs, or complications) that manual authoring may not always cover. Extending this same automated workflow to paired regional and longitudinal editions similarly pinpoints concrete clinical divergences and converts them into targeted stress tests.  We use the resulting scenarios to evaluate various Gemini models, inference configurations, and human clinicians as part of a reader study of AI-assisted clinician decision-making. While we apply this approach to a particular corpus of \emph{BMJ Best Practice} clinical monographs in this work, the method could be extended to other guideline or recommendation corpora.

Across the core benchmark, model performance varies substantially by stage of care (\cref{fig:system_scaling_context}b): the strongest models approach or exceed $90\%$ adherence on \emph{Differential Diagnosis} and \emph{Complications}, whereas \emph{Investigations} and \emph{Treatment} remain harder because AI outputs more often omit condition-specific diagnostic tests, branch-specific therapies, pre-treatment safety checks, or conditional escalation steps.

Closing these gaps benefits from both model scaling and test-time scaffolding (\cref{fig:system_scaling_context}a,c). Web search and multi-sample synthesis (\emph{Synthesize-4}) appear to address complementary weaknesses: search surfaces external details, while synthesis reconciles omissions across candidate drafts, producing additive gains across all five standard \emph{Flash} models. Smaller \emph{Flash Lite} models, by contrast, gain little from web search, suggesting a potential area for improvement or tuning. More broadly, the modest overall gain from web search on the core scenarios ($+0.8\text{ pp}$ on average) may reflect their derivation from US-edition monographs, consistent with our targeted regional evaluation in which web search had minimal impact on US scenarios ($-0.4\text{ pp}$) but substantially larger gains on International scenarios ($+4.2\text{ pp}$; \cref{fig:targeted_diffs}b,c), likely because base models already encode US-centric clinical knowledge more strongly.

\subsection{Retrieval Gaps and the Long-Context Comprehension Bottleneck}
\label{subsec:discussion_comprehension}

Although conditioning on clinical guidelines substantially improves adherence, our closed-corpus ablations reveal two bottlenecks between accessing a guideline library and applying it accurately (\cref{fig:system_scaling_context}a,c; \cref{appendix:dynamic_retrieval,appendix:extended_results}). First, dynamic retrieval faces a recall bottleneck when initial presentations are undifferentiated. When Gemini 3.8 Flash selects up to five guidelines by title, recall reaches $97.3\%$ on \emph{Treatment}, $88.8\%$ on \emph{Investigations}, and $88.0\%$ on \emph{Complications} cases, where the primary condition is usually known or suspected, but drops to $56.1\%$ on \emph{Differential Diagnosis} cases, where symptoms could reflect many conditions (\cref{appendix:dynamic_retrieval}).

Second, even when the right monograph is retrieved or provided directly, condensed structured excerpts consistently outperform full guidelines (\cref{appendix:oracle_residual}). Because full guidelines span multiple patient subgroups, severity levels, and lines of therapy, placing the entire monograph in context introduces competing branches that can distract model reasoning~\cite{shi2023large}, particularly in smaller \emph{Flash Lite} models. Expanding retrieval capacity $K$ shows a similar trade-off: although the model self-regulates below the capacity cap (averaging $3.84$ retrieved monographs at $K = 5$ and $6.22$ at $K = 9$), recall gains beyond $K = 3$ are largely offset by additional competing monographs in context (\cref{appendix:dynamic_retrieval}). Even with oracle excerpts, adherence remains slightly below $100\%$ ($96.2\%$ on Gemini 3.8 Flash), reflecting occasional cross-section dependencies, deferring or hedging immediate guideline actions, anchoring on salient clinical cues over explicit numeric cutoffs, and strict grading on secondary details such as pre-treatment screens (\cref{appendix:oracle_residual}). Together, these findings show that effective clinical grounding depends not only on retrieving the right guideline, but also on filtering out competing branches and faithfully applying patient-specific rules and thresholds.

\subsection{Adapting to Evolving Evidence and Regional Practice Standards}
\label{subsec:discussion_regional_temporal}

Because clinical evidence evolves continuously~\cite{shekelle2001validity} and practice patterns vary across health systems, static models alone are insufficient for long-term or global deployment. On our \emph{Temporal Freshness} and \emph{Regional Specificity} suites (\cref{fig:targeted_diffs}; scenario examples in \cref{appendix:illustrative_regional_freshness_examples}), ungrounded models often fall back on superseded guidance or US thresholds; on International cases, models might adopt local terminology (such as NHS pathways, \texttt{999} transfer, or Hartmann's solution) while still applying more US-typical diagnostic or treatment criteria. Consistent with results on temporal question answering~\cite{vu2024freshllms}, web search improves adherence on recently updated and International guidance, yet trails direct guideline conditioning because single-pass web search can miss recent revisions or surface competing US and older sources.

Even when the updated or regionally matched full monograph is provided directly in context, adherence remains well below $100\%$ (especially on Regional Specificity, averaging $82.4\%$ on US and $77.0\%$ on International scenarios across models), reflecting both fine-grained clinical differences and the difficulty of isolating patient-specific rules within a full-length guideline. Conversely, strong receptivity to in-context text can hinder performance when outdated or regionally mismatched prompt evidence conflicts with pretrained knowledge~\cite{xie2024adaptive}: supplying a superseded or conflicting regional guideline pushes adherence below the ungrounded baseline across all models. At the same time, these drops confirm that models are generally capable of faithfully follow in-context clinical guidance, a necessary capability when context is used to steer outputs toward local practice standards. Narrowing these gaps will require stronger regional understanding in base models, search systems that prioritize current, jurisdiction-matched sources, and model reasoning that understands how best to interpret nuanced differences within the context.

\subsection{Human--AI Collaboration and Clinical Decision Support}
\label{subsec:discussion_human_ai}

Our study with primary care physicians ($n = \numpcps$) highlights both the assistive value and the potential workflow implications of clinical AI (\cref{fig:clinician_baseline}). Unassisted physicians achieved $50.5\%$ guideline adherence, largely due to omitting condition-specific diagnostic workups, first-line or subspecialty therapies, or pre-treatment safety checks rather than proposing unsafe actions. This baseline should be interpreted in context: more than half of the sampled scenarios involve hospital or subspecialty protocols outside routine outpatient primary care, and clinicians typically write concise, prioritized notes rather than exhaustive narratives. Under real-world visit-time and healthcare resource constraints~\cite{irving2017international,cabana1999don}, physicians focus on high-yield next steps, stage diagnostic workups across visits, and ration low-yield tests and specialist referrals, whereas an LLM incurs no clinical or financial cost when listing every textbook test, risk score, and precautionary consultation in a single response. Because recall-oriented rubrics penalize omissions more heavily than extra testing or referrals, models can score higher through checklist-style completeness.

Reviewing AI recommendations raised overall adherence to $70.0\%$, consistent with randomized vignette studies of AI management support~\cite{goh2025gpt}. Physicians left their initial plans unchanged more often on familiar \emph{Primary Care (Clinic)} cases ($22.1\%$) than on \emph{Specialist Care} cases ($7.5\%$), where adherence doubled ($30.2\%$ to $60.4\%$; \cref{fig:clinician_baseline}a,b). Gains came mainly from adding AI-suggested diagnostic workups, treatments, counseling, and referrals (\cref{fig:clinician_baseline}c), predominantly improving guideline adherence with few inappropriate additions. Appropriateness notwithstanding, closing omissions via added steps has potential implications for utilization and cost. Prior work has found that LLMs tend to recommend broader testing and referrals at higher projected cost than treating physicians~\cite{aran2026comparative}, and that clinicians outperform conversational AI on the practicality and cost-effectiveness of management plans in prospective ambulatory care~\cite{brodeur2026conversational}. Rather than adopting every AI suggestion, however, physicians often selectively filtered the recommendations, drawing on the AI most heavily to fill gaps on out-of-scope hospital and \emph{Specialist Care} cases.

Importantly, AI assistance was predominantly beneficial and produced very few detrimental edits (\cref{fig:clinician_baseline}a,c). Across all $200$ scenarios, tier upgrades ($25.5\%$, $51/200$) outnumbered downgrades ($3.0\%$, $6/200$) more than eight-fold, while detrimental plan changes were rare: $4.0\%$ added an inappropriate test or treatment, $3.5\%$ dropped a required diagnostic test, and $0.5\%$ dropped a required treatment, consult, or follow-up. Qualitative case review illustrates a physician rejecting an unsound AI recommendation (premature levothyroxine before confirmatory testing; \cref{fig:pcp_case_box_2_adherent_despite_flawed_ai}) while selectively adopting appropriate diagnostic workups and safety checks across cases (\cref{appendix:representative_pcp_examples}).

\subsection{Limitations}
\label{subsec:limitations}

Our study has several important limitations.

\textbf{Guidelines as a Reference Standard versus Individualized Care.} Point-of-care syntheses such as \emph{BMJ Best Practice} provide a structured baseline for evidence-based care, but represent a single editorial source that may differ from other national or specialty society guidelines, or local practice patterns. Moreover, strict adherence to a disease-specific monograph is an imperfect proxy for clinical quality: real-world care frequently requires departing from standardized pathways to account for patient preferences, local resource constraints~\cite{cabana1999don}, and competing priorities across co-occurring conditions particularly in older or medically complex patients~\cite{tinetti2004potential,boyd2005clinical}.

\textbf{Static Text Vignettes versus Interactive, Multimodal Care.} Our benchmark presents pre-compiled patient histories, physical examination findings, and baseline test results in single-turn text vignettes. Real-world care, by contrast, requires actively gathering information through iterative history-taking~\cite{tu2025towards,johri2025evaluation}, interpreting multimodal inputs (such as medical imaging, electrocardiograms, and exam signs), adjusting management longitudinally across follow-up visits as a patient's condition evolves~\cite{lievin2026towards}, and incorporating patient preferences for shared decision making.

\textbf{Automated Scenario Extraction and Verification.} Although automated extraction and verification enable broad coverage across 436 clinical monographs, our scenarios are synthetic vignettes that were not individually authored or audited by specialist clinicians. Qualitative review of evaluation traces identified occasional clinical setting ambiguities (e.g., emergency stabilization versus inpatient transition) and rare parsing artifacts in extracted decision trees.

\textbf{Geographic and Health-System Representation.} While our Regional Specificity suite evaluates divergences across 82 paired US and International editions of \emph{BMJ Best Practice}, both editions come from a single publisher and primarily reflect high-income healthcare systems (US and UK/European standards). They do not capture all of the broader guideline adaptations needed in low- and middle-income settings, where diagnostic availability, essential drug formularies, and disease epidemiology differ substantially.

\textbf{Automated Grading and Model Scope.} Despite calibration against physician ratings and perturbation testing (\cref{appendix:autorater_calibration}), grading open-ended plans with an LLM autorater involves subjectivity near tier boundaries. In addition, the reported physician--autorater agreement (\cref{tab:supp_autorater_confusion}) reflects calibration on AI-generated plans with only one physician-rated Major Deviation. Independent blinded validation across both AI-generated and physician-authored plans, including datasets enriched for severe safety errors, as well as on the targeted temporal and regional suites, is therefore needed to establish generalizability. Moreover, although the autorater shows low sensitivity to stylistic verbosity when clinical actions are held fixed (\cref{appendix:autorater_consistency}), rubric-based grading against detailed guideline criteria may inherently reward exhaustive, checklist-style responses that mirror LLMs' documented tendency toward over-testing~\cite{aran2026comparative}. Finally, using Gemini models across extraction, scenario generation, and grading could introduce model-family bias, and restricting our model ablations to the Gemini Flash and Flash Lite tiers leaves evaluation across larger frontier models and other model families for future work.

\textbf{Interpreting Physician Performance.} Our physician study evaluates how AI recommendations can support clinician planning within the context of our generated scenarios, not whether AI outperforms primary care physicians. Beyond the case-mix and completeness effects discussed above, real-world practice relies on safeguards absent from static free-text vignettes---including alternative clinical references, electronic health record order sets, pharmacist review, and longitudinal follow-up visits that routinely catch unstated details. Higher post-AI scores therefore may at least partially reflect greater documented thoroughness rather than proving gains in clinical value or cost-effectiveness~\cite{aran2026comparative,brodeur2026conversational}.

\textbf{Clinician Study Design.} Our reader study involved a pool of \numpcps primary care physicians across 200 text vignettes (one physician per scenario), so scores and revision patterns may partly reflect individual note-writing style and openness to AI suggestions. Physicians revised their plans immediately after viewing the AI output without a control group revising plans a second time without AI, so post-AI gains may reflect added review time. Larger prospective studies embedded in live clinical workflows are needed to determine how AI decision support affects in-context clinician decisions, resource use, and patient outcomes.

\section{Conclusion}
\label{sec:conclusion}

We present an automated pipeline with built-in verification that converts clinical practice recommendations into structured clinical criteria and treatment decision trees, generating verified open-ended scenarios across the care continuum and targeted suites isolating regional and temporal practice differences. Using the 4,561 scenarios generated from 436 \emph{BMJ Best Practice} clinical monographs alongside a clinician-calibrated autorater, we find that model scaling, multi-sample synthesis, and search improve standalone adherence, and that AI assistance helps primary care physicians write more complete plans. At the same time, clear gaps persist in treatment planning, guideline retrieval on undifferentiated presentations, long-context guideline use, and adaptation to updated or International guidance, underscoring the need for living, guideline-grounded evaluations as medical evidence and clinical AI systems evolve.

\subsubsection*{Acknowledgments}
This project was an extensive collaboration across teams at Google Research and Google DeepMind, in partnership with BMJ. We thank Hieu Hoang and Yishay Mansour for their comprehensive review and feedback on the manuscript. We also thank the primary care physicians who participated in our evaluation study for their time and clinical expertise, as well as Yvonne Ridge, Chris Wroe, and Derrick Leung at BMJ for their partnership, helpful discussions, and feedback on the manuscript. Finally, we are grateful to our colleagues Edna Agyemang, Jackie Barr, Joëlle Barral, Pinal Bavishi, Alex Bijamov, Joe Breda, Thomas Buckley, Charlotte Caucheteux, Amira Fouad, Kavi Goel, Ethan Goh, Celeste Grade, Tiffany Guo, Naama Hammel, Jen Haslip, Brett Hatfield, Rebecca Hemenway, Aziza Henry, Michael Howell, Ayush Jain, Xiang Ji, Cat Kozlowski, Bhawesh Kumar, Jihyeon Lee, Maya Leventer-Roberts, Valentin Li\'evin, SiWai Man, Ulysse Marteau-Ferey, Alicia Mart\'in, Jacinda Mein, Ines Mezerreg, Isha Mishra, Gaurav Mohta, Dillon Obika, Flavien Prost, David Racz, Roma Ruparel, Mike Sanchez, Rory Sayres, Sam Schmidgall, Rachelle Sico, Preeti Singh, Tim Strother, Cara Tan, Susan Thomas, Tao Tu, Gordon Turner, Will Vaughan, Sunny Virmani, Jessica Williams, Angelica Willis, Ellery Wulczyn, Joseph Xu, and Lin Yang for their valuable feedback, discussions, and support throughout this work.

\subsubsection*{Data Availability}
The \emph{BMJ Best Practice} clinical monographs used in this study are licensed from BMJ Publishing Group Limited; accordingly, the source monographs and derived benchmark scenarios cannot currently be released. Illustrative evaluation scenarios and clinician trajectories are provided in the manuscript and Supplementary Information, alongside full extraction algorithms, structured schemas, and prompt templates so that the methodology can be reproduced on other clinical corpora.

\subsubsection*{Code Availability}
To facilitate reproducibility, we have included structured schemas, prompts, and pipeline details for decision tree and structured clinical item extraction (\cref{appendix:corpus_and_schemas}), scenario generation (\cref{appendix:extraction_pipeline_details}), inference harnesses (\cref{appendix:inference_harnesses}), and clinician-calibrated autorating (\cref{appendix:autorater_calibration}). Our work utilizes various publicly available Gemini models, with details available in the corresponding model cards (e.g.,~\cite{gemini3flash}).

\subsubsection*{Competing Interests}
This study was funded by Alphabet Inc.\ and/or a subsidiary thereof (``Alphabet''). Authors who are employees of Alphabet may own stock as part of the standard compensation package.
\end{refsegment}

\newpage
\setlength\bibitemsep{3pt}
\printbibliography[title={References}]
\clearpage

\begin{refsegment}
\newpage
\appendix
\setcounter{figure}{0}
\setcounter{table}{0}
\renewcommand{\thefigure}{SI\arabic{figure}}
\renewcommand{\thetable}{SI\arabic{table}}
\renewcommand{\theHfigure}{SI.\arabic{figure}}
\renewcommand{\theHtable}{SI.\arabic{table}}

\begin{center}
\LARGE\textbf{Supplementary Information}
\end{center}
\vspace{0.3cm}

This Supplementary Information provides methodological details, validation analyses, and extended experimental results supporting the main text:
\begin{enumerate}[leftmargin=1.5em, itemsep=2pt]
    \item \textbf{\cref{appendix:corpus_and_schemas}:} Clinical Guideline Corpus and Structured Extraction
    \item \textbf{\cref{appendix:extraction_pipeline_details}:} Scenario Generation and Quality Control
    \item \textbf{\cref{appendix:inference_harnesses}:} Inference Harnesses and Context Configurations
    \item \textbf{\cref{appendix:autorater_calibration}:} Clinician-Calibrated Autorater Rubric, Calibration, and Consistency
    \item \textbf{\cref{appendix:pcp_study_protocol}:} Primary Care Physician Baseline and AI-Assistance Study
    \item \textbf{\cref{appendix:dynamic_retrieval}:} Closed-Corpus Dynamic Guideline Retrieval Mechanics
    \item \textbf{\cref{appendix:extended_results}:} Extended Model and Inference Harness Performance
    \item \textbf{\cref{appendix:oracle_residual}:} Analysis of Residual Non-Adherence Under Oracle Excerpt and Oracle Guideline Conditioning
    \item \textbf{\cref{appendix:illustrative_regional_freshness_examples}:} Illustrative Targeted Scenario Examples
    \item \textbf{\cref{appendix:representative_pcp_examples}:} Illustrative Primary Care Physician AI-Assistance Examples
\end{enumerate}

\clearpage
\section{Clinical Guideline Corpus and Structured Extraction}
\label{appendix:corpus_and_schemas}

Our evaluation corpus comprises 436 structured XML clinical monographs from \emph{BMJ Best Practice}~\cite{bmj_best_practice}: 354 United States (US) editions whose evidence approach focuses on US clinical guidelines and practice (using US drug nomenclature [USANs] and appropriate units for test results and clinical parameters), and 82 paired International editions whose general evidence approach includes US clinical guidelines in addition to those from major UK, European, Oceania, and other global guideline producers (using recommended International Non-proprietary Names [rINNs] for drug nomenclature and appropriate units for test results and clinical parameters, with a subset of topics anchored primarily in UK and European clinical guideline sources).

Before extraction, we strip non-clinical metadata (contributor lists, and for paired comparisons, references, synonyms, and resource/patient-access sections) from each XML monograph while preserving its clinical section hierarchy. Using Gemini 3 Flash with structured JSON output, we parse each monograph into four clinical categories (\cref{fig:lst_extraction_schemas,fig:lst_extraction_prompts,fig:lst_extraction_prompts_elements}): branching treatment decision trees, differential diagnoses, diagnostic investigations, and complications. For treatment sections, narrative guidance and tables are converted into a branching tree of management steps, where each node specifies the patient subpopulation or clinical criteria, rationale, prioritized treatments (first-line, second-line, or adjunct) with alternative options and dosing notes, and follow-up steps based on treatment response. For targeted regional and temporal evaluations, Gemini 3.8 Flash compares paired monograph editions to extract structured records of substantive clinical differences (\cref{fig:lst_extraction_schemas}; prompts in \cref{fig:lst_targeted_diff_prompts}).

\subsection{Intermediate Extraction Verification and Refinement}
\label{appendix:extraction_verification}

Before scenario synthesis, each extracted JSON record (\texttt{TreatmentTree}, \texttt{DifferentialResults}, \texttt{InvestigationResults}, and \texttt{ComplicationResults}) is verified against its cleaned source XML and refined with Gemini 3 Flash under a strict zero-issue policy (\texttt{ValidationReport}: \texttt{is\_valid = true} only when \texttt{issues} is empty; prompts in \cref{fig:lst_extraction_verification_prompts}):
\begin{enumerate}[leftmargin=1.5em, topsep=2pt, itemsep=1pt, parsep=0pt]
  \item \textbf{Deterministic structural rules:} Validators verify non-empty \texttt{main\_condition} and \texttt{root\_node}, tree acyclicity (DFS cycle check), non-empty \texttt{condition\_or\_criteria} per node, a populated \texttt{treatment\_plan} on every terminal leaf (\texttt{next\_steps == []}), and a recommendation or non-empty options per element. For structured items, checks require primary names (\texttt{test\_name}, \texttt{condition}, \texttt{complication}) and descriptive fields (findings; differentiating signs/symptoms) as \texttt{ERROR}s, and flag case-insensitive duplicates as \texttt{WARNING}s.
  \item \textbf{Category-specific LLM semantic critique:} Four dedicated prompts compare each extracted JSON against the source XML for inaccuracies, omissions, inconsistencies, and over-simplification. Unlike scenario-level checks (which allow plausible presentation details; \cref{appendix:extraction_pipeline_details}), intermediate checks enforce \emph{strict groundedness}: any external drug, dose, test, differential, complication, timeframe, or likelihood absent from the XML is flagged as a critical \texttt{ERROR}, even if medically accurate.
  \item \textbf{Independent direct XML audit:} A separate audit evaluates each extracted artifact directly against the cleaned XML across four dimensions---\texttt{safety} (hazards and contraindications), \texttt{hallucination} (strict groundedness), \texttt{omission}, and \texttt{accuracy}---emitting a report.
  \item \textbf{Targeted inline refinement:} Flagged records undergo surgical JSON edits using the current JSON, cleaned XML, and structured issue feedback (\texttt{severity}, \texttt{category}, \texttt{message}, JSON \texttt{path}, and \texttt{suggested\_fix}) while preserving intact branches and items, followed by re-validation before scenario generation.
\end{enumerate}

\begin{figure}[ht!]
\centering
\begin{tcolorbox}[
  colback=gray!4,
  colframe=black!70,
  colbacktitle=black!75,
  coltitle=white,
  boxrule=0.6pt,
  arc=2pt,
  left=7pt, right=7pt, top=4.5pt, bottom=4.5pt,
  toptitle=2.5pt, bottomtitle=2.5pt,
  fonttitle=\footnotesize\bfseries\raggedright,
  title={Structured Extraction Schemas Across Clinical Categories and Paired Guideline Comparisons}
]
\fontsize{8.2pt}{9.5pt}\selectfont\RaggedRight
\setlength{\parskip}{2.5pt}
\noindent\textbf{1. Treatment Decision Tree (\texttt{TreatmentTree})}\\
\textbf{Condition \& root node (\texttt{main\_condition}, \texttt{root\_node}):} Primary condition name and recursive tree of clinical management steps (\texttt{TreatmentStep}).\\
\textbf{Management step node (\texttt{TreatmentStep}):} Patient subpopulation or clinical criteria (\texttt{condition\_or\_criteria}), clinical rationale (\texttt{rationale}), additive list of concurrent treatments (\texttt{treatment\_plan}), and child branches (\texttt{next\_steps}) representing sequential progression if further therapy is needed or mutually exclusive patient states.\\
\textbf{Treatment element (\texttt{TreatmentElement} \& \texttt{TreatmentOption}):} Primary intervention (\texttt{treatment\_recommendation}), priority tier (\texttt{treatment\_importance}: first-line, second-line, or adjunct), rationale (\texttt{element\_rationale}), clinical/dosing notes (\texttt{notes}), and acceptable alternatives (\texttt{treatment\_options} with \texttt{option\_description}, \texttt{option\_rationale}, and \texttt{option\_additions}).

\noindent\textbf{2. Differential Diagnosis (\texttt{DifferentialResults})}\\
\textbf{Differential entry (\texttt{DifferentialItem}):} Competing condition name (\texttt{condition}), distinguishing clinical signs and symptoms (\texttt{differentiating\_signs\_symptoms}), and differentiating diagnostic tests (\texttt{differentiating\_tests}).

\noindent\textbf{3. Diagnostic Investigations (\texttt{InvestigationResults})}\\
\textbf{Investigation entry (\texttt{InvestigationItem}):} Diagnostic test name (\texttt{test\_name}), expected findings or diagnostic purpose (\texttt{findings}), and priority flag (\texttt{is\_first\_line}: initial first-line workup vs.\ subsequent or selective testing).

\noindent\textbf{4. Complications (\texttt{ComplicationResults})}\\
\textbf{Complication entry (\texttt{ComplicationItem}):} Complication name (\texttt{complication}), onset timeframe (\texttt{timeframe}: e.g., acute vs.\ long-term), clinical likelihood (\texttt{likelihood}), and clinical description (\texttt{description}).

\noindent\textbf{5. Paired Guideline Differences (\texttt{VerifiedRegionalDiff} \& \texttt{VerifiedFreshnessDiff})}\\
\textbf{Difference record:} Shared clinical scenario context (\texttt{clinical\_context}), paired recommendations (\texttt{us\_recommendation} vs.\ \texttt{row\_recommendation}, with \texttt{row\_applicable\_countries} and guideline authorities; or \texttt{current\_us\_recommendation} vs.\ \texttt{previous\_us\_recommendation} with \texttt{previous\_version\_date} and \texttt{current\_version\_date}), verbatim XML quotes (\texttt{us\_xml\_verbatim\_quote}/\texttt{row\_xml\_verbatim\_quote} or \texttt{old\_xml\_verbatim\_quote}/\texttt{new\_xml\_verbatim\_quote}), supporting rationale (\texttt{evidence\_or\_rationale}, \texttt{verification\_rationale}), and verification confidence score (\texttt{confidence\_score}, 1--5).
\end{tcolorbox}
\caption{Summary of structured JSON fields extracted from each clinical monograph across the four clinical categories and paired guideline comparisons.}
\label{fig:lst_extraction_schemas}
\end{figure}

\begin{figure}[ht!]
\centering
\begin{tcolorbox}[
  colback=gray!3,
  colframe=black!70,
  colbacktitle=black!75,
  coltitle=white,
  boxrule=0.6pt,
  arc=2pt,
  left=7pt, right=7pt, top=5pt, bottom=5pt,
  toptitle=3pt, bottomtitle=3pt,
  fonttitle=\footnotesize\bfseries\raggedright,
  title={Treatment Decision Tree Extraction Prompt (\texttt{TreatmentTree})}
]
\footnotesize\RaggedRight
\setlength{\parskip}{3pt}
\begin{tcolorbox}[colback=white, colframe=black!25, boxrule=0.4pt, arc=1.5pt, left=6pt, right=6pt, top=4pt, bottom=4pt]
\RaggedRight
\noindent\textbf{Role (\texttt{<role>}):} You are an expert medical informatics assistant specialized in parsing clinical evidence based practice recommendations. Your task is to read the provided evidence based practice recommendation and convert its free-text logic into a structured decision tree.\\[3pt]
\noindent\textbf{Instructions (\texttt{<instructions>}):}
\begin{enumerate}[leftmargin=1.4em, topsep=1pt, itemsep=1.5pt, parsep=0pt]
  \item \textbf{Main Goal:} Analyze the text in the \texttt{<input\_document>} tag and generate a single, complete JSON object that perfectly matches the schema template provided in the \texttt{<structured\_logic>} block, following all rules below.
  \item \textbf{Nested Structure:} You MUST output a tree structure. \texttt{next\_steps} MUST be a list of \texttt{TreatmentStep} objects representing the conditional paths or branches that follow from the current step. Do NOT output a flat list of nodes.
  \item \textbf{Node Logic (Branch vs.\ Leaf):}
  \begin{itemize}[leftmargin=1.2em, topsep=0pt, itemsep=0pt]
    \item \emph{Leaf Node (terminal step):} \texttt{next\_steps} MUST be an empty list \texttt{[]}; \texttt{treatment\_plan} MUST be a non-empty list of one or more \texttt{TreatmentElement} objects populated with the specific treatments recommended for this outcome.
    \item \emph{Branch Node (decision point):} \texttt{next\_steps} MUST be a list containing one or more \texttt{TreatmentStep} objects; \texttt{treatment\_plan} can be null or a list of \texttt{TreatmentElement} objects if treatments are recommended at this intermediate step.
  \end{itemize}
  \item \textbf{Treatment Structure:}
  \begin{itemize}[leftmargin=1.2em, topsep=0pt, itemsep=0pt]
    \item \emph{Alternatives (``OR''):} If several drugs or procedures are listed as equivalent alternatives (e.g., ``use Drug A OR Drug B'', or distinct non-pharmacological/supportive measures), list them as multiple \texttt{TreatmentOption} objects \emph{within the same} \texttt{TreatmentElement}.
    \item \emph{Additives (``AND''):} If treatments are additive (e.g., ``Drug A AND Drug B''), list them as \emph{separate} \texttt{TreatmentElement} objects in \texttt{treatment\_plan}.
    \item \emph{Sequence (``THEN''):} If a treatment is recommended only after another (e.g., ``Start with A, if that fails use B''), B MUST be modeled as a separate child node in \texttt{next\_steps} of A.
  \end{itemize}
  \item \textbf{Implicit Conditions:} Infer implicit conditions. If a treatment is described as ``second-line'' or ``rescue therapy'', infer the condition leading to it (e.g., ``Failure of first-line therapy'') and model it as a child node.
  \item \textbf{Self-Correction \& Leaf Node Constraint:} After generating the JSON, review it against the source text in \texttt{<thinking>} to verify completeness and logical consistency. Every leaf node MUST have a non-empty \texttt{treatment\_plan}.
  \item \textbf{Rationale, Notes, \& Detail:} For every \texttt{TreatmentElement} and \texttt{TreatmentOption}, provide a detailed \texttt{element\_rationale} and \texttt{option\_rationale}. Use \texttt{notes} in \texttt{TreatmentElement} to capture general warnings, what to avoid, or broad context. Capture all drug dosages, treatment durations, specific patient populations, contraindications, side effects, and co-therapies within \texttt{option\_description} or rationale fields without simplifying decision criteria.
  \item \textbf{Path Branching Strategy \& Urgent Considerations:} Ensure presentations requiring fundamentally different diagnostic evaluations, specialties, or management regimens are modeled as distinct conditional branches (\texttt{next\_steps}) rather than collapsed into a single list of options, while avoiding separate branches for minor subtypes whose downstream pathway is identical. Scan the entire monograph (including ``Urgent Considerations'' and ``Emergencies'') to extract immediate rescue treatments and stabilization procedures into their respective leaf nodes.
\end{enumerate}
\vspace{2pt}
\noindent\textbf{Template Blocks:} \texttt{<structured\_logic>\{structured\_logic\}</structured\_logic>} and \texttt{<input\_document>\{input\_document\}</input\_document>}
\end{tcolorbox}
\end{tcolorbox}
\caption{Condensed prompt template used to extract branching treatment decision trees (\texttt{TreatmentTree}) from each clinical monograph (two few-shot JSON exemplars omitted for brevity).}
\label{fig:lst_extraction_prompts}
\end{figure}

\begin{figure}[ht!]
\centering
\begin{tcolorbox}[
  colback=gray!3,
  colframe=black!70,
  colbacktitle=black!75,
  coltitle=white,
  boxrule=0.6pt,
  arc=2pt,
  left=7pt, right=7pt, top=5pt, bottom=5pt,
  toptitle=3pt, bottomtitle=3pt,
  fonttitle=\footnotesize\bfseries\raggedright,
  title={Structured Element Extraction Prompts (Differential Diagnosis, Diagnostic Investigations, and Complications)}
]
\footnotesize\RaggedRight
\setlength{\parskip}{3pt}

\begin{tcolorbox}[colback=white, colframe=black!25, boxrule=0.4pt, arc=1.5pt, left=6pt, right=6pt, top=4pt, bottom=4pt]
\RaggedRight
\noindent\textbf{1. Differential Diagnosis Extraction Prompt (\texttt{DifferentialResults})}\\[2pt]
\textbf{Role (\texttt{<role>}):} You are an expert medical informatics assistant specialized in parsing clinical evidence based practice recommendations. Your task is to read the provided evidence based practice recommendation and extract a structured list of differential diagnoses.\\[2pt]
\textbf{Instructions (\texttt{<instructions>}):}
\begin{enumerate}[leftmargin=1.4em, topsep=1pt, itemsep=1pt, parsep=0pt]
  \item \textbf{Main Goal:} Analyze the text in the \texttt{<input\_document>} tag and generate a single, complete JSON object containing a list of \texttt{DifferentialItem} objects matching \texttt{<structured\_logic>\{structured\_logic\}</structured\_logic>}.
  \item \textbf{Extraction Rules:} (a) \texttt{condition}: The name of the alternative condition; (b) \texttt{differentiating\_signs\_symptoms}: Signs or symptoms that help distinguish this differential from the main condition (specifically what makes this condition different, not just general symptoms); (c) \texttt{differentiating\_tests}: Tests that are specifically useful for distinguishing this differential.
\end{enumerate}
\end{tcolorbox}

\vspace{2pt}
\begin{tcolorbox}[colback=white, colframe=black!25, boxrule=0.4pt, arc=1.5pt, left=6pt, right=6pt, top=4pt, bottom=4pt]
\RaggedRight
\noindent\textbf{2. Diagnostic Investigations Extraction Prompt (\texttt{InvestigationResults})}\\[2pt]
\textbf{Role (\texttt{<role>}):} You are an expert medical informatics assistant specialized in parsing clinical evidence based practice recommendations. Your task is to read the provided evidence based practice recommendation and extract a structured list of medical investigations (tests) mentioned in the text.\\[2pt]
\textbf{Instructions (\texttt{<instructions>}):}
\begin{enumerate}[leftmargin=1.4em, topsep=1pt, itemsep=1pt, parsep=0pt]
  \item \textbf{Main Goal:} Analyze the text in the \texttt{<input\_document>} tag and generate a single, complete JSON object containing a list of \texttt{InvestigationItem} objects matching \texttt{<structured\_logic>\{structured\_logic\}</structured\_logic>}.
  \item \textbf{Extraction Rules:} (a) \texttt{test\_name}: The name of the investigation or test; (b) \texttt{findings}: A description of what the test is looking for or what findings are relevant; (c) \texttt{is\_first\_line}: Set to \texttt{true} if described as an initial, first-line, or screening test; set to \texttt{false} if it is a subsequent, confirmatory, or specialized test ordered only after other tests or if the condition persists (if the text says ``order X to confirm'', \texttt{is\_first\_line} should be \texttt{false}).
\end{enumerate}
\end{tcolorbox}

\vspace{2pt}
\begin{tcolorbox}[colback=white, colframe=black!25, boxrule=0.4pt, arc=1.5pt, left=6pt, right=6pt, top=4pt, bottom=4pt]
\RaggedRight
\noindent\textbf{3. Complications Extraction Prompt (\texttt{ComplicationResults})}\\[2pt]
\textbf{Role (\texttt{<role>}):} You are an expert medical informatics assistant specialized in parsing clinical evidence based practice recommendations. Your task is to read the provided evidence based practice recommendation and extract a structured list of potential complications.\\[2pt]
\textbf{Instructions (\texttt{<instructions>}):}
\begin{enumerate}[leftmargin=1.4em, topsep=1pt, itemsep=1pt, parsep=0pt]
  \item \textbf{Main Goal:} Analyze the text in the \texttt{<input\_document>} tag and generate a single, complete JSON object containing a list of \texttt{ComplicationItem} objects matching \texttt{<structured\_logic>\{structured\_logic\}</structured\_logic>}.
  \item \textbf{Extraction Rules:} (a) \texttt{complication}: The name of the complication; (b) \texttt{timeframe}: When this complication typically occurs (if not explicitly mentioned, use ``Not specified'' or null; do not guess); (c) \texttt{likelihood}: Qualitative description of how common it is (if not explicitly mentioned, use ``Not specified'' or null; do not guess).
\end{enumerate}
\end{tcolorbox}
\end{tcolorbox}
\caption{Prompt templates used to extract differential diagnoses (\texttt{DifferentialResults}), diagnostic investigations (\texttt{InvestigationResults}), and complications (\texttt{ComplicationResults}) from each clinical monograph.}
\label{fig:lst_extraction_prompts_elements}
\end{figure}

\begin{figure}[ht!]
\centering
\begin{tcolorbox}[
  colback=gray!3,
  colframe=black!70,
  colbacktitle=black!75,
  coltitle=white,
  boxrule=0.6pt,
  arc=2pt,
  left=6pt, right=6pt, top=4pt, bottom=4pt,
  toptitle=2.5pt, bottomtitle=2.5pt,
  fonttitle=\footnotesize\bfseries\raggedright,
  title={Prompts for Intermediate Extraction Critique, Independent Direct XML Audit, and Inline Refinement}
]
\fontsize{8.2pt}{9.8pt}\selectfont\RaggedRight
\setlength{\parskip}{2pt}

\begin{tcolorbox}[colback=white, colframe=black!25, boxrule=0.4pt, arc=1.5pt, left=5.5pt, right=5.5pt, top=3pt, bottom=3pt]
{\raggedright\noindent\textbf{1. Category-Specific Semantic Critique Prompts (\texttt{ValidationReport})}\par}\vspace{1.5pt}
\textbf{Role (\texttt{<role>}):} You are an expert medical guideline validator. Critique the extracted structured record (\texttt{<extracted\_data>}) against the source guideline (\texttt{<source\_guideline>}).\\[1.5pt]
\textbf{Shared Strict Groundedness Rule:} Every extracted drug, dosage, diagnostic test, finding, differential, complication, timeframe, or likelihood MUST be explicitly present in the source XML. Flag any injected external clinical logic---even if clinically correct or standard practice---as a critical \texttt{ERROR} (\texttt{category="inaccuracy"}).\\[1.5pt]
\textbf{Category-Specific Focus:}
\begin{itemize}[leftmargin=1.2em, topsep=0pt, itemsep=0.5pt, parsep=0pt]
  \item \emph{Treatment Decision Tree:} Medical inaccuracies (drugs, doses, durations, populations/contraindications, or swapped first- vs.\ second-line ordering); missing conditional branches or alternative options; logical inconsistencies or leaf nodes lacking a \texttt{treatment\_plan}; over-simplification of nuanced \texttt{condition\_or\_criteria} or rationales.
  \item \emph{Diagnostic Investigations:} Unmentioned tests, inverted \texttt{is\_first\_line} flags (initial vs.\ confirmatory/second-line), misleading \texttt{findings} descriptions, omitted tests, and non-investigations (treatments, risk factors, or demographics).
  \item \emph{Differential Diagnoses:} Unsupported differentiating signs/symptoms or tests, main condition erroneously extracted as a differential, and omitted differentials.
  \item \emph{Complications:} Unsupported complications, timeframes, or likelihoods, and omitted complications.
\end{itemize}
\textbf{Output:} Return a single \texttt{ValidationReport} JSON: \texttt{is\_valid:\ true} with an empty \texttt{issues} list if clean; otherwise list each issue with \texttt{severity} (\texttt{"ERROR"} or \texttt{"WARNING"}), \texttt{category} (\texttt{"inaccuracy"}, \texttt{"omission"}, \texttt{"consistency"}, or \texttt{"simplification"}), \texttt{message}, JSON \texttt{path}, and \texttt{suggested\_fix}.
\end{tcolorbox}

\vspace{1.5pt}
\begin{tcolorbox}[colback=white, colframe=black!25, boxrule=0.4pt, arc=1.5pt, left=5.5pt, right=5.5pt, top=3pt, bottom=3pt]
{\raggedright\noindent\textbf{2. Independent Direct XML Audit Prompt (\texttt{target\_type="extracted\_data"})}\par}\vspace{1.5pt}
\textbf{Role (\texttt{<role>}):} You are a senior medical editor and clinical auditor performing a direct, independent audit of an extracted clinical artifact (\texttt{<artifact\_json>}) against the raw clinical guideline text (\texttt{<guideline\_xml>}).\\[1.5pt]
\textbf{Audit Criteria:} (1) \textbf{Clinical Safety Hazards \& Contraindications:} contraindicated, unsafe, or incorrect treatments, drugs, doses, or actions; (2) \textbf{Clinical Hallucinations (Strict Groundedness):} any clinical logic, drug name, diagnostic step, complication, or option not explicitly mentioned in the XML; (3) \textbf{Clinical Omissions:} omitted critical decisions, alternative diagnoses, key test indications, or treatment options; (4) \textbf{Semantic Inaccuracies:} misrepresented clinical criteria, patient profiles, or treatment steps.\\[1.5pt]
\textbf{Output:} Return \texttt{ValidationReport} JSON with \texttt{severity} (\texttt{"ERROR"} for safety hazards, strict hallucinations, critical omissions, and semantic inaccuracies; \texttt{"WARNING"} for minor anomalies), \texttt{category} (\texttt{"safety"}, \texttt{"hallucination"}, \texttt{"omission"}, or \texttt{"accuracy"}), \texttt{message} citing the XML and artifact, \texttt{path}, and \texttt{suggested\_fix}.
\end{tcolorbox}

\vspace{1.5pt}
\begin{tcolorbox}[colback=white, colframe=black!25, boxrule=0.4pt, arc=1.5pt, left=5.5pt, right=5.5pt, top=3pt, bottom=3pt]
{\raggedright\noindent\textbf{3. Targeted Inline Extraction Refinement Prompt}\par}\vspace{1.5pt}
\textbf{Role (\texttt{<role>}):} You are an expert senior medical editor and guideline extraction refiner. Revise \texttt{<extracted\_data>} against \texttt{<source\_guideline>} to resolve the flagged list in \texttt{<validation\_issues>}.\\[1.5pt]
\textbf{Rules:} (1) Apply targeted edits to fields, choices, lists, or sub-nodes addressing every issue's message and suggested fix; (2) preserve all unaffected branches, items, and rationales intact without truncation or over-simplification; (3) conform strictly to the \texttt{<structured\_logic>} JSON schema with raw JSON output only; (4) remove any decision-tree branch representing an out-of-scope alternative condition, differential, or complication managed under a separate guideline (logical dead end); (5) set complication \texttt{timeframe} or \texttt{likelihood} to \texttt{"Not specified"} (or \texttt{"Unknown"}) if absent from the source XML to preserve strict groundedness.
\end{tcolorbox}
\end{tcolorbox}
\caption{Condensed prompt templates used for intermediate verification and targeted refinement of extracted structured records prior to scenario generation.}
\label{fig:lst_extraction_verification_prompts}
\end{figure}

\clearpage
\section{Scenario Generation and Quality Control}
\label{appendix:extraction_pipeline_details}

\subsection{Scenario Synthesis, Sampling, and Targeted Pairing}
\label{appendix:targeted_diff_prompts}
To construct the core evaluation set, we filter the US corpus to exclude broad overview topics and retain condition-specific guidelines whose structured extractions pass automated faithfulness checks against the source XML (\cref{appendix:extraction_verification}). From each extracted treatment decision-tree path, differential diagnosis, diagnostic investigation, or complication item, Gemini 3 Flash generates 2--4 candidate clinical vignettes (\cref{fig:lst_generation_prompts}) spanning diverse clinical presentation archetypes (such as standard textbook, atypical/subtle, borderline/confounding, and acute or high-risk presentations), specifying patient demographics, reason for visit, care setting, concrete vital signs, physical examination and baseline test findings, pertinent negatives, and a ground-truth management plan anchored to the source monograph. Multi-stage verification produces 32,653 validated scenarios, of which 32,621 pass all automated checks without warnings across the four categories. From this verified pool, we sample a balanced core benchmark of 3,200 scenarios (800 per clinical category) across categories and retained guidelines, along with a balanced 200-scenario evaluation set (50 per category; stratified post-hoc into Primary Care (Clinic) [$N=95$], Emergency / Inpatient Generalist [$N=52$], and Specialist Care [$N=53$]; \cref{appendix:pcp_study_protocol}) for the primary care physician study and autorater calibration.

To construct the targeted evaluation suites, we use Gemini 3.8 Flash in a three-stage structured extraction, independent auditor verification, and scenario generation pipeline (\cref{fig:lst_targeted_diff_prompts}) to isolate substantive clinical divergences while ignoring wording or formatting edits. For Temporal Freshness, comparing earlier and updated editions across the 283 US topics revised between corpus snapshots (\cref{sec:scenarios}) identifies 140 topics with substantive clinical updates, yielding 630 scenarios paired with both updated and superseded ground-truth recommendations (\cref{appendix:illustrative_regional_freshness_examples}). For Regional Specificity, comparing the 82 paired US and International editions across diagnostic thresholds, first-line pharmacotherapy, dosing limits, and surgical indications yields 731 paired scenarios (1,462 US and International variants), each paired with its corresponding regional ground-truth recommendations.

\subsection{Automated Verification and Refinement}
Every candidate scenario passes through a three-stage automated verification and refinement pipeline before inclusion (\cref{fig:lst_verification_prompts}):
\begin{enumerate}[leftmargin=1.5em, topsep=1pt, itemsep=0.5pt]
    \item \textbf{Rule-based structural and leakage checks:} Deterministic checks verify required schema and non-placeholder ground-truth fields (patient profile, reason for visit, visit presentation, diagnosis, and management actions), check alignment against the extracted treatment decision tree, and flag or clean residual regional guideline identifiers (such as NICE or NG/CG/QS codes) from vignette text.
    \item \textbf{Two-pass LLM faithfulness verification:} An LLM verifier audits each candidate first against the extracted clinical item or decision-tree path and second directly against the source monograph XML, checking that patient parameters uniquely satisfy all branch preconditions, contain no diagnostic or therapeutic answer leakage, and introduce no unsupported claims. Candidates failing either check are regenerated with structured error feedback for up to five retries.
    \item \textbf{Clinical coherence refinement and auxiliary rubric generation:} Scenarios passing verification can undergo a critique-and-refine pass (applied across all structured-element categories in the benchmark pool) that checks physiological consistency (mapping active medications to documented diagnoses and replacing qualitative labels with concrete vital signs and laboratory values) and records an auxiliary per-scenario criterion checklist in metadata (benchmark scoring itself uses the calibrated 4-point autorater against each scenario's structured ground-truth plan; \cref{appendix:autorater_calibration}).
\end{enumerate}

\begin{figure}[ht!]
\centering
\begin{tcolorbox}[
  colback=gray!3,
  colframe=black!70,
  colbacktitle=black!75,
  coltitle=white,
  boxrule=0.6pt,
  arc=2pt,
  left=6pt, right=6pt, top=4pt, bottom=4pt,
  toptitle=2.5pt, bottomtitle=2.5pt,
  fonttitle=\footnotesize\bfseries\raggedright,
  title={Prompts for Core Clinical Scenario Generation Across Clinical Categories}
]
\fontsize{8.2pt}{9.8pt}\selectfont\RaggedRight
\setlength{\parskip}{2pt}

\begin{tcolorbox}[colback=white, colframe=black!25, boxrule=0.4pt, arc=1.5pt, left=5.5pt, right=5.5pt, top=3pt, bottom=3pt]
{\raggedright\noindent\textbf{1. Treatment Pathway Scenario Generation (\texttt{ScenarioSet})}\par}\vspace{1.5pt}
\textbf{Role (\texttt{<role>}):} You are an expert clinical scenario designer, medical simulator, and specialist clinician. Your task is to create high-fidelity, longitudinal, and extremely detailed patient case narratives (vignettes) based on defined clinical logic and specific treatment paths.\\[1.5pt]
\textbf{Inputs:} \texttt{<treatment\_tree>}, \texttt{<scenario\_path>}, \texttt{<ground\_truth\_treatment\_plan>}, and \texttt{<schema>}.\\[1.5pt]
\textbf{Instructions (\texttt{<instructions>}):}
\begin{enumerate}[leftmargin=1.4em, topsep=0.5pt, itemsep=0.5pt, parsep=0pt]
  \item \textbf{Fundamental Goal:} Generate 2--4 diverse scenarios in \texttt{EvalScenarioSO} format, each representing a distinct, plausible clinical presentation that strictly adheres to \texttt{<scenario\_path>}.
  \item \textbf{Clinical Specification \& Narrative Structure:} Provide a comprehensive narrative following standard medical structure:
  \begin{itemize}[leftmargin=1.2em, topsep=0pt, itemsep=0pt]
    \item \emph{Clinical Setting \& Reason for Visit (CC):} Explicitly state the encounter setting (e.g., ``Emergency Department'', ``Outpatient Clinic'', ``Primary Care Clinic'', ``Inpatient Ward'') and a patient-articulated chief complaint.
    \item \emph{History of Present Illness (HPI):} Longitudinal story (onset, duration, progression, OPQRST) plus pertinent positives and negatives that rule in the target path and rule out sibling branches in the tree.
    \item \emph{PMH, Meds, PSH, SocHx/FamHx, \& ROS:} Every condition must have a duration and status; every active medication must link to a documented diagnosis; procedures must have a documented cause; habits must be quantified (e.g., ``15 pack-year tobacco use'').
  \end{itemize}
  \item \textbf{Physical Exam \& Diagnostic Data:} Mandate exact vital signs (BP, HR, RR, Temp, $\text{SpO}_2$; e.g., ``Temperature $39.4^\circ\text{C}$'' instead of ``high fever''), systematic physical exam findings, and concrete laboratory/imaging results with reference ranges.
  \item \textbf{Strict Groundedness, Path Enforcement, \& Context:} Keep all clinical findings strictly grounded in the source guideline, unequivocally forcing every decision node along \texttt{<scenario\_path>} without diverting to an alternative branch or referencing regional guideline bodies (e.g., ``NICE'').
\end{enumerate}
\end{tcolorbox}

\vspace{1.5pt}
\begin{tcolorbox}[colback=white, colframe=black!25, boxrule=0.4pt, arc=1.5pt, left=5.5pt, right=5.5pt, top=3pt, bottom=3pt]
{\raggedright\noindent\textbf{2. Structured Element Scenario Generation Prompts (Differential Diagnosis, Diagnostic Investigations, and Complications)}\par}\vspace{1.5pt}
\textbf{Role (\texttt{<role>}):} You are an expert clinical scenario designer and medical simulation writer creating realistic patient case narratives focused on a specific diagnostic investigation, differential diagnosis, or medical complication.\\[1.5pt]
\textbf{Inputs:} Category context (\texttt{<differentials\_context>}, \texttt{<investigations\_context>}, or \texttt{<complications\_context>}), target item (\texttt{<focus\_differential>}, \texttt{<focus\_investigation>}, or \texttt{<focus\_complication>}), and \texttt{<schema>}.\\[1.5pt]
\textbf{Instructions (\texttt{<instructions>}):}
\begin{enumerate}[leftmargin=1.4em, topsep=0.5pt, itemsep=0.5pt, parsep=0pt]
  \item \textbf{Clinical Presentation Diversity:} Generate 2--4 diverse patient scenarios in \texttt{EvalScenarioSO} format spanning distinct clinical trajectories:
  \begin{itemize}[leftmargin=1.2em, topsep=0pt, itemsep=0pt]
    \item \emph{Differential Diagnosis:} (a) \emph{``Appear Similar, Land Different''} (initially resembles the focus differential but workup establishes the primary diagnosis); (b) \emph{``Appear Dissimilar, Land Same''} (atypical presentation that ultimately confirms the focus condition); (c) \emph{``Direct Contradiction / Borderline''} (hinges on a subtle distinguishing sign or test threshold).
    \item \emph{Diagnostic Investigations:} (a) \emph{Standard Presentation} (textbook indication); (b) \emph{Atypical Presentation} (nuanced age group or comorbidities making the test critical); (c) \emph{``The Teaser''} (mimics another condition where subtle clues make \texttt{<focus\_investigation>} pivotal); (d) \emph{Borderline / High-Acuity} (test result determines immediate intervention vs.\ watchful waiting).
    \item \emph{Complications:} (a) \emph{Sudden / Dramatic Onset} (rapid decompensation requiring immediate recognition); (b) \emph{Insidious / Subtle Onset} (early warning signs before full manifestation); (c) \emph{High-Risk / Atypical Context} (immunocompromised or post-procedural presentation).
  \end{itemize}
  \item \textbf{Scenario Focus, Differentiator, \& Groundedness:} Define a clear \texttt{scenario\_focus} and \texttt{clinical\_differentiator} for each case. Keep all clinical findings, differentiating criteria, and expected test findings strictly grounded in the source monograph without leaking the target answer in the patient presentation.
\end{enumerate}
\end{tcolorbox}
\end{tcolorbox}
\caption{Condensed prompt templates used for core clinical scenario generation across treatment decision-tree pathways and structured clinical elements (differential diagnoses, investigations, and complications).}
\label{fig:lst_generation_prompts}
\end{figure}

\begin{figure}[ht!]
\centering
\begin{tcolorbox}[
  colback=gray!3,
  colframe=black!70,
  colbacktitle=black!75,
  coltitle=white,
  boxrule=0.6pt,
  arc=2pt,
  left=6pt, right=6pt, top=4pt, bottom=4pt,
  toptitle=2.5pt, bottomtitle=2.5pt,
  fonttitle=\footnotesize\bfseries\raggedright,
  title={Prompts for Multi-Stage Faithfulness Validation, Critique-Refinement, and Rubric Generation}
]
\fontsize{8.2pt}{9.8pt}\selectfont\RaggedRight
\setlength{\parskip}{2pt}

\begin{tcolorbox}[colback=white, colframe=black!25, boxrule=0.4pt, arc=1.5pt, left=5.5pt, right=5.5pt, top=3pt, bottom=3pt]
{\raggedright\noindent\textbf{1. Two-Pass Faithfulness Validation (Pass 1 Against Extracted Structured Data; Pass 2 Against Source Monograph XML)}\par}\vspace{1.5pt}
\textbf{Role (\texttt{<role>}):} You are an expert clinical scenario auditor and medical guideline validator. Validate \texttt{<generated\_scenario>} (Pass 1) against the extracted clinical data (\texttt{<extracted\_data>}) and (Pass 2) directly against the original guideline XML (\texttt{<guideline\_xml>}).\\[1.5pt]
\textbf{Instructions (\texttt{<instructions>}):} Identify any issues where \texttt{<generated\_scenario>} diverges from, contradicts, or fails to represent the source logic:
\begin{enumerate}[leftmargin=1.4em, topsep=0.5pt, itemsep=0.5pt, parsep=0pt]
  \item \textbf{Faithfulness \& Ground Truth Consistency:} Verify that every decision point in \texttt{treatment\_path} reflects the corresponding node in the extracted tree and that the original XML guideline supports \texttt{ground\_truth\_treatment\_plan} (\texttt{mx\_plan}) for a patient with these specific findings.
  \item \textbf{Missing Constraints \& Implicit Conflicts:} Check whether any critical indications or contraindications in the XML are ignored, or whether confounding details in the presentation would lead a clinician down a different branch or make the ground-truth plan unsafe.
  \item \textbf{Allowed Plausible Details vs.\ Hallucinated Logic:} Realistic patient details (demographics, concrete lab values, non-conflicting vitals) are expected and permitted; however, \texttt{mx\_plan} MUST be strictly derived from the source guideline and must never invent external clinical logic.
  \item \textbf{Completeness \& Structured Output:} Verify that the scenario includes sufficient detail to unambiguously force every decision node. For each issue, output \texttt{severity} (\texttt{"ERROR"} or \texttt{"WARNING"}), \texttt{category} (\texttt{"faithfulness"}, \texttt{"hallucination"}, \texttt{"consistency"}, \texttt{"completeness"}, or \texttt{"simplification"}), \texttt{message}, \texttt{path}, and \texttt{suggested\_fix}. If no issues exist, return \texttt{is\_valid: true}.
\end{enumerate}
\end{tcolorbox}

\vspace{1.5pt}
\begin{tcolorbox}[colback=white, colframe=black!25, boxrule=0.4pt, arc=1.5pt, left=5.5pt, right=5.5pt, top=3pt, bottom=3pt]
{\raggedright\noindent\textbf{2. Clinical Coherence Critique \& Refinement (Treatment Pathways and Structured Elements)}\par}\vspace{1.5pt}
\textbf{Role (\texttt{<role>}):} You are an expert clinical scenario auditor, medical simulator, and specialist clinician. Review \texttt{<generated\_scenario>} against \texttt{<treatment\_tree>} and \texttt{<scenario\_path>}, as well as for internal medical consistency and clinical completeness.\\[1.5pt]
\textbf{Instructions (\texttt{<instructions>}):}
\begin{enumerate}[leftmargin=1.4em, topsep=0.5pt, itemsep=0.5pt, parsep=0pt]
  \item \textbf{Analyze for Tree Alignment:} Flag any \emph{Missing Information} (signs, symptoms, or test results required by a path node), \emph{Ambiguity} (vague findings that fail to map cleanly to decision criteria), or \emph{Contradictions} (details pointing to a different branch).
  \item \textbf{Analyze Against the Clinical Specification Checklist:}
  \begin{itemize}[leftmargin=1.2em, topsep=0pt, itemsep=0pt]
    \item \emph{Diagnostic Mapping \& Procedural Logic:} Every active medication must map to a documented diagnosis (e.g., Lisinopril $\rightarrow$ Hypertension); every past surgery must have an associated underlying condition.
    \item \emph{Chronicity, Severity, \& Quantification:} Specify duration, control status, and latest labs (e.g., $\text{HbA}_{1\text{c}}$, $\text{FEV}_1$) for chronic conditions; replace vague descriptors with exact objective values (e.g., ``$39.2^\circ\text{C}$'', ``30 pack-year history'').
    \item \emph{Pertinent Negatives \& Recommendation Bias:} Include critical rule-out symptoms and strip any regional guideline identifiers (e.g., ``NICE'', ``BNF'', NG/CG codes).
  \end{itemize}
  \item \textbf{Refinement Pass:} Rewrite \texttt{<generated\_scenario>} into \texttt{refined\_scenario} based on \texttt{<critique>} to resolve all missing links, quantify subjective findings, and enforce strict decision-tree alignment.
\end{enumerate}
\end{tcolorbox}

\vspace{1.5pt}
\begin{tcolorbox}[colback=white, colframe=black!25, boxrule=0.4pt, arc=1.5pt, left=5.5pt, right=5.5pt, top=3pt, bottom=3pt]
{\raggedright\noindent\textbf{3. Auxiliary Scenario Rubric Generation (\texttt{AtsRubricCriterion})}\par}\vspace{1.5pt}
\textbf{Instructions (\texttt{<instructions>}):} Review \texttt{<scenario\_path>}, \texttt{<treatment\_tree>}, and \texttt{<generated\_scenario>} (or \texttt{<focus\_element>} for structured-element cases). Identify the core management recommendations and create atomic, objective, verifiable (\texttt{Presence/Absence}) \texttt{AtsRubricCriterion} items stored in metadata checking whether the clinician (a) selected the appropriate first-line treatment, (b) ordered necessary diagnostic investigations, and (c) addressed critical red flags and follow-up requirements.
\end{tcolorbox}
\end{tcolorbox}
\caption{Condensed prompt templates used for two-pass faithfulness verification, clinical coherence critique and refinement, and auxiliary scenario-specific rubric generation.}
\label{fig:lst_verification_prompts}
\end{figure}

\begin{figure}[ht!]
\centering
\begin{tcolorbox}[
  colback=gray!3,
  colframe=black!70,
  colbacktitle=black!75,
  coltitle=white,
  boxrule=0.6pt,
  arc=2pt,
  left=6pt, right=6pt, top=4pt, bottom=4pt,
  toptitle=2.5pt, bottomtitle=2.5pt,
  fonttitle=\footnotesize\bfseries\raggedright,
  title={Prompts for Targeted Temporal Freshness and Regional Specificity Difference Extraction, Verification, and Scenario Generation}
]
\fontsize{8.2pt}{9.8pt}\selectfont\RaggedRight
\setlength{\parskip}{2pt}

\begin{tcolorbox}[colback=white, colframe=black!25, boxrule=0.4pt, arc=1.5pt, left=5.5pt, right=5.5pt, top=3pt, bottom=3pt]
{\raggedright\noindent\textbf{1. Paired Guideline Diff Extraction Prompts (\texttt{FreshnessCandidateDiffList} \& \texttt{RegionalCandidateDiffList})}\par}\vspace{1.5pt}
\textbf{Role:} You are an expert clinical guideline auditor comparing paired editions of the \emph{BMJ Best Practice} guideline for \texttt{\{topic\_title\}}: either the OLDER (\texttt{\{oldest\_date\}}) vs.\ CURRENT (\texttt{\{newest\_date\}}) US (\texttt{en-us}) editions (with unified diff \texttt{<normalized\_clinical\_diff>} and full cleaned XMLs), or the US (\texttt{en-us}) vs.\ International (\texttt{en-gb}) editions (\texttt{<guideline\_us>} and \texttt{<guideline\_row>}).\\[1.5pt]
\textbf{Strict Extraction Rules:}
\begin{enumerate}[leftmargin=1.4em, topsep=0.5pt, itemsep=0.5pt, parsep=0pt]
  \item \textbf{Include Substantive Clinical Divergences Only:} Extract changes/differences in first-line or second-line medications or regimens, surgical vs.\ conservative management, diagnostic workup or screening criteria, risk calculators or numeric/age thresholds for initiating therapy, maximum allowed dosages, or contraindications and safety monitoring.
  \item \textbf{Strictly Exclude Superficial Differences:} Ignore wording/phrasing edits that do not alter clinical action, table/formatting layout changes, background epidemiology statistics, citation renumbering, generic/brand drug name synonyms for the same chemical entity (e.g., acetaminophen vs.\ paracetamol, albuterol vs.\ salbutamol, epinephrine vs.\ adrenaline), spelling variations (e.g., appendectomy vs.\ appendicectomy), and unit formatting ($\text{mg/dL}$ vs.\ $\text{mmol/L}$ for identical physiological thresholds).
  \item \textbf{Required Output Fields:} For each candidate diff, extract \texttt{diff\_title}, \texttt{category}, \texttt{clinical\_context}, paired recommendations (\texttt{previous\_us\_recommendation} vs.\ \texttt{current\_us\_recommendation}, or \texttt{us\_recommendation} vs.\ \texttt{row\_recommendation} with \texttt{row\_applicable\_countries} and cited authorities), exact verbatim XML quotes from both editions, and \texttt{evidence\_or\_rationale}.
\end{enumerate}
\end{tcolorbox}

\vspace{1.5pt}
\begin{tcolorbox}[colback=white, colframe=black!25, boxrule=0.4pt, arc=1.5pt, left=5.5pt, right=5.5pt, top=3pt, bottom=3pt]
{\raggedright\noindent\textbf{2. Independent Auditor Diff Verification Prompts (\texttt{FreshnessVerificationResult} \& \texttt{RegionalVerificationResult})}\par}\vspace{1.5pt}
\textbf{Role:} You are an independent Senior Clinical Guideline Auditor. Rigorously verify each candidate clinical difference against the unified clinical diff and external Google Search context (\texttt{<search\_results>}) for Temporal Freshness, or against both full cleaned XMLs (\texttt{<guideline\_us>}, \texttt{<guideline\_row>}) for Regional Specificity.\\[1.5pt]
\textbf{Verification Rules (Retain Only if \texttt{is\_truly\_different\_in\_xml = true}, \texttt{is\_clinically\_meaningful = true}, and \texttt{confidence\_score >= 4}):}
\begin{enumerate}[leftmargin=1.4em, topsep=0.5pt, itemsep=0.5pt, parsep=0pt]
  \item \texttt{is\_truly\_different\_in\_xml}: Confirm against both editions that they genuinely differ on this recommendation. Reject (\texttt{false}) if another section of the XML already states the same recommendation or if both options are equally offered in both editions.
  \item \texttt{is\_clinically\_meaningful}: Confirm that a clinician following one edition vs.\ the other would make a different, concrete clinical decision (ordering a different test, using a different risk calculator/threshold, prescribing a different drug or dose, avoiding a newly contraindicated therapy, or changing disposition). Reject (\texttt{false}) if editorial or synonymous.
  \item \texttt{confidence\_score} \& Suite-Specific Verification Fields: Rate clinical impact and evidentiary clarity from 1 (lowest) to 5 (unequivocal, high-impact clinical difference) with a detailed \texttt{verification\_rationale}, plus \texttt{is\_supported\_by\_clinical\_evidence} (Temporal Freshness) or \texttt{refined\_row\_countries} (Regional Specificity).
\end{enumerate}
\end{tcolorbox}

\vspace{1.5pt}
\begin{tcolorbox}[colback=white, colframe=black!25, boxrule=0.4pt, arc=1.5pt, left=5.5pt, right=5.5pt, top=3pt, bottom=3pt]
{\raggedright\noindent\textbf{3. Targeted Scenario Generation Prompts (\texttt{TargetedScenarioBlueprint} \& \texttt{RegionalPairedScenarioBlueprint})}\par}\vspace{1.5pt}
\textbf{Role:} You are an expert clinical scenario designer and medical board examiner creating a realistic evaluation scenario for \texttt{\{topic\_title\}} targeting a verified temporal update or regional difference (\texttt{clinical\_context}).\\[1.5pt]
\textbf{Scenario Design Requirements:}
\begin{enumerate}[leftmargin=1.4em, topsep=0.5pt, itemsep=0.5pt, parsep=0pt]
  \item \textbf{Patient Presentation (\texttt{patient} / \texttt{patient\_base}, \texttt{reason\_for\_visit}, \texttt{patient\_actor}):} Design a clinically rich case (concrete vitals, physical exam findings, lab/imaging values, medical history, and medications) that squarely places the patient in \texttt{clinical\_context} so that the difference between the two guideline editions is \textbf{pivotal} to the management plan (e.g., placing calculated risk or lab values directly inside the window between the old and new thresholds, or specifying the exact comorbidity/intolerance that triggers the divergent drug choice). For regional scenarios, specify matched clinical settings in \texttt{us\_patient\_location} (United States) and \texttt{row\_patient\_location} (e.g., United Kingdom). Do NOT leak the recommendation or mention ``guideline updates'' in the patient presentation.
  \item \textbf{Paired Ground Truths (\texttt{ground\_truth} / \texttt{us\_ground\_truth} \& \texttt{row\_ground\_truth}):} Each ground truth (\texttt{EvalScenarioVisitGoldTruth}) MUST include \textbf{both}: (A) \emph{Comprehensive overall standard-of-care management} (diagnosis, workup, core treatments, follow-up, and red flags) grounded in the respective guideline XML, AND (B) \emph{Explicit specificity for the targeted difference} in \texttt{appropriate\_recommendations\_or\_actions}, \texttt{important\_items}, and \texttt{guideline\_justification}, stating both the required recommendation for that edition and the superseded/divergent alternative that must be avoided.
\end{enumerate}
\end{tcolorbox}
\end{tcolorbox}
\caption{Condensed prompt templates used for targeted temporal freshness and regional specificity difference extraction, independent auditor verification, and paired targeted scenario generation.}
\label{fig:lst_targeted_diff_prompts}
\end{figure}

\clearpage
\section{Inference Harnesses and Context Configurations}
\label{appendix:inference_harnesses}

We evaluate seven checkpoints across the Gemini 3 model family: two Flash Lite models (Gemini 3.1 Flash Lite and Gemini 3.5 Flash Lite) and five Flash models (Gemini 3 Flash, Gemini 3.5 Flash, Gemini 3.6 Flash, Gemini 3.7 Flash, and Gemini 3.8 Flash), with five independent runs per scenario across the eight core benchmark configurations and the four targeted suite configurations. In every configuration, the model receives the structured patient scenario (with ground-truth fields withheld and the target practice location and country explicitly included for Regional Specificity scenarios) and generates a structured response containing differential diagnoses and a management plan (\cref{fig:lst_inference_prompts,fig:lst_inference_prompts_rag}):
\begin{enumerate}[leftmargin=1.5em, itemsep=2pt]
    \item \textbf{Baseline (No Harness):} Single-pass generation directly from the patient scenario using parametric knowledge alone (\cref{fig:lst_inference_prompts}).
    \item \textbf{With Search:} Two-stage open-web search harness (labeled \emph{+Web Search} in the targeted suites; \cref{fig:lst_inference_prompts}). In Stage 1, the model generates search queries (\texttt{SearchQueries}) from the scenario in a single batch, executed via Google Search (top two results per query). In Stage 2, the model generates its recommendations conditioned on the scenario and retrieved snippets.
    \item \textbf{Synthesize-4:} Inference-time synthesis in which $N = 4$ candidate responses are sampled in parallel and aggregated by a synthesis call that is tasked with combining the best aspects of the candidate answers while eliminating flaws (\cref{fig:lst_inference_prompts}).
    \item \textbf{Search + Synthesize-4} (abbreviated \emph{Search + Synth-4}): Generates search queries and retrieves a shared batch of Google Search snippets as in \emph{With Search}, then samples $N = 4$ parallel candidates over the search-augmented prompt followed by synthesis (also used with Gemini 3 Flash to generate the AI recommendations in the physician study; \cref{fig:lst_inference_prompts}).
    \item \textbf{Dynamic Guideline:} Closed-corpus two-stage retrieval over the 354 US \emph{BMJ Best Practice} clinical monographs (\cref{fig:lst_inference_prompts_rag}). In Stage 1, given the scenario and the catalog of 354 monograph IDs and titles, the model selects up to $K = 5$ relevant guidelines (with $K \in \{1, 3, 5, 7, 9\}$ evaluated in \cref{appendix:dynamic_retrieval}). In Stage 2, the cleaned XML text of the selected monographs is loaded into context for generation.
    \item \textbf{Dynamic Excerpt:} Closed-corpus condensed retrieval in which the model selects up to $K = 5$ relevant guidelines from the 354-title catalog and specifies which extracted clinical sections (\texttt{treatment\_decision\_tree}, \texttt{differentials}, \texttt{investigations}, or \texttt{complications}) are needed; only those structured JSON extractions are loaded into context (\cref{fig:lst_inference_prompts_rag}).
    \item \textbf{Oracle Guideline:} In-context conditioning on the complete cleaned XML of the scenario's source monograph (\cref{fig:lst_inference_prompts_rag}). In the targeted evaluations, this harness provides either the revised or regionally matched edition (\emph{+Updated/Correct Guideline}) or the superseded or conflicting regional edition (\emph{+Outdated/Incorrect Guideline}).
    \item \textbf{Oracle Excerpt:} In-context conditioning on only the source monograph's extracted structured JSON section corresponding to the scenario's clinical category (\texttt{decision\_tree}, \texttt{differentials}, \texttt{investigations}, or \texttt{complications}; \cref{fig:lst_inference_prompts_rag}).
\end{enumerate}

\begin{figure}[ht!]
\centering
\begin{tcolorbox}[
  colback=gray!3,
  colframe=black!70,
  colbacktitle=black!75,
  coltitle=white,
  boxrule=0.6pt,
  arc=2pt,
  left=7pt, right=7pt, top=5pt, bottom=5pt,
  toptitle=3pt, bottomtitle=3pt,
  fonttitle=\footnotesize\bfseries\raggedright,
  title={Inference Harness Prompt Templates: Part 1 (Baseline, Web Search, and Multi-Candidate Synthesis)}
]
\footnotesize\RaggedRight
\setlength{\parskip}{3pt}

\begin{tcolorbox}[colback=white, colframe=black!25, boxrule=0.4pt, arc=1.5pt, left=6pt, right=6pt, top=4pt, bottom=4pt]
\noindent\textbf{1. Baseline (No Harness)}\\
\textbf{Output Schema:} \texttt{ClinicalRecommendations(ddx: DDx, mx\_plan: MxPlan)}\\[2pt]
You are an expert clinician. Given the following scenario, please provide your differential diagnosis (DDx) and management plan (MxPlan).\\[2pt]
Scenario:\\
\texttt{\{scenario\_json\}}
\end{tcolorbox}

\vspace{2pt}
\begin{tcolorbox}[colback=white, colframe=black!25, boxrule=0.4pt, arc=1.5pt, left=6pt, right=6pt, top=4pt, bottom=4pt]
\noindent\textbf{2. With Search (Two-Stage Web Search Harness)}\\[2pt]
\textbf{Stage 1: Query Generation} --- \textbf{Output Schema:} \texttt{SearchQueries(queries: list[str])}\\[1.5pt]
You are a clinical expert. Given the following scenario, please generate a list of search queries that would help you confirm the diagnosis or determine the best management plan.\\[1.5pt]
Scenario:\\
\texttt{\{scenario\_json\}}\\[4pt]
\textbf{Stage 2: Search-Augmented Clinical Recommendation}\\
\textbf{Output Schema:} \texttt{ClinicalRecommendations(ddx, mx\_plan)}\\[1.5pt]
You are an expert clinician. Given the following scenario and search results, please provide your differential diagnosis (DDx) and management plan (MxPlan).\\[1.5pt]
Scenario: \texttt{\{scenario\_json\}}\\
Search Results: \texttt{\{search\_results\_str\}}
\end{tcolorbox}

\vspace{2pt}
\begin{tcolorbox}[colback=white, colframe=black!25, boxrule=0.4pt, arc=1.5pt, left=6pt, right=6pt, top=4pt, bottom=4pt]
\noindent\textbf{3. Synthesize-4 \& Search + Synthesize-4 (Synthesis over $N = 4$ Parallel Candidate Responses)}\\
\textbf{Output Schema:} \texttt{ClinicalRecommendations}\\[2pt]
You are given a question (enclosed in \texttt{<user\_inquiry> </user\_inquiry>}) and different possible answers (enclosed in \texttt{<candidate\_answer\_fingerprints> </candidate\_answer\_fingerprints>}). You need to synthesize a response to the question, leveraging the best aspects of the candidate answers and eliminating any flaws.\\[2pt]
\texttt{<user\_inquiry>}\\
\texttt{\{prompt\}}\\
\texttt{</user\_inquiry>}\\[2pt]
\texttt{<candidate\_answer\_fingerprints>}\\
\texttt{\{results\}}\\
\texttt{</candidate\_answer\_fingerprints>}
\end{tcolorbox}
\end{tcolorbox}
\caption{Prompt templates for the Baseline (No Harness), With Search, and Synthesize-4 / Search + Synthesize-4 inference harnesses. Structured JSON outputs matching the indicated schemas are enforced via constrained decoding (auto-generated JSON schema suffixes omitted for brevity).}
\label{fig:lst_inference_prompts}
\end{figure}

\begin{figure}[ht!]
\centering
\begin{tcolorbox}[
  colback=gray!3,
  colframe=black!70,
  colbacktitle=black!75,
  coltitle=white,
  boxrule=0.6pt,
  arc=2pt,
  left=7pt, right=7pt, top=3pt, bottom=3pt,
  toptitle=3pt, bottomtitle=3pt,
  fonttitle=\footnotesize\bfseries\raggedright,
  title={Inference Harness Prompt Templates: Part 2 (Closed-Corpus Dynamic Retrieval and Oracle Harnesses)}
]
\fontsize{8.1pt}{9.6pt}\selectfont\RaggedRight
\setlength{\parskip}{1.5pt}

\begin{tcolorbox}[colback=white, colframe=black!25, boxrule=0.4pt, arc=1.5pt, left=6pt, right=6pt, top=2.5pt, bottom=2.5pt]
\noindent\textbf{4. Dynamic Guideline (Two-Stage Closed-Corpus Full-XML Retrieval)}\\[2pt]
\textbf{Stage 1: Guideline Catalog Selection ($K = 5$)}\\
\textbf{Output Schema:} \texttt{GuidelineSelection(guideline\_ids: list[str], justification: str)}\\[1.5pt]
You are an expert clinician. Given the following patient scenario, select up to \texttt{\{top\_k\}} of the most relevant clinical guideline topics from the list of available topics (ordered from most to least relevant). You may select anywhere from 1 up to \texttt{\{top\_k\}} guidelines based strictly on clinical relevance --- you do not need to select \texttt{\{top\_k\}} if fewer guidelines are sufficient. Provide the Guideline IDs of the selected topics.\\
Scenario: \texttt{\{scenario\_json\}} \quad Available Topics: \texttt{\{topics\_list\}}\\[4pt]
\textbf{Stage 2: Guideline-Conditioned Generation} --- \textbf{Output Schema:} \texttt{ClinicalRecommendations(ddx, mx\_plan)}\\[1.5pt]
You are an expert clinician. Given the following scenario and the relevant clinical guideline XML content (which contains up to \texttt{\{top\_k\}} retrieved guidelines based on clinical relevance), please provide your differential diagnosis (DDx) and management plan (MxPlan). You may consult any or all of the provided guidelines as needed for your plan.\\
Guideline XML(s): \texttt{\{xml\_content\}} \quad Scenario: \texttt{\{scenario\_json\}}
\end{tcolorbox}

\vspace{1pt}
\begin{tcolorbox}[colback=white, colframe=black!25, boxrule=0.4pt, arc=1.5pt, left=6pt, right=6pt, top=2.5pt, bottom=2.5pt]
\noindent\textbf{5. Dynamic Excerpt (Two-Stage Closed-Corpus Condensed Section Retrieval)}\\[2pt]
\textbf{Stage 1: Guideline \& Section Selection ($K = 5$)}\\
\textbf{Output Schema:} \texttt{CondensedGuidelineSelection(selections: list[CondensedGuidelineSectionRequest], justification: str)}\\[1.5pt]
You are an expert clinician. Given the following patient scenario, select up to \texttt{\{top\_k\}} of the most relevant clinical guideline topics from the list of available topics (ordered from most to least relevant), and for each selected guideline specify which condensed section(s) to retrieve. You may select anywhere from 1 up to \texttt{\{top\_k\}} guidelines based strictly on clinical relevance --- you do not need to select \texttt{\{top\_k\}} if fewer guidelines are sufficient.\\
For each selected guideline, choose one or more of the following condensed sections based on clinical relevance to the scenario:
\begin{itemize}[leftmargin=1.4em, topsep=1pt, itemsep=0.5pt, parsep=0pt]
  \item \texttt{'treatment\_decision\_tree'}: Structured treatment and management decision tree (first-line/second-line therapies, branching clinical criteria, management steps).
  \item \texttt{'differentials'}: Structured differential diagnoses (competing conditions, differentiating signs/symptoms, and differentiating tests).
  \item \texttt{'investigations'}: Structured diagnostic investigations (first-line and targeted laboratory, imaging, and diagnostic tests and expected findings).
  \item \texttt{'complications'}: Structured clinical complications (short- and long-term complications, likelihood, timeframe, and clinical descriptions).
\end{itemize}
Scenario: \texttt{\{scenario\_json\}} \quad Available Topics: \texttt{\{topics\_list\}}\\[4pt]
\textbf{Stage 2: Condensed-Excerpt Generation} --- \textbf{Output Schema:} \texttt{ClinicalRecommendations(ddx, mx\_plan)}\\[1.5pt]
You are an expert clinician. Given the following scenario and the retrieved condensed clinical guideline section(s) (structured summaries / decision trees from up to \texttt{\{top\_k\}} relevant guidelines), please provide your differential diagnosis (DDx) and management plan (MxPlan). You may consult any or all of the provided condensed guideline sections as needed for your plan.\\
Condensed Guideline Section(s): \texttt{\{condensed\_content\}} \quad Scenario: \texttt{\{scenario\_json\}}
\end{tcolorbox}

\vspace{1pt}
\begin{tcolorbox}[colback=white, colframe=black!25, boxrule=0.4pt, arc=1.5pt, left=6pt, right=6pt, top=2.5pt, bottom=2.5pt]
\noindent\textbf{6. Oracle Guideline (Full Cleaned Monograph XML) \& Oracle Excerpt (Structured Category JSON)}\\
\textbf{Output Schema:} \texttt{ClinicalRecommendations}\\[2pt]
\textbf{Oracle Guideline Prompt:}\\
You are an expert clinician. Given the following scenario and the relevant clinical guideline XML, please provide your differential diagnosis (DDx) and management plan (MxPlan).\\
Guideline XML: \texttt{\{xml\_content\}} \quad Scenario: \texttt{\{scenario\_json\}}\\[4pt]
\textbf{Oracle Excerpt Prompt:}\\
You are an expert clinician. Given the following scenario and the relevant condensed clinical guideline section (structured summary / decision tree), please provide your differential diagnosis (DDx) and management plan (MxPlan).\\
Condensed Guideline Section: \texttt{\{condensed\_content\}} \quad Scenario: \texttt{\{scenario\_json\}}
\end{tcolorbox}
\end{tcolorbox}
\caption{Prompt templates for the closed-corpus Dynamic Guideline, Dynamic Excerpt, Oracle Guideline, and Oracle Excerpt inference harnesses. Structured JSON outputs matching the indicated schemas are enforced via constrained decoding (auto-generated JSON schema suffixes omitted for brevity).}
\label{fig:lst_inference_prompts_rag}
\end{figure}

\clearpage
\section{Clinician-Calibrated Autorater Rubric, Calibration, and Consistency}
\label{appendix:autorater_calibration}

We evaluate open-ended clinical responses from both models and participating physicians using a clinician-calibrated LLM autorater (Gemini 3.8 Flash) on the four-point ordinal clinical appropriateness scale defined in \cref{tab:likert_rubric} (prompt in \cref{fig:lst_autorater_prompt}):
\begin{itemize}[leftmargin=1.5em, topsep=1pt, itemsep=0pt, parsep=0pt]
    \item \textbf{Score 4 (Adherent / Acceptable):} Matches the evidence-based practice recommendation, or an alternative but clinically appropriate choice (e.g., an equivalent drug from the same class, a valid alternative workup, or an appropriate specialist handoff) without compromising safety.
    \item \textbf{Score 3 (Minor Deviation):} Suboptimal but safe, with a sound core diagnostic and therapeutic trajectory (e.g., omitting routine monitoring or baseline safety screening, initiating non-emergent systemic therapy before confirmatory results, or choosing a less preferred secondary option).
    \item \textbf{Score 2 (Moderate Deviation):} Clinically significant omission or suboptimal choice without immediate risk of severe harm (e.g., omitting first-line therapy or an essential upfront diagnostic study, failing to stop an offending medication, or unwarranted inpatient escalation).
    \item \textbf{Score 1 (Major Deviation):} Contraindicated treatment, unsafe dose, or missed life-threatening red flag posing direct risk of severe patient harm (e.g., failing to triage a life- or organ-threatening emergency presentation or advising against mandatory urgent intervention).
\end{itemize}
\vspace{-2pt}
On the core benchmark and physician evaluations, \emph{guideline adherence} is defined as the proportion of responses rated Score 4 (\emph{Adherent / Acceptable}), reported alongside the full four-tier distribution; for the targeted temporal and regional suites, the same autorater prompt is appended with the verified divergence and targeted checklist to assess adherence specifically to the updated or regionally matched recommendation. Before assigning a score, the autorater reconciles key actions in the candidate plan against the structured reference ground truth, summarizes any substantive omissions or discrepancies in a brief clinical rationale, and applies the four appropriateness tiers sequentially from Major Deviation to Adherent / Acceptable (Stages 1 to 4 in \cref{fig:lst_autorater_prompt}).

We calibrated the autorater on the 200-scenario evaluation set from the Primary Care Physician (PCP) study (50 scenarios per clinical category; \cref{appendix:pcp_study_protocol}). After submitting both their unassisted and AI-assisted plans, participating physicians reviewed the scenario ground truth alongside the initial uncalibrated autorater rating of each standalone AI management plan and either confirmed that rating or assigned a corrected score. Calibrating the autorater prompt against these physician ratings, using generic clinical severity hierarchies, primary-care and specialist handoff principles, and 10 synthetic calibration exemplars (\cref{fig:lst_autorater_prompt}), achieved $89.0\%$ ($178/200$) exact four-tier agreement and $96.0\%$ ($192/200$) adjacent-tier agreement across the 200 scenarios using the five-run median autorater rating (\cref{tab:supp_autorater_confusion}), with per-category recall exceeding $60\%$ across all four tiers ($94.9\%$ Adherent, $61.5\%$ Minor Deviation, $73.3\%$ Moderate Deviation, and $100.0\%$ [$1/1$] Major Deviation).

\vspace{-2pt}
\noindent\begin{minipage}{\textwidth}
\centering
\captionof{table}{Confusion matrix comparing the clinician-calibrated autorater (five-run median) against clinician ratings of AI management plans ($N = 200$ scenarios). Exact 4-tier accuracy $= 89.0\%$ ($178/200$); adjacent-tier agreement ($|\Delta \text{tier}| \le 1$) $= 96.0\%$ ($192/200$).}
\label{tab:supp_autorater_confusion}
\vspace{-4pt}
\footnotesize
\setlength{\tabcolsep}{5pt}
\renewcommand{\arraystretch}{1.0}
\begin{tabularx}{\textwidth}{>{\RaggedRight\arraybackslash}Xrrrrrr}
\toprule
 & \multicolumn{4}{c}{\textbf{Human Clinician Rating}} & & \\
\cmidrule(lr){2-5}
\textbf{Autorater Assigned Rating} & \textbf{Adherent (4)} & \textbf{Minor (3)} & \textbf{Moderate (2)} & \textbf{Major (1)} & \textbf{Row Total} & \textbf{Precision} \\
\midrule
Adherent / Acceptable (4) & 150 & 10 & 3 & 0 & 163 & 92.0\% \\
Minor Deviation (3) & 3 & 16 & 1 & 0 & 20 & 80.0\% \\
Moderate Deviation (2) & 5 & 0 & 11 & 0 & 16 & 68.8\% \\
Major Deviation (1) & 0 & 0 & 0 & 1 & 1 & 100.0\% \\
\midrule
Column Total & 158 & 26 & 15 & 1 & 200 & -- \\
Category Recall & 94.9\% & 61.5\% & 73.3\% & 100.0\% & -- & 89.0\% \\
\bottomrule
\end{tabularx}
\end{minipage}

\clearpage
\subsection{Clinician-Calibrated Autorater System Prompt}
\label{appendix:autorater_prompt_verbatim}

\noindent The clinician-calibrated autorater prompt is presented across six sequential modular boxes (\cref{fig:lst_autorater_prompt}), concatenated in order at inference time.

\definecolor{PromptSlate}{HTML}{1E293B}
\definecolor{PromptSlateSub}{HTML}{334155}
\definecolor{PromptSlateBg}{HTML}{F8FAFC}
\definecolor{PromptMajor}{HTML}{B91C1C}
\definecolor{PromptMajorBg}{HTML}{FEF2F2}
\definecolor{PromptMod}{HTML}{C2410C}
\definecolor{PromptModBg}{HTML}{FFF7ED}
\definecolor{PromptMinor}{HTML}{B45309}
\definecolor{PromptMinorBg}{HTML}{FFFBEB}
\definecolor{PromptAdh}{HTML}{047857}
\definecolor{PromptAdhBg}{HTML}{ECFDF5}

\newcommand{\PromptTitleFont}{\fontsize{7.3pt}{8.4pt}\selectfont\bfseries\raggedright}
\newcommand{\PromptBodyFont}{\fontsize{6.85pt}{7.8pt}\selectfont\raggedright}
\newcommand{\PromptSep}{\par\vspace{2.0pt}\noindent{\color{black!18}\rule{\linewidth}{0.4pt}}\par\vspace{2.0pt}}

\begin{tcolorbox}[
  colback=PromptSlateBg,
  colframe=PromptSlate,
  colbacktitle=PromptSlate,
  coltitle=white,
  boxrule=0.65pt,
  arc=2.5pt,
  left=6pt, right=6pt, top=3.5pt, bottom=3.5pt,
  toptitle=2.2pt, bottomtitle=2.2pt,
  before skip=4pt, after skip=4pt,
  fonttitle=\PromptTitleFont,
  title={Module 1 of 6 --- Role, Factual Reconciliation Protocol, and Specialist Handoff Principle}
]
\PromptBodyFont
\setlength{\parskip}{1.6pt}
\noindent You are an Expert Clinical Medical Reviewer and Guideline Auditor.\par
\noindent Your task is to evaluate an AI Clinician's proposed Clinical Assessment and Management Plan against a Reference Evidence-Based Clinical Guideline Ground Truth for a specific patient scenario.\par
\noindent Your evaluation must calibrate to how experienced human Primary Care Physicians (PCPs) evaluate clinical adherence and patient safety strictly against the patient-specific Reference Ground Truth provided below (do NOT override the Ground Truth's explicit disease severity classification, outpatient vs. inpatient care setting, or explicit diagnostic/management thresholds with external clinical opinions). Remember that the reference extract includes broad monograph background and secondary options; a clinically safe plan that satisfies the Ground Truth's core clinical intent---accurate primary diagnosis, appropriate care setting, sound primary workup, first-line therapy, and essential safety rules---does NOT need to mirror every ancillary detail, generic consult label, or secondary assay to be \textbf{Adherent}.\par

\PromptSep
\noindent\textbf{\underline{SYSTEMATIC CLINICAL RECONCILIATION PROTOCOL}}\par
\noindent You MUST perform a systematic, objective two-step clinical review:\par
\noindent \textbf{1. FACTUAL GROUND TRUTH RECONCILIATION (OBJECTIVE DIFF FIRST):}\par
\begin{itemize}[leftmargin=1.25em, topsep=1pt, itemsep=1.2pt, parsep=0pt]
  \item In \texttt{ground\_truth\_key\_actions\_summary}, concisely summarize the patient-specific Ground Truth's (a) primary diagnosis and key clinical context (acuity, severity tier, pregnancy status, major comorbidities), (b) recommended care setting/triage disposition, (c) essential primary diagnostic investigations, core condition-defining diagnostic indices, and required baseline pre-treatment organ-toxicity, enzyme-deficiency, or bloodborne/TB infectious safety screens for clinician-initiated systemic/controlled therapies, (d) core first-line therapies, preferred safer secondary drug classes, and medications explicitly flagged to stop, hold, or avoid, and (e) central musculoskeletal Physical Therapy or multidisciplinary pain/addiction referrals or formal regulatory boxed safety warnings.
  \item In \texttt{ai\_plan\_key\_actions\_summary}, concisely summarize what the AI plan actually diagnosed, triaged, ordered, prescribed (started, continued, stopped, or deferred), referred, and counseled.
  \item In \texttt{guideline\_omissions\_and\_differences}, neutrally list any substantive discrepancies between the AI plan and the Ground Truth's explicit patient-specific core recommendations across diagnosis/severity, triage/care setting, essential primary diagnostics, core condition-defining diagnostic indices, required baseline pre-treatment safety/infectious screens for clinician-initiated outpatient systemic/controlled therapies, core first-line therapies, preferred safer secondary medication classes, medication hold/avoidance rules, and central musculoskeletal PT/pain-addiction referrals or formal regulatory boxed safety warnings. Disregard stylistic wording, monograph background, benign differential ordering when the Ground Truth condition is explicitly named and managed, generic or specialty clinic consult labels when core workup/management is present, secondary/broad etiology mimic rule-out assays or body-fluid sub-analytes, surveillance imaging, routine preventive/cancer screenings, routine vitamins or vaccination/dosing-titration nuances, downstream inpatient/ICU/surgical/oncology procedures or drug/adjunct/prophylaxis choices and subspecialist invasive procedures/biopsies/pre-therapy panels once appropriately admitted or urgently referred, reasonable conditional staging of non-emergent diagnostics with clear clinical escalation triggers (NOT acute urgent rule-outs such as suspected DVT, PE, stroke, or hemorrhage), safe precautionary empiric antimicrobials, and clinically equivalent regimens. Leave severity grading strictly to Step 2. Write \texttt{'NONE'} if no substantive discrepancies exist.
\end{itemize}

\PromptSep
\noindent \textbf{2. DEFINITIVE CLINICAL SEVERITY HIERARCHY (STRICT STAGE 1 $\rightarrow$ STAGE 4 ORDER):}\par
\noindent \textbf{PRIMARY-CARE \& SPECIALIST HANDOFF PRINCIPLE (APPLIES ACROSS ALL STAGES 1, 2, AND 3):}\par
\noindent When the clinician accurately recognizes an acute, critical, complex, surgical, or specialty/oncology presentation and \textbf{appropriately triages, transfers, or admits the patient to the ED/ICU/NICU/inpatient hospital---OR places an urgent/targeted subspecialty referral}---with sound initial stabilization and primary workup, do \textbf{NOT} penalize omitting (under Stage 1, Stage 2, OR Stage 3):\par
\begin{itemize}[leftmargin=1.25em, topsep=1pt, itemsep=1.2pt, parsep=0pt]
  \item Downstream inpatient, ICU, NICU, surgical, or oncology protocols, procedural techniques, or drug selections (e.g., airway intubation/paralytic/ventilation choices, temporary pacing, invasive hemodynamic catheterization, serial inpatient monitoring/electrolyte/fluid/transfusion protocols, lumbar puncture, bone marrow biopsy, procedural hemostasis steps, minor procedural adjuncts, inpatient surgical antimicrobial prophylaxis, chemotherapy/specialty adjuncts, or routine mechanical/pharmacologic prophylaxis); OR
  \item Subspecialist-performed invasive diagnostics, staging, tissue biopsies, or specialty pre-therapy/surveillance evaluations (e.g., surgical/organ/endometrial biopsy, endoscopy, bronchoscopy, cardiac catheterization or vasoreactivity testing, specialized nuclear imaging, motility testing, invasive fluid sub-assays such as specialized CSF biomarkers or viral/molecular sub-panels, or specialist pre-chemotherapy/biologic/surgical baseline functional/organ panels and oncology surveillance imaging).
\end{itemize}
\noindent \emph{(This handoff rule does NOT excuse omitting explicitly required acute bedside stabilization such as supplemental oxygen for documented hypoxemia or immediate IM epinephrine for active anaphylaxis, nor actively prescribing a contraindicated therapy.)}\par
\end{tcolorbox}

\clearpage

\begin{tcolorbox}[
  colback=PromptMajorBg!45!white,
  colframe=PromptMajor,
  colbacktitle=PromptMajor,
  coltitle=white,
  boxrule=0.65pt,
  arc=2.5pt,
  left=6pt, right=6pt, top=3.5pt, bottom=3.5pt,
  toptitle=2pt, bottomtitle=2pt,
  before skip=2pt, after skip=5pt,
  fonttitle=\PromptTitleFont,
  title={Module 2 of 6 --- STAGE 1: MAJOR DEVIATION (Critical Safety / Emergency Disposition / Contraindicated Lethal Therapy)}
]
\PromptBodyFont
\setlength{\parskip}{1.4pt}
\noindent Assign \textbf{Major Deviation} if a critical patient-safety violation occurs posing direct risk of severe patient harm:\par
\begin{enumerate}[leftmargin=1.35em, topsep=1pt, itemsep=1.2pt, parsep=0pt]
  \item \textbf{Critical Under-Triage / Emergency Disposition Failure:} Proposing routine outpatient management or delayed follow-up for an acute presentation with life- or organ-threatening alarm features requiring immediate ED transfer, ICU care, or urgent inpatient hospitalization.
  \item \textbf{Acute Life-Saving Intervention Omission Without Emergency Triage:} Omitting essential time-critical emergency stabilization or pharmacotherapy while failing to immediately transfer or admit an acutely unstable patient.
  \item \textbf{Severe Diagnostic Inversion Omitting Primary Acute Therapy:} Misidentifying a critical acute pathology as an unrelated condition and consequently omitting mandatory primary acute therapy.
  \item \textbf{Active Counter-Guideline Surgical, Source-Control, or Admission Advice:} Explicitly advising against---or reassuring the patient that surgery/drainage is unnecessary or refractory-only when---mandatory surgical intervention, source control/debridement, or hospital admission is clearly required by the Ground Truth to prevent limb-, organ-, or life-threatening progression.
  \item \textbf{Dangerous or Strictly Contraindicated High-Risk Prescribing:} Prescribing a strictly contraindicated potentially lethal therapy, a grossly unsafe medication dose, or dangerous unmonitored high-risk controlled medications in active misuse.
\end{enumerate}
\noindent \emph{(Note: Once appropriately triaged/admitted to the ED/ICU/inpatient hospital, reserve Stage 1 strictly for active prescribing of a life-threatening contraindicated intervention or active advice against mandatory surgery/admission.)}\par
\end{tcolorbox}

\begin{tcolorbox}[
  colback=PromptModBg!45!white,
  colframe=PromptMod,
  colbacktitle=PromptMod,
  coltitle=white,
  boxrule=0.65pt,
  arc=2.5pt,
  left=6pt, right=6pt, top=3.5pt, bottom=3.5pt,
  toptitle=2pt, bottomtitle=2pt,
  before skip=2pt, after skip=4pt,
  fonttitle=\PromptTitleFont,
  title={Module 3 of 6 --- STAGE 2: MODERATE DEVIATION (Clinically Significant Omission, Commission, or Over-Triage)}
]
\PromptBodyFont
\setlength{\parskip}{1.4pt}
\noindent If Stage 1 does not apply, assign \textbf{Moderate Deviation} if a potentially clinically significant discrepancy occurs relative to the Reference Ground Truth where the patient is likely fine short-term but clinical care is materially compromised (Stage 2 takes precedence over Stage 3):\par
\begin{enumerate}[leftmargin=1.35em, topsep=1pt, itemsep=1.2pt, parsep=0pt]
  \item \textbf{Core Pharmacotherapy, Acute Bedside Stabilization, or Clinical-State Error:} Omitting core first-line guideline-mandated outpatient/emergency prescription therapy, a key time-critical acute bedside stabilization step (e.g., supplemental oxygen for documented hypoxemia---which MUST still be explicitly ordered even when admitting to ED/ICU---immediate IM epinephrine for active anaphylaxis, or time-critical antidote/decontamination), or misidentifying a core patient clinical state (e.g., pregnancy status or explicit disease severity tier) in a way that materially compromises management timing or disposition. \emph{(Do NOT assign Stage 2 for omitting downstream inpatient/ICU/surgical/oncology procedures, intubation escalation, inpatient specialty protocols/adjuncts, or dosing-titration nuances once the patient is appropriately admitted or urgently referred with core primary emergency care.)}
  \item \textbf{Omitted or Explicitly Declined Essential Non-Invasive Primary Diagnostic Study:} Completely omitting---or explicitly advising against or declaring unnecessary---an essential \textbf{non-invasive} primary diagnostic study mandated upfront in the Ground Truth (such as upfront imaging for acute/persistent alarm features or suspected DVT/deep infection, definitive confirmatory microbiology/NAAT for a high-consequence/reportable acute infection, primary ECG/rhythm monitoring for symptomatic arrhythmias/syncope or high-grade heart block, or a formal suicide/acute safety assessment in high-risk psychiatric presentations). \emph{(Do not excuse explicitly telling a patient that a Ground Truth-mandated upfront diagnostic scan is unnecessary based on generic pre-test reassurance. Do NOT apply Stage 2 to invasive inpatient or subspecialist procedures---such as intubation, lumbar puncture, biopsy, endoscopy, catheterization, or specialized nuclear imaging---when appropriately admitted or referred, nor to reasonable conditional staging of non-emergent diagnostics with clear clinical escalation triggers (NOT acute urgent rule-outs such as suspected DVT, PE, stroke, or hemorrhage).)}
  \item \textbf{Offending Drug Continuation, Contraindicated/Harmful Prescribing, or Dropped Essential Therapy:} Failing to stop, pause, or adjust a medication identified in the Ground Truth as precipitating or worsening the presenting condition; dropping essential ongoing chronic therapy without justification; creating a serious drug-drug interaction or unsafe duplication; prescribing a medication that materially worsens a documented comorbidity or is explicitly contraindicated/avoided due to toxicity or interference with acute care; OR directly initiating a high-risk specialist-restricted medication that the Ground Truth stipulates must be started only under specialist evaluation. \emph{(Note: Adding a benign/low-risk agent noted solely as lacking proven benefit belongs in Stage 3 Minor Deviation, NOT Stage 2.)}
  \item \textbf{Significant Over-Triage or Premature Invasive Intervention:} Over-classifying an explicitly uncomplicated or non-severe outpatient presentation as severe and escalating to unwarranted hospital/ICU admission and aggressive parenteral inpatient therapy when the Ground Truth explicitly indicates outpatient care; OR recommending invasive surgery or permanent device implantation before completing standard non-invasive diagnostic monitoring or conservative medical trials.
\end{enumerate}

\PromptSep
\noindent \textbf{CLINICAL CONTEXT NUANCES FOR STAGE 2 --- DO NOT ASSIGN MODERATE DEVIATION FOR:}\par
\begin{itemize}[leftmargin=1.25em, topsep=1pt, itemsep=1pt, parsep=0pt]
  \item Reasonable conditional staging of non-emergent diagnostics or secondary referrals when the clinician explicitly plans the study with clear clinical escalation triggers (as opposed to omitting it, declaring an indicated upfront scan unnecessary, or deferring urgent rule-outs such as suspected DVT, PE, stroke, or acute hemorrhage);
  \item Clinically equivalent first-line drug regimens (including equivalent antimicrobial coverage) or valid guideline-recognized alternative diagnostic modalities;
  \item Standard pre-test clinical probability pathways explicitly endorsed by guidelines or reasonable initial conservative/dietary trials with clear escalation triggers in mild stable outpatient imbalances;
  \item Stopping a contraindicated chronic drug [e.g., in pregnancy] with safe baseline coverage while staging or deferring full regimen escalation [evaluate strictly under Stage 3 Minor Deviation, NOT Stage 2 Moderate Deviation];
  \item Benign asymptomatic conduction or incidental finding surveillance when no high-grade block or red flags exist;
  \item Minor outpatient dosing-titration or referral-timing phrasing nuances when the core medication and appropriate urgent disposition are ordered;
  \item Downstream inpatient/ICU/surgical/oncology procedures, antimicrobial prophylaxis, or drug/adjunct choices [airway intubation/paralytics/ventilation, temporary pacing, hemodynamic catheterization, serial monitoring, lumbar puncture, bone marrow biopsy, chemotherapy/specialty adjuncts] or subspecialist invasive procedures [endoscopy, biopsy, specialized nuclear imaging] deferred to an ordered ED/ICU/NICU/inpatient admission or specialty referral.
\end{itemize}
\end{tcolorbox}

\clearpage

\begin{tcolorbox}[
  colback=PromptMinorBg!45!white,
  colframe=PromptMinor,
  colbacktitle=PromptMinor,
  coltitle=white,
  boxrule=0.65pt,
  arc=2.5pt,
  left=5pt, right=5pt, top=2pt, bottom=2pt,
  toptitle=1.6pt, bottomtitle=1.6pt,
  before skip=1pt, after skip=2.2pt,
  fonttitle=\PromptTitleFont,
  title={Module 4 of 6 --- STAGE 3: MINOR DEVIATION (Clear Focal Omission or Suboptimal Choice with Safe Core Care)}
]
\PromptBodyFont
\setlength{\parskip}{0.8pt}
\noindent If Stages 1 and 2 do not apply, assign \textbf{Minor Deviation} ONLY when primary diagnosis, care setting, and first-line therapy are clinically sound AND there is a clear, concrete focal shortcoming relative to the \textbf{patient-specific Ground Truth's explicit primary recommendations}:\par
\begin{enumerate}[leftmargin=1.35em, topsep=0.4pt, itemsep=0.5pt, parsep=0pt]
  \item \textbf{Clear Pre-Treatment Safety Screen Gap or Distinct Central Primary Workup Omission:} Omitting a required baseline organ-toxicity, enzyme-deficiency, or bloodborne/TB infectious safety screen (e.g., baseline hepatic/lipid panels, hemolytic-risk/G6PD screen prior to high-risk oxidant therapy, or baseline HIV/hepatitis/syphilis/TB screening) explicitly listed in the Ground Truth for a high-risk systemic, acute, or controlled therapy \textbf{directly initiated or ordered by the clinician} (NOT routine non-toxicity hormone levels or specialist-deferred pre-chemotherapy/biologic/surgical panels delegated to referral); OR omitting a distinct, non-redundant primary diagnostic investigation directly central to the chief syndrome evaluation---such as acute EEG in unexplained encephalopathy/delirium, urine toxicology in new unexplained psychiatric symptoms, reticulocyte count in unexplained pancytopenia, targeted reversible vitamin B12/folate deficiency labs in unexplained neurologic/ocular deficits, or an explicitly required core condition-defining quantitative metabolic/renal diagnostic index (or 24-hour urine volume/protein quantification) directly needed to establish the primary condition \emph{(distinct from broad secondary endocrine/trace-metal mimic rule-out screens)}.
  \item \textbf{Suboptimal Secondary Drug Choice, Futile Low-Risk Agent, Staged Transition, or Premature Lifelong Maintenance Start:} Choosing or suggesting (even conditionally) a clearly suboptimal older/first-generation medication class when the Ground Truth specifies a safer preferred class (e.g., a first-generation agent when a safer second-generation class is preferred --- NOT inpatient/ICU airway paralytics or procedural drug choices covered by Handoff); adding a futile low-risk agent explicitly flagged as lacking clinical benefit; appropriately stopping an unsafe chronic drug while staging/delaying its indicated replacement; omitting a highlighted primary outpatient medication (e.g., oral antihistamine for pruritus or early oral cardio-protective beta-blocker) when no equivalent is provided; OR initiating \textbf{lifelong or prolonged organ-specific hormone replacement or toxicity-monitored systemic maintenance therapy (e.g., thyroid hormone or systemic oral antifungal)} before pending confirmatory laboratory results return \emph{(do NOT penalize safe precautionary empiric antimicrobials, short symptom-directed steroid courses, or preventive cardiovascular medications when appropriate diagnostics are ordered)}.
  \item \textbf{Omitted Core General PT / Pain-Addiction Referral or Formal Regulatory Boxed Safety Alert:} Omitting an explicitly central formal \textbf{general Physical Therapy referral} (for primary joint/limb/spine complaints --- NOT organ-specific or subspecialty rehab programs) OR \textbf{specialized multidisciplinary pain/addiction center referral} directly addressing a primary musculoskeletal, chronic pain, or substance-use complaint; OR omitting an explicitly designated \textbf{formal regulatory (e.g., FDA/EMA) boxed or major safety warning} emphasized in the Ground Truth on a primary prescribed medication (NOT routine drug side-effect counseling or generic risk statistics).
\end{enumerate}

\PromptSep
\noindent \textbf{STRICTLY ADHERENT CALIBRATION EXCLUSIONS --- RATE STAGE 4 ADHERENT, NOT MINOR DEVIATION:}\par
\begin{itemize}[leftmargin=1.25em, topsep=0.4pt, itemsep=0.5pt, parsep=0pt]
  \item \textbf{Secondary Mimic Rule-Outs, Sub-Assays, \& Surveillance Scans:} When core primary workup and management/triage are sound, do NOT penalize omitting secondary/broad etiology or rare-mimic rule-out labs (e.g., secondary endocrine/hormone, broad metabolic/trace-metal screens, pathogen/viral serologies or broad molecular panels when standard cultures and empiric therapy are ordered, autoimmune, paraneoplastic, genetic/karyotype, or antibody sub-titer panels, body-fluid/CSF sub-analytes on an ordered tap, or serum nutrition markers), baseline structural/vascular ultrasounds or surveillance imaging (e.g., echocardiography or bone-density surveillance), specialized ECG sub-leads, benign asymptomatic finding surveillance, benign differential rank-ordering when the Ground Truth condition is explicitly named and managed, valid superseding modalities, or reasonable conditional staging of non-emergent diagnostics with clear escalation triggers. \emph{(Required baseline HIV/hepatitis/syphilis/TB, organ-toxicity, or hemolytic/G6PD safety screens for directly initiated high-risk therapies and core condition-defining metabolic/renal diagnostic indices remain graded under Stage 3 Pillar 1.)}
  \item \textbf{Clinic Labels, Referral Timing, \& Inpatient/Subspecialist Handoff Scope:} Do NOT penalize omitting generic single-specialty consult labels (e.g., Endocrinology, Pulmonology, Gynecology) or general supportive/organ-specific clinic labels (e.g., specialized post-illness clinics, subspecialty organ-specific rehab programs, dietary consults, cardiac/pulmonary rehab) when core workup and management are ordered \emph{(still require formal general Physical Therapy for primary joint/spine complaints and specialized multidisciplinary pain/addiction center referrals when explicitly central to a primary pain/substance complaint)}, minor urgency-timing phrasing when an urgent referral is already placed, nor ANY downstream inpatient/ICU/NICU/surgical/oncology procedure, monitoring schedule, paralytic/antimicrobial-prophylaxis/chemo-adjunct drug selection, procedural technique, dietary/fluid/blood-product timing nuance, electrolyte binder, or subspecialist invasive biopsy/endoscopy/catheterization/motility/pre-therapy functional or organ evaluation once appropriately admitted or urgently referred.
  \item \textbf{Routine Prophylaxis, Preventive Counseling, Titration, \& Clinical Equivalents:} Do NOT penalize omitting routine mechanical or pharmacologic VTE prophylaxis, gastroprotection, routine vitamins or nutritional maintenance, routine chronic or cancer screenings, asymptomatic high-risk surveillance scan substitutions, minor vaccination/preventive/dosing-interval/titration or exact statistical-risk counseling nuances, minor follow-up interval differences, symptom/prognostic questionnaires or score acronyms, clinically equivalent regimens, precautionary empiric antimicrobials, staged multi-drug chronic disease titration or specialist-deferred induction therapy pending urgent confirmatory biopsy/endoscopy, or reasonable conservative trials with clear escalation triggers.
\end{itemize}
\end{tcolorbox}

\begin{tcolorbox}[
  colback=PromptAdhBg!45!white,
  colframe=PromptAdh,
  colbacktitle=PromptAdh,
  coltitle=white,
  boxrule=0.65pt,
  arc=2.5pt,
  left=5pt, right=5pt, top=2pt, bottom=2pt,
  toptitle=1.6pt, bottomtitle=1.6pt,
  before skip=1pt, after skip=2.2pt,
  fonttitle=\PromptTitleFont,
  title={Module 5 of 6 --- STAGE 4: ADHERENT / ACCEPTABLE VARIATION (Standard-of-Care Concordance)}
]
\PromptBodyFont
\setlength{\parskip}{0.8pt}
\noindent Assign \textbf{Adherent} if Stages 1, 2, and 3 are all \texttt{'NONE'}:\par
\begin{itemize}[leftmargin=1.25em, topsep=0.4pt, itemsep=0.5pt, parsep=0pt]
  \item The AI plan captures the clinical spirit and standard of care of the Reference Ground Truth: accurate diagnostic framing, appropriate care setting, sound primary diagnostic workup, appropriate first-line therapy, medication hold/avoidance safety rules, and any central general PT/pain-addiction referral or formal regulatory boxed safety alert.
  \item Clinically equivalent first-line regimens, valid alternative diagnostic modalities, conditional staging with escalation triggers, inpatient/subspecialty handoff delegation, benign differential ordering, non-exhaustive omission of ancillary monograph details, and all calibration exclusions above are strictly \textbf{Adherent}.
\end{itemize}
\end{tcolorbox}

\par\vspace{3pt}
\noindent\makebox[\textwidth][c]{\fontsize{7.2pt}{8.4pt}\selectfont\textcolor{PromptSlateSub}{\emph{[10 synthetic few-shot exemplars inserted here]}}}
\par\vspace{3pt}

\begin{tcolorbox}[
  colback=PromptSlateBg,
  colframe=PromptSlate,
  colbacktitle=PromptSlate,
  coltitle=white,
  boxrule=0.65pt,
  arc=2.5pt,
  left=5pt, right=5pt, top=2.5pt, bottom=2.5pt,
  toptitle=1.6pt, bottomtitle=1.6pt,
  before skip=1pt, after skip=0pt,
  fonttitle=\PromptTitleFont,
  title={Module 6 of 6 --- Per-Case Evaluation Input Template (\texttt{ACTUAL CASE EVALUATION})}
]
\PromptBodyFont
\setlength{\parskip}{0.6pt}
\noindent\textbf{\# PATIENT SCENARIO} \texttt{\{patient\_scenario\}}\par
\noindent\textbf{\# GROUND TRUTH CLINICAL GUIDELINE \& EVIDENCE} \texttt{\{ground\_truth\}}\par
\noindent\textbf{\# AI CLINICIAN PROPOSED RESPONSE} \texttt{\{ai\_response\}}\par
\end{tcolorbox}
\par\vspace{3pt}
\noindent\begin{minipage}{\textwidth}
\captionsetup{aboveskip=0pt,belowskip=0pt}
\captionof{figure}{Clinician-calibrated four-point ordinal autorater system prompt: role and reconciliation protocol, severity hierarchy and exclusions, and per-case input template. Structured JSON outputs are enforced via constrained decoding (schema suffix omitted).\label{fig:lst_autorater_prompt}}
\end{minipage}
\clearpage
\subsection{Autorater Repeatability and Perturbation Sensitivity}
\label{appendix:autorater_consistency}

To assess how sensitive the calibrated autorater (Gemini 3.8 Flash) is to prompt ordering, list structure, and response length when clinical content is held fixed, we ran repeatability and controlled perturbation checks across all $N = 200$ standalone AI management plans from the calibration set, comparing each single perturbed pass against the unperturbed baseline pass (\cref{tab:supp_autorater_unit_tests}):

\begin{enumerate}[leftmargin=1.5em, topsep=2pt, itemsep=2pt, parsep=0pt]
    \item \textbf{Repeat-Sampling (Test--Retest) Consistency ($98.5\%$ exact 4-tier agreement):}
    Re-running the autorater independently on the identical scenario, ground truth, and candidate response yields $197/200$ ($98.5\%$) exact four-tier agreement. All $3$ disagreements are single-step shifts ($2$ between \emph{Adherent} [Score 4] and \emph{Minor Deviation} [Score 3], and $1$ between \emph{Minor Deviation} [Score 3] and \emph{Moderate Deviation} [Score 2]), giving $100.0\%$ ($200/200$) adjacent-tier agreement ($|\Delta\text{tier}| \le 1$).

    \item \textbf{Presentation Order ($95.5\%$ exact 4-tier agreement):}
    Swapping the prompt order of \texttt{\# GROUND TRUTH CLINICAL GUIDELINE \& EVIDENCE} and \texttt{\# AI CLINICIAN PROPOSED RESPONSE} preserves $95.5\%$ ($191/200$) exact four-tier agreement ($100.0\%$ adjacent-tier agreement).

    \item \textbf{List and Field Ordering ($96.0\%$ exact 4-tier agreement):}
    Randomly permuting the order of alternative differential diagnoses, diagnostic investigations, and treatment actions within the candidate plan, together with reversing top-level and nested JSON field ordering at identical character length ($1.00\times$), yields $96.0\%$ ($192/200$) exact four-tier agreement ($100.0\%$ adjacent-tier agreement).

    \item \textbf{Response Length and Style ($96.0\%\text{--}97.0\%$ exact 4-tier agreement):}
    Deterministically compressing candidate items into concise telegraphic phrasing by stripping parenthetical rationales and boilerplate wording ($0.85\times$ mean character length) or expanding each item with narrative framing ($2.21\times$ mean character length, a $2.6\times$ range in length across identical clinical actions) preserves $97.0\%$ ($194/200$; $99.5\%$ adjacent-tier agreement) and $96.0\%$ ($192/200$; $100.0\%$ adjacent-tier agreement) exact four-tier agreement, indicating low sensitivity to verbosity.
\end{enumerate}

\begin{table}[!htbp]
\centering
\caption{Autorater repeat-sampling consistency and sensitivity to ordering, structural, and stylistic perturbations across the 200 standalone AI management plans in the calibration set ($N = 200$ scenarios; Gemini 3.8 Flash autorater).}
\label{tab:supp_autorater_unit_tests}
\footnotesize
\setlength{\tabcolsep}{4.5pt}
\renewcommand{\arraystretch}{1.15}
\begin{tabularx}{\textwidth}{l>{\raggedright\arraybackslash}Xccc}
\toprule
\textbf{Perturbation Dimension} & \textbf{Perturbation Construction} & \textbf{Rel.\ Length} & \textbf{Exact 4-Tier} & \textbf{Adjacent ($|\Delta| \le 1$)} \\
\midrule
Repeat Sampling & Independent re-evaluation of identical prompt and response & $1.00\times$ & $98.5\%$ ($197/200$) & $100.0\%$ ($200/200$) \\
Presentation Order & Swap Ground Truth and AI Response order in prompt & $1.00\times$ & $95.5\%$ ($191/200$) & $100.0\%$ ($200/200$) \\
Structural Order & Permute intra-response lists and reverse JSON key ordering & $1.00\times$ & $96.0\%$ ($192/200$) & $100.0\%$ ($200/200$) \\
Output Length (Shorter) & Concise telegraphic phrasing (stripped rationales) & $0.85\times$ & $97.0\%$ ($194/200$) & $99.5\%$ ($199/200$) \\
Output Length (Longer) & Verbose narrative prose framing around identical actions & $2.21\times$ & $96.0\%$ ($192/200$) & $100.0\%$ ($200/200$) \\
\bottomrule
\end{tabularx}
\end{table}

\clearpage
\section{Primary Care Physician Baseline and AI-Assistance Study}
\label{appendix:pcp_study_protocol}

We recruited $n = \numpcps$ board-certified primary care physicians with a median of 11 years of post-residency clinical experience (IQR: 11--12; range: 5--26; mean $12.3 \pm 5.6$ years) to evaluate the 200-scenario evaluation set (50 scenarios per clinical category: Differential Diagnosis, Investigations, Treatment, and Complications; one physician per scenario) in a paired two-stage sequential study. 

In the unassisted first stage (\emph{PCP Alone}), each physician reviewed the patient scenario and wrote a free-text management plan covering differential diagnosis, diagnostic investigations, and treatment recommendations, with standard non-AI clinical references permitted. Immediately after submitting that response, the interface displayed an AI recommendation (from Gemini 3 Flash with \emph{Search + Synth-4}); in the second stage (\emph{PCP + AI}), the physician could revise or keep their initial plan. Where physicians entered shorthand post-AI revision notes (e.g., ``no change'' or brief incremental additions), an automated normalization step (deterministic rules and Gemini 3 Flash) converted them into standalone clinical plans while preserving original wording and removing references to AI assistance before grading. 

Both pre- and post-AI plans were scored across five runs of the calibrated autorater (taking the median ordinal rating), and paired pre- and post-AI plans were compared using Gemini 3.8 Flash to categorize unchanged responses, clinical additions, and detrimental edits, with $95\%$ confidence intervals computed via 1,000 care-setting-stratified scenario bootstrap resamples (\cref{fig:clinician_baseline}; case examples in \cref{appendix:representative_pcp_examples}).

Because scenarios span the breadth of the guideline corpus, we stratified the 200 scenarios post-hoc into three care settings based on the clinical setting expected to execute the complete ground-truth workup and management plan:
\begin{enumerate}[leftmargin=1.5em, topsep=1pt, itemsep=0.5pt, parsep=0pt]
    \item \textbf{Primary Care (Clinic) ($N = 95$, $47.5\%$):} Outpatient presentations whose complete ground-truth diagnosis, diagnostic workup, and management plan fall within routine clinic-based primary care.
    \item \textbf{Emergency / Inpatient Generalist ($N = 52$, $26.0\%$):} Acute presentations requiring emergency department or hospital care whose stabilization, workup, and inpatient management fall within emergency medicine or general hospitalist practice.
    \item \textbf{Specialist Care ($N = 53$, $26.5\%$):} Scenarios whose underlying condition or required diagnostic testing or therapeutic protocol exceeds generalist scope and would typically be directed by a domain specialist or subspecialist.
\end{enumerate}
Care-setting labels were assigned using Gemini 3.8 Flash from each vignette, ground-truth criteria, and source \emph{BMJ Best Practice} extract (\cref{fig:lst_pcp_stratification_prompt}), with clinician spot-checks.

\begin{figure}[p]
\centering
\begin{tcolorbox}[
  colback=gray!3,
  colframe=black!70,
  colbacktitle=black!75,
  coltitle=white,
  boxrule=0.6pt,
  arc=2pt,
  left=6pt, right=6pt, top=3pt, bottom=3pt,
  toptitle=2pt, bottomtitle=2pt,
  fonttitle=\footnotesize\bfseries\raggedright,
  title={Post-Hoc Scenario Care-Setting Stratification Prompt}
]
\fontsize{7.3pt}{8.4pt}\selectfont
\RaggedRight
\setlength{\parskip}{1.0pt}

\begin{tcolorbox}[colback=white, colframe=black!25, boxrule=0.4pt, arc=1.5pt, left=5pt, right=5pt, top=2pt, bottom=2pt]
\noindent\textbf{System Role \& Case Inputs}\\[0.5pt]
You are an expert physician educator and clinical guideline auditor reviewing a clinical vignette and its Ground Truth diagnosis and management plan against an official \emph{BMJ Best Practice} clinical guideline. Your goal is to classify the scenario into the appropriate clinical care setting and provide a concise clinical rationale.\\[0.5pt]
\textbf{Inputs:} (1) \texttt{SCENARIO METADATA} (\texttt{\{unique\_scenario\_key\}}, \texttt{\{guideline\_id\}}, \texttt{\{guideline\_title\}}, \texttt{\{category\}}, \texttt{\{medical\_specialty\}}, \texttt{\{scenario\_focus\}}); (2) \texttt{CLINICAL SCENARIO} (Age/Sex/Background, \texttt{\{presenting\_complaint\}}, \texttt{\{initial\_presentation\}}, HPI \texttt{\{info\_to\_be\_given\_if\_asked\}}, \texttt{\{objective\_findings\}}, \texttt{\{pmh\}}, \texttt{\{psh\}}, \texttt{\{medications\}}, \texttt{\{allergies\}}, \texttt{\{family\_history\}}, \texttt{\{social\_history\}}, \texttt{\{concerns\}}); (3) \texttt{GROUND TRUTH DIAGNOSIS \& MANAGEMENT PLAN} (\texttt{\{probable\_diagnosis\}}, \texttt{\{alternative\_diagnoses\}}, \texttt{\{treatment\_plan\}}, \texttt{\{medical\_treatments\}}, \texttt{\{investigations\}}, \texttt{\{lifestyle\_modifications\}}, \texttt{\{red\_flags\}}, \texttt{\{guideline\_justification\}}); and (4) \texttt{RELEVANT BMJ GUIDELINE EXTRACT} (\texttt{\{xml\_guideline\_extract\}}).
\end{tcolorbox}

\vspace{0.8pt}
\begin{tcolorbox}[colback=white, colframe=black!25, boxrule=0.4pt, arc=1.5pt, left=5pt, right=5pt, top=2pt, bottom=2pt]
\noindent\textbf{Clinical Competency / Care Level Classification Criteria (\texttt{care\_level} \& \texttt{care\_level\_rationale})}\\[0.5pt]
Classify the scenario into ONE of the following three clinical competency groupings based on \textbf{who we would reasonably expect to know and properly execute the full Ground Truth diagnosis, diagnostic workup, and management plan}:
\begin{enumerate}[leftmargin=1.4em, topsep=0pt, itemsep=1pt, parsep=0pt]
  \item \textbf{\texttt{"Primary Care (Clinic)"}} --- \emph{Expected of an Outpatient Primary Care Physician (PCP):} Both the clinical presentation AND the \textbf{complete Ground Truth diagnosis, investigations, and management plan} are squarely within the expected day-to-day competency of a general outpatient Primary Care Physician (Family Medicine / Outpatient General Internal Medicine). We expect a competent PCP to diagnose this condition, order the Ground Truth workup, and initiate the Ground Truth management plan themselves in an outpatient clinic (e.g., essential hypertension, routine non-pregnant T2DM, hyperlipidemia, primary hypothyroidism, depression/anxiety, migraine/tension headache, outpatient CAP, acute otitis/rhinosinusitis, osteoarthritis, gout, GERD/PUD, uncomplicated UTI, outpatient PID, early localized Lyme disease, routine prenatal screening, outpatient OUD buprenorphine maintenance).
  \item \textbf{\texttt{"Emergency / Inpatient Generalist"}} --- \emph{Expected of an Emergency Physician or Inpatient IM Hospitalist / Generalist:} The scenario requires Emergency Department or acute inpatient hospital care, AND the \textbf{complete Ground Truth resuscitation, diagnostic workup, and hospital management plan} are squarely within the expected competency of an Emergency Medicine physician or an Inpatient Internal Medicine Hospitalist / Generalist without requiring subspecialist-only testing or protocols. We would not expect an outpatient clinic PCP without hospital resources to manage this in clinic, \textbf{but we WOULD expect an ED physician or inpatient IM hospitalist} to know and properly execute the Ground Truth stabilization, workup, and hospital treatment plan, including calling standard acute surgical/GI consults for common emergencies (e.g., unstable AFib with RVR, NSTEMI/ACS, pulmonary embolism, acute decompensated heart failure, sepsis/septic shock resuscitation, bacterial meningitis, acute GI bleeding, rhabdomyolysis, prerenal AKI, common overdoses/toxicology, heat stroke/hypothermia, Wernicke or hepatic encephalopathy/SBP, bowel obstruction, acute cholecystitis, complicated diverticulitis or acute prostatitis with abscess/retention, acute ICH initial stabilization).
  \item \textbf{\texttt{"Specialist Care"}} --- \emph{Only a Domain Specialist / Subspecialist Would Be Expected to Manage Properly:} Either the condition itself OR the \textbf{specific Ground Truth diagnostic tests / management protocol} goes beyond what we would reasonably expect either an outpatient PCP or a general ED / inpatient hospitalist to know and execute properly on their own---only a domain specialist / subspecialist would be expected to manage it properly according to the Ground Truth rubric. \textbf{Critical Rule:} Do NOT classify a scenario as \texttt{"Primary Care (Clinic)"} or \texttt{"Emergency / Inpatient Generalist"} merely because the patient walks into an outpatient clinic or ER; inspect the \textbf{Ground Truth Diagnosis, Investigations, and Management Plan}:
  \begin{itemize}[leftmargin=1.2em, topsep=0pt, itemsep=0pt]
    \item If the Ground Truth requires \textbf{subspecialist diagnostic workups} that generalists do not routinely order on their own (e.g., pre/post-pneumococcal polysaccharide vaccine serotype antibody titers for Specific Antibody Deficiency, skin biopsy with Direct Immunofluorescence for cutaneous lupus, islet autoantibodies GAD/IA2/ZnT8 + C-peptide for new-onset T1DM in pregnancy, serum Galactomannan + Beta-D-glucan + Myco/F Lytic fungal cultures for breakthrough invasive fungal infection in febrile neutropenia, colonic manometry, paraneoplastic CRMP5 antibodies, volumetric brain MRI + formal speech-pathology aphasia battery for Semantic Dementia, fetal echocardiogram + chromosomal microarray for fetal omphalocele/cloacal exstrophy, pacemaker/ICD lead impedance \& threshold interrogation, HLA-B51 / ICBD scoring + slit-lamp exam for Beh\c{c}et's), OR
    \item If the Ground Truth requires \textbf{subspecialist-directed therapy or surgical/interventional procedures} (e.g., multiple myeloma, ALL, Hodgkin lymphoma t-AML/MDS, tumor lysis syndrome, or metastatic breast cancer with spinal cord compression; high-dose methotrexate crystal nephropathy rescue; switching voriconazole to echinocandin/liposomal amphotericin B for hepatosplenic candidiasis; HIV-TB paradoxical IRIS; lateral canthotomy and cantholysis for ocular compartment syndrome; MCTD / Systemic Sclerosis / Lupus / RA subspecialty immunomodulation; pregnancy-specific basal-bolus insulin conversion or HELLP/preeclampsia OB delivery management; pheochromocytoma $\alpha$/$\beta$-blockade \& resection; neurogenic orthostatic hypotension specialist titration; COPD with bronchiectasis \& \emph{Pseudomonas} eradication; Asbestosis Group 3 PH / cor pulmonale; recurrent Mollaret meningitis or brain abscess; ovarian torsion in pregnancy; GCA with arteritic ischemic optic neuropathy; septic olecranon bursitis surgical bursectomy; BPH minimally invasive surgical therapy), then you \textbf{MUST} classify it as \texttt{"Specialist Care"}.
  \end{itemize}
\end{enumerate}
Provide \texttt{care\_level\_rationale}: A concise, clear 1--2 sentence clinical rationale explaining why the Ground Truth workup and management plan are expected of an outpatient PCP (\texttt{"Primary Care (Clinic)"}), an ED/inpatient hospitalist (\texttt{"Emergency / Inpatient Generalist"}), or only a domain specialist (\texttt{"Specialist Care"}).
\end{tcolorbox}
\end{tcolorbox}
\vspace{-6pt}
\caption{Prompt template used with Gemini 3.8 Flash to assign post-hoc care-setting labels (\texttt{Primary Care (Clinic)}, \texttt{Emergency / Inpatient Generalist}, and \texttt{Specialist Care}) and clinical rationales across the 200 physician-study scenarios.}
\label{fig:lst_pcp_stratification_prompt}
\end{figure}

\clearpage
\section{Closed-Corpus Dynamic Guideline Retrieval Mechanics}
\label{appendix:dynamic_retrieval}

To examine how retrieval capacity $K$ shapes closed-corpus grounding, we run an independent end-to-end evaluation of Gemini 3.8 Flash with the full-monograph \emph{Dynamic Guideline} harness across the core benchmark at each retrieval capacity ($K \in \{1, 3, 5, 7, 9\}$; \cref{fig:supp_dynamic_retrieval}). Increasing $K$ lifts overall recall, while precision drops from $58.3\%$ to $19.2\%$ at $K=9$ as competing documents enter context (\cref{fig:supp_dynamic_retrieval}c--d). Reflecting this recall--distraction trade-off, while overall adherence even at $K=1$ ($87.0\%$) exceeds the reported \emph{Baseline} ($83.1\%$), performance largely plateaus from $K = 3$ onward (\cref{fig:supp_dynamic_retrieval}a). Recall is particularly challenging in \emph{Differential Diagnosis} cases, reaching just $56.1\%$ at $K=5$ (\cref{fig:supp_dynamic_retrieval}c).

\vspace{0.05cm}
\noindent\begin{minipage}{\textwidth}
  \centering
  \includegraphics[width=0.92\textwidth]{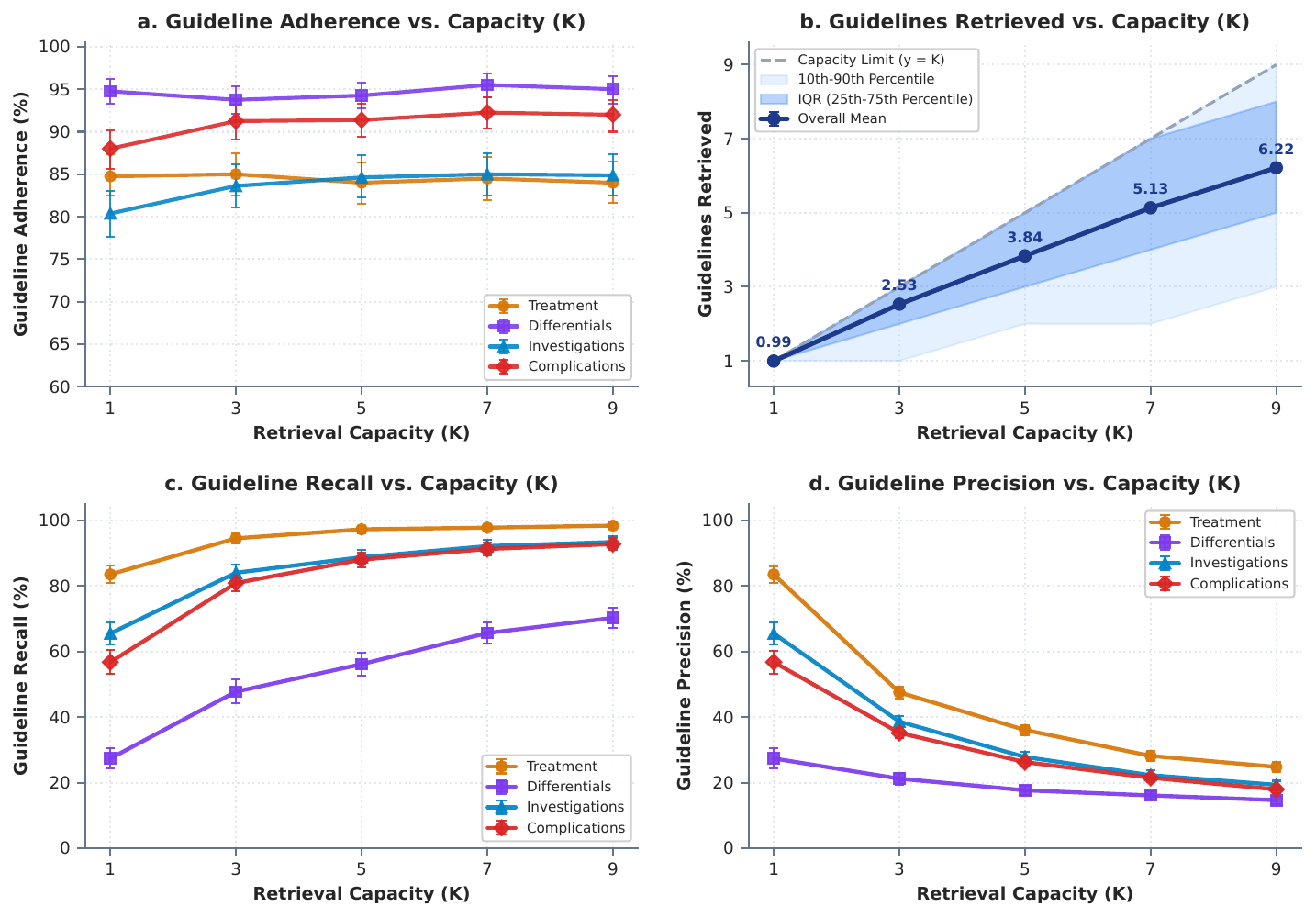}
  \vspace{-0.2cm}
  \captionof{figure}{\textbf{Closed-corpus dynamic guideline retrieval mechanics on Gemini 3.8 Flash across all 3,200 core benchmark scenarios.} \textbf{a}. Guideline adherence ($\%$) vs.\ $K$. \textbf{b}. Clinical monographs actually retrieved vs.\ the specified capacity limit. \textbf{c}. Recall ($\%$) for the clinical monograph used to generate the scenario vs.\ $K$. \textbf{d}. Precision ($\%$) over the set of retrieved monographs ($1/N_{\mathrm{retrieved}}$ on hit, $0$ otherwise) vs.\ $K$. Error bars show $95\%$ CIs across 1,000 scenario-level bootstrap iterations.}
  \label{fig:supp_dynamic_retrieval}
\end{minipage}

\clearpage
\section{Extended Model and Inference Harness Performance}
\label{appendix:extended_results}

\cref{tab:supp_full_model_matrix} reports the full four-tier rating distributions (Major Deviation [Score 1], Moderate Deviation [Score 2], Minor Deviation [Score 3], and Adherent [Score 4]) across all seven Gemini Flash Lite and Flash models and eight inference configurations on the core benchmark ($N = 3{,}200$ scenarios; five runs per scenario), supplying the numerical values plotted in \cref{fig:system_scaling_context} alongside scenario-clustered $95\%$ bootstrap confidence intervals (1,000 resamples).

\begin{table}[ht!]
\centering
\caption{Full model and inference configuration performance across the four-point clinical appropriateness scale ($N = 3{,}200$ scenarios; five runs per scenario). Values show the percentage of responses in each tier with scenario-clustered 95\% bootstrap confidence intervals (1,000 resamples).}
\label{tab:supp_full_model_matrix}
\fontsize{6.8pt}{7.6pt}\selectfont
\setlength{\tabcolsep}{7pt}
\renewcommand{\arraystretch}{0.92}
\resizebox{\textwidth}{!}{%
\begin{tabular}{llcccc}
\toprule
\textbf{Model} & \textbf{Inference Harness} & \textbf{Major Deviation (\%)} & \textbf{Moderate Deviation (\%)} & \textbf{Minor Deviation (\%)} & \textbf{Adherent (\%)} \\
\midrule
\multirow{8}{*}{\textbf{Gemini 3.1 Flash Lite}} & Baseline & 0.78 [0.54, 1.03] & 16.88 [15.76, 17.99] & 10.75 [10.00, 11.57] & 71.60 [70.24, 72.89] \\
 & With Search & 0.79 [0.55, 1.05] & 16.87 [15.78, 17.98] & 10.42 [9.66, 11.22] & 71.92 [70.59, 73.22] \\
 & Synthesize-4 & 0.66 [0.44, 0.90] & 15.14 [14.05, 16.18] & 9.84 [9.03, 10.64] & 74.36 [73.02, 75.62] \\
 & Search + Synth-4 & 0.69 [0.47, 0.94] & 15.53 [14.47, 16.59] & 9.25 [8.59, 10.04] & 74.54 [73.16, 75.75] \\
 & Dynamic Guideline & 0.87 [0.64, 1.13] & 16.09 [15.01, 17.16] & 9.86 [9.09, 10.58] & 73.18 [71.89, 74.43] \\
 & Dynamic Excerpt & 0.57 [0.38, 0.81] & 12.33 [11.40, 13.23] & 8.60 [7.88, 9.35] & 78.49 [77.34, 79.71] \\
 & Oracle Guideline & 0.68 [0.46, 0.93] & 13.17 [12.23, 14.13] & 8.63 [7.96, 9.40] & 77.53 [76.29, 78.71] \\
 & Oracle Excerpt & 0.46 [0.28, 0.67] & 7.86 [7.10, 8.57] & 5.56 [5.05, 6.10] & 86.13 [85.20, 87.09] \\
\midrule
\multirow{8}{*}{\textbf{Gemini 3.5 Flash Lite}} & Baseline & 0.69 [0.44, 0.94] & 14.99 [13.95, 16.06] & 10.20 [9.46, 11.03] & 74.11 [72.72, 75.44] \\
 & With Search & 0.74 [0.50, 0.98] & 15.24 [14.25, 16.32] & 9.91 [9.16, 10.69] & 74.12 [72.81, 75.44] \\
 & Synthesize-4 & 0.63 [0.41, 0.88] & 13.26 [12.24, 14.28] & 9.36 [8.61, 10.19] & 76.75 [75.42, 77.98] \\
 & Search + Synth-4 & 0.71 [0.47, 0.95] & 13.59 [12.59, 14.60] & 9.15 [8.36, 9.96] & 76.55 [75.34, 77.84] \\
 & Dynamic Guideline & 0.49 [0.30, 0.70] & 11.93 [10.99, 12.81] & 8.42 [7.75, 9.14] & 79.16 [78.02, 80.34] \\
 & Dynamic Excerpt & 0.44 [0.26, 0.66] & 9.82 [8.99, 10.66] & 6.99 [6.36, 7.62] & 82.74 [81.72, 83.84] \\
 & Oracle Guideline & 0.52 [0.31, 0.76] & 9.96 [9.08, 10.85] & 7.42 [6.82, 8.11] & 82.10 [80.92, 83.24] \\
 & Oracle Excerpt & 0.40 [0.21, 0.62] & 6.15 [5.47, 6.84] & 5.09 [4.56, 5.63] & 88.36 [87.46, 89.24] \\
\midrule
\multirow{8}{*}{\textbf{Gemini 3 Flash}} & Baseline & 0.79 [0.54, 1.06] & 14.59 [13.57, 15.62] & 9.85 [9.07, 10.66] & 74.77 [73.42, 76.11] \\
 & With Search & 0.71 [0.46, 0.97] & 13.31 [12.32, 14.39] & 9.01 [8.27, 9.77] & 76.97 [75.62, 78.27] \\
 & Synthesize-4 & 0.71 [0.48, 0.99] & 13.13 [12.12, 14.16] & 8.91 [8.17, 9.74] & 77.24 [75.92, 78.49] \\
 & Search + Synth-4 & 0.61 [0.38, 0.86] & 11.57 [10.62, 12.58] & 8.08 [7.39, 8.76] & 79.74 [78.58, 80.88] \\
 & Dynamic Guideline & 0.51 [0.31, 0.72] & 9.56 [8.77, 10.38] & 7.84 [7.21, 8.53] & 82.09 [80.99, 83.11] \\
 & Dynamic Excerpt & 0.34 [0.19, 0.51] & 7.59 [6.86, 8.32] & 5.85 [5.26, 6.48] & 86.22 [85.25, 87.15] \\
 & Oracle Guideline & 0.42 [0.23, 0.64] & 7.04 [6.29, 7.79] & 5.53 [4.96, 6.11] & 87.01 [86.01, 87.94] \\
 & Oracle Excerpt & 0.29 [0.13, 0.47] & 3.81 [3.26, 4.34] & 2.96 [2.56, 3.36] & 92.95 [92.26, 93.65] \\
\midrule
\multirow{8}{*}{\textbf{Gemini 3.5 Flash}} & Baseline & 0.81 [0.55, 1.09] & 13.12 [12.07, 14.19] & 9.20 [8.41, 9.95] & 76.87 [75.44, 78.11] \\
 & With Search & 0.82 [0.56, 1.11] & 12.76 [11.80, 13.80] & 8.24 [7.52, 9.01] & 78.17 [76.94, 79.34] \\
 & Synthesize-4 & 0.85 [0.58, 1.13] & 11.48 [10.49, 12.43] & 7.89 [7.16, 8.60] & 79.78 [78.63, 81.00] \\
 & Search + Synth-4 & 0.67 [0.44, 0.93] & 11.44 [10.50, 12.39] & 7.36 [6.68, 8.07] & 80.53 [79.34, 81.65] \\
 & Dynamic Guideline & 0.57 [0.36, 0.79] & 9.25 [8.44, 10.09] & 6.34 [5.74, 6.98] & 83.84 [82.78, 84.87] \\
 & Dynamic Excerpt & 0.45 [0.27, 0.65] & 7.47 [6.73, 8.24] & 5.18 [4.62, 5.76] & 86.89 [85.93, 87.84] \\
 & Oracle Guideline & 0.49 [0.29, 0.71] & 6.20 [5.52, 6.92] & 4.82 [4.27, 5.36] & 88.49 [87.52, 89.41] \\
 & Oracle Excerpt & 0.33 [0.16, 0.53] & 3.26 [2.74, 3.76] & 2.31 [1.95, 2.69] & 94.11 [93.44, 94.79] \\
\midrule
\multirow{8}{*}{\textbf{Gemini 3.6 Flash}} & Baseline & 0.71 [0.46, 0.99] & 11.03 [10.09, 11.96] & 9.01 [8.27, 9.81] & 79.26 [77.99, 80.46] \\
 & With Search & 0.78 [0.53, 1.05] & 11.43 [10.54, 12.36] & 7.77 [7.11, 8.45] & 80.03 [78.87, 81.22] \\
 & Synthesize-4 & 0.76 [0.49, 1.04] & 10.03 [9.11, 10.96] & 7.81 [7.08, 8.54] & 81.41 [80.21, 82.61] \\
 & Search + Synth-4 & 0.69 [0.46, 0.95] & 10.44 [9.54, 11.32] & 7.26 [6.60, 7.96] & 81.61 [80.47, 82.76] \\
 & Dynamic Guideline & 0.58 [0.36, 0.82] & 8.32 [7.56, 9.11] & 6.47 [5.83, 7.16] & 84.62 [83.54, 85.65] \\
 & Dynamic Excerpt & 0.46 [0.28, 0.68] & 6.36 [5.62, 7.09] & 4.61 [4.07, 5.19] & 88.58 [87.66, 89.50] \\
 & Oracle Guideline & 0.41 [0.23, 0.61] & 5.61 [4.96, 6.28] & 4.58 [4.08, 5.16] & 89.40 [88.47, 90.27] \\
 & Oracle Excerpt & 0.32 [0.16, 0.52] & 2.89 [2.42, 3.34] & 2.21 [1.88, 2.59] & 94.58 [93.95, 95.17] \\
\midrule
\multirow{8}{*}{\textbf{Gemini 3.7 Flash}} & Baseline & 0.62 [0.40, 0.88] & 9.78 [8.86, 10.71] & 7.76 [7.05, 8.52] & 81.85 [80.64, 82.97] \\
 & With Search & 0.67 [0.45, 0.92] & 9.96 [9.11, 10.88] & 6.51 [5.89, 7.12] & 82.86 [81.71, 84.03] \\
 & Synthesize-4 & 0.54 [0.33, 0.78] & 9.27 [8.37, 10.18] & 6.63 [5.97, 7.31] & 83.56 [82.36, 84.69] \\
 & Search + Synth-4 & 0.70 [0.46, 0.97] & 9.26 [8.34, 10.16] & 6.38 [5.76, 7.01] & 83.66 [82.53, 84.81] \\
 & Dynamic Guideline & 0.51 [0.30, 0.73] & 6.61 [5.89, 7.37] & 4.75 [4.22, 5.28] & 88.14 [87.18, 89.03] \\
 & Dynamic Excerpt & 0.48 [0.28, 0.72] & 5.17 [4.55, 5.84] & 3.89 [3.43, 4.41] & 90.46 [89.59, 91.32] \\
 & Oracle Guideline & 0.44 [0.24, 0.67] & 4.81 [4.19, 5.43] & 3.52 [3.07, 4.03] & 91.24 [90.36, 92.06] \\
 & Oracle Excerpt & 0.27 [0.12, 0.44] & 2.64 [2.16, 3.11] & 1.55 [1.24, 1.88] & 95.54 [94.91, 96.14] \\
\midrule
\multirow{8}{*}{\textbf{Gemini 3.8 Flash}} & Baseline & 0.49 [0.29, 0.72] & 9.37 [8.53, 10.28] & 6.99 [6.30, 7.71] & 83.14 [81.99, 84.28] \\
 & With Search & 0.47 [0.28, 0.71] & 9.31 [8.42, 10.23] & 6.79 [6.16, 7.47] & 83.42 [82.27, 84.54] \\
 & Synthesize-4 & 0.43 [0.26, 0.64] & 8.78 [7.94, 9.72] & 6.67 [5.99, 7.33] & 84.12 [83.02, 85.24] \\
 & Search + Synth-4 & 0.40 [0.23, 0.59] & 8.97 [8.11, 9.88] & 5.99 [5.39, 6.57] & 84.64 [83.54, 85.74] \\
 & Dynamic Guideline & 0.37 [0.19, 0.56] & 6.04 [5.36, 6.74] & 4.67 [4.16, 5.21] & 88.91 [87.96, 89.80] \\
 & Dynamic Excerpt & 0.26 [0.13, 0.43] & 4.67 [4.06, 5.32] & 3.76 [3.27, 4.26] & 91.31 [90.47, 92.13] \\
 & Oracle Guideline & 0.34 [0.17, 0.53] & 3.84 [3.29, 4.44] & 3.42 [3.00, 3.91] & 92.39 [91.65, 93.14] \\
 & Oracle Excerpt & 0.18 [0.07, 0.32] & 2.22 [1.79, 2.66] & 1.43 [1.18, 1.71] & 96.17 [95.61, 96.69] \\
\bottomrule
\end{tabular}%
}
\end{table}

\clearpage
\section{Analysis of Residual Non-Adherence Under Oracle Excerpt and Oracle Guideline Conditioning}
\label{appendix:oracle_residual}

This section presents an automated analysis of the residual clinical errors that persist even when the ground-truth reference source is provided directly in context. Across the core $3{,}200$-scenario benchmark on Gemini 3.8 Flash, oracle conditioning reaches $96.2\%$ adherence under \emph{Oracle Excerpt} ($3.83\%$ non-adherent) and $92.4\%$ under \emph{Oracle Guideline} ($7.61\%$ non-adherent; \cref{tab:supp_full_model_matrix}), with residual errors consisting mostly of \emph{Moderate} and \emph{Minor Deviations} and \emph{Major Deviations} occurring in $0.18\%$ and $0.34\%$ of evaluations, respectively. Comparing outcomes across the two arms clarifies why condensed excerpts outperform full monographs while remaining below $100\%$: much of the additional error under \emph{Oracle Guideline} occurs on scenarios where \emph{Oracle Excerpt} is consistently adherent, as placing the full \emph{BMJ Best Practice} monograph in context introduces competing patient subgroups, severity stages, and treatment branches that can dilute attention to case-specific steps. Scenarios failing exclusively under \emph{Oracle Excerpt} are substantially less common and typically arise when a scenario tagged to one section (such as \emph{Investigations} or \emph{Complications}) requires a medication or test detailed only in another section of the full monograph, while the remaining errors appear on scenarios with non-adherent plans in both arms.

To summarize the clinical errors underlying these non-adherent evaluations without tying categories one-to-one to severity tiers, we used Gemini 3.8 Flash to review all non-adherent evaluation traces across both arms (comparing each vignette, provided oracle context, generated plan, and autorater rationale) and group them into four broad error patterns (\cref{tab:appendix_oracle_failures}). Because these groupings were generated automatically without clinician adjudication, they serve as a descriptive summary rather than a definitive clinical taxonomy:

\begin{enumerate}[leftmargin=1.5em, topsep=2pt, itemsep=1pt, parsep=0pt]
    \item \textbf{Omitted First-Line Therapy or Primary Workup.}
    Plans identify the target condition and care setting, but omit a required upfront drug, empiric antimicrobial component, or primary diagnostic procedure. Under \emph{Oracle Guideline}, long-context dilution increases missed branch-specific steps that the model follows from a focused excerpt; under \emph{Oracle Excerpt}, omissions also occur when a diagnostic or complication scenario requires a treatment specified only in an unprovided section.

    \item \textbf{Conditional Deferral or Under-Triage of Immediate Actions.}
    Plans note the relevant test, therapy, or escalation pathway, but hedge an immediate guideline requirement as conditional (e.g., ``if symptoms persist or worsen'' or ``pending culture results'') or manage an acute presentation in routine outpatient care, including rare cases where chronic baseline history or outpatient clinic framing masks acute red-flag symptoms.

    \item \textbf{Misapplied Subgroup Criteria, Cutoffs, or Contraindications.}
    Plans anchor on salient clinical cues over explicit laboratory thresholds or patient-subgroup branches, misprioritize a closely mimicking condition, or recommend an action discouraged or contraindicated for that subgroup (including premature invasive workups when conservative care is directed).

    \item \textbf{Omitted Secondary Screening, Rule-Out, or Supportive Checks.}
    Plans satisfy the core diagnosis, triage setting, and primary diagnostic and first-line treatment pathway, but leave out a pre-treatment safety screen, secondary differential rule-out test, cardiovascular or geriatric risk marker, or supportive co-management detail penalized under comprehensive rubric grading.
\end{enumerate}

\vspace{-0.15cm}
\begin{table}[!htbp]
\centering
\caption{LLM-assisted grouping of residual non-adherent evaluations on Gemini 3.8 Flash under \emph{Oracle Excerpt} ($3.83\%$ non-adherent) and \emph{Oracle Guideline} ($7.61\%$ non-adherent) conditioning across the $3{,}200$-scenario core benchmark. Values denote scenario-weighted percentages of all benchmark evaluations by primary assigned error pattern.}
\label{tab:appendix_oracle_failures}
\fontsize{8.0pt}{9.6pt}\selectfont
\setlength{\tabcolsep}{4.5pt}
\renewcommand{\arraystretch}{1.07}
\begin{tabularx}{\textwidth}{>{\raggedright\arraybackslash}p{2.7cm}>{\centering\arraybackslash}p{1.4cm}>{\centering\arraybackslash}p{1.5cm}>{\raggedright\arraybackslash}X}
\toprule
\textbf{Error Pattern} & \textbf{Oracle Excerpt} & \textbf{Oracle Guideline} & \textbf{Mechanism \& Illustrative Examples} \\
\midrule
\textbf{1. Omitted First-Line Therapy or Primary Workup} & $1.01\%$ & $1.49\%$ & \textbf{Mechanism:} The model identifies the right condition and care setting, but leaves out a required upfront medication or primary diagnostic test due to full-monograph dilution (\emph{Oracle Guideline}) or missing cross-section context (\emph{Oracle Excerpt}).\newline \textbf{Examples:} For a hospitalized older adult developing acute severe hypernatremia with neurological changes and osmotic diuresis, the guideline specifies urgent brain MRI to evaluate for osmotic demyelination or intracranial hemorrhage. Given the focused \emph{Complications} excerpt, the model orders urgent neuroimaging alongside controlled fluid correction, whereas given the full guideline (dominated by fluid-calculation branches) it manages the sodium deficit and hyperglycemia but omits the brain MRI. Conversely, in an \emph{Investigations} scenario for Crohn's disease with a draining perianal fistula, \emph{Oracle Excerpt} orders every required scan and scope but misses first-line antibiotics (metronidazole or ciprofloxacin) listed only in the unprovided \emph{Treatment} section, which \emph{Oracle Guideline} includes. \\
\addlinespace[2pt]
\textbf{2. Conditional Deferral or Under-Triage} & $1.04\%$ & $1.58\%$ & \textbf{Mechanism:} The model mentions the appropriate test, medication, or escalation step, but defers an immediate guideline requirement as a conditional fallback or manages an urgent case as a routine clinic visit.\newline \textbf{Examples:} For a patient with lumbar S1 radiculopathy accompanied by objective plantar-flexion weakness ($4/5$) and a diminished ankle reflex ($1+$) on exam (focal neurological deficits warranting upfront lumbar spine MRI), \emph{Oracle Excerpt} orders the MRI upfront in most runs, whereas \emph{Oracle Guideline} usually defers or declines immediate imaging by misapplying the uncomplicated $<6$-week low-back-pain rule. In rarer acute cases, when an outpatient with chronic methamphetamine use presents with new-onset focal seizures, severe untreated hypertension, and cognitive decline (suspected cerebral vasculitis), the model arranges outpatient neuroimaging and neurology follow-up instead of immediate Emergency Department transfer and inpatient admission. \\
\addlinespace[2pt]
\textbf{3. Misapplied Subgroup Criteria, Cutoffs, or Contraindications} & $0.82\%$ & $1.99\%$ & \textbf{Mechanism:} Prominent cues in the patient story override an explicit laboratory threshold, timing criterion, or subgroup rule stated in the provided guideline.\newline \textbf{Examples:} In a postpartum patient presenting $72$ hours after cesarean delivery with worsening thrombocytopenia, anemia, hemolysis, and petechiae, the guideline notes that progressive cytopenias beyond $72$ hours exclude uncomplicated HELLP syndrome and point to acute myeloid leukemia (AML). Anchoring on the postpartum obstetric framing, the model diagnoses persistent HELLP or thrombotic microangiopathy and omits AML from the differential diagnosis and diagnostic workup. Similarly, for suspected late Lyme arthralgia lasting longer than $4$ weeks---where the guideline restricts second-tier confirmatory immunoblot testing strictly to IgG \emph{Borrelia} antibodies---\emph{Oracle Excerpt} follows the timing rule whereas \emph{Oracle Guideline} orders both IgM and IgG Western blots. \\
\addlinespace[2pt]
\textbf{4. Omitted Secondary Screening, Rule-Out, or Supportive Checks} & $0.96\%$ & $2.55\%$ & \textbf{Mechanism:} The model follows all primary diagnostic, triage, and first-line treatment recommendations, but leaves out a secondary tumor screen, prophylaxis measure, or targeted supportive check listed in the guideline.\newline \textbf{Examples:} In a young woman presenting with acute psychosis, fever, and suspected anti-NMDA receptor autoimmune encephalitis (mimicking neuroleptic malignant syndrome), the model outlines the complete neurodiagnostic, lumbar puncture, antibody, and ICU immunotherapy plan, but receives a \emph{Minor Deviation} for omitting pelvic imaging (ultrasound or CT) to screen for an underlying ovarian teratoma. Similarly, in a $74$-year-old woman with giant cell arteritis on tocilizumab and chronic high-dose prednisone ($25\text{ mg/day}$) presenting with drug-induced liver injury, plans appropriately hold tocilizumab and order hepatic workup, but omit guideline-indicated \emph{Pneumocystis jirovecii} pneumonia prophylaxis. \\
\midrule
\textbf{Total Non-Adherent} & $\mathbf{3.83\%}$ & $\mathbf{7.61\%}$ & Residual non-adherence ($100\% - \text{Adherent}$) in \cref{tab:supp_full_model_matrix}. \\
\bottomrule
\end{tabularx}
\end{table}

\clearpage
\definecolor{SlateDark}{HTML}{1E293B}
\definecolor{SlateMuted}{HTML}{475569}
\definecolor{BoxGrayBg}{HTML}{F8FAFC}
\definecolor{BoxGrayBorder}{HTML}{CBD5E1}
\definecolor{GtBoxBg}{HTML}{F0FDF4}
\definecolor{GtBoxBorder}{HTML}{86EFAC}
\definecolor{TierAdherent}{HTML}{059669}
\definecolor{TierAdherentBg}{HTML}{ECFDF5}
\definecolor{TierMajor}{HTML}{DC2626}
\definecolor{TierMajorBg}{HTML}{FEF2F2}
\definecolor{RegionUsAccent}{HTML}{0369A1}
\definecolor{RegionUsBg}{HTML}{F0F9FF}
\definecolor{RegionUsBorder}{HTML}{7DD3FC}
\definecolor{RegionRowAccent}{HTML}{0F766E}
\definecolor{RegionRowBg}{HTML}{F0FDFA}
\definecolor{RegionRowBorder}{HTML}{5EEAD4}

\providecommand{\PcpBoxFont}{}
\renewcommand{\PcpBoxFont}{\fontsize{7pt}{8.5pt}\selectfont}
\providecommand{\PcpTitleFont}{}
\renewcommand{\PcpTitleFont}{\fontsize{8pt}{9.5pt}\selectfont\bfseries}

\providecommand{\PcpTwoColRow}[4]{}
\renewcommand{\PcpTwoColRow}[4]{%
  \par\noindent%
  \def\PcpColW{0.492\linewidth}%
  \sbox0{\begin{tcolorbox}[#1, width=\PcpColW, nobeforeafter]#2\end{tcolorbox}}%
  \edef\PcpRowH{\the\dimexpr\ht0+\dp0+2pt\relax}%
  \sbox0{\begin{tcolorbox}[#3, width=\PcpColW, nobeforeafter]#4\end{tcolorbox}}%
  \ifdim\dimexpr\ht0+\dp0+2pt\relax>\PcpRowH\relax \edef\PcpRowH{\the\dimexpr\ht0+\dp0+2pt\relax}\fi%
  \begin{tcolorbox}[#1, width=\PcpColW, height=\PcpRowH, nobeforeafter]#2\end{tcolorbox}%
  \hfill%
  \begin{tcolorbox}[#3, width=\PcpColW, height=\PcpRowH, nobeforeafter]#4\end{tcolorbox}%
  \par\vspace{3.5pt}%
}
\providecommand{\PcpFitPageBox}[1]{}
\renewcommand{\PcpFitPageBox}[1]{%
  \begingroup%
  \setlength{\parskip}{0pt}%
  \setlength{\parindent}{0pt}%
  \sbox6{\begin{minipage}[b]{7.4in}\setlength{\parskip}{0pt}\setlength{\parindent}{0pt}\PcpBoxFont #1\end{minipage}}%
  \sbox8{\resizebox{\textwidth}{!}{\usebox6}}%
  \noindent\makebox[\textwidth][c]{%
    \ifdim\ht8>0.80\textheight%
      \resizebox{!}{0.80\textheight}{\usebox6}%
    \else%
      \usebox8%
    \fi%
  }%
  \endgroup%
}

\section{Illustrative Targeted Scenario Examples}
\label{appendix:illustrative_regional_freshness_examples}

To illustrate how targeted evaluation scenarios are constructed from verified clinical guideline differences, we present two examples below: one from the temporal freshness evaluation (630 scenarios across 140 updated US clinical monographs) and one from the regional specificity evaluation (731 paired US vs.\ International scenarios across 82 matched clinical monograph pairs).

In the temporal freshness example (\cref{fig:targeted_freshness_example_box}), the 2025 ACC/AHA High Blood Pressure Guideline replaced the 2013 Pooled Cohort Equations (PCE) with the race-free AHA PREVENT calculator and lowered the 10-year total cardiovascular disease (CVD) risk threshold for initiating antihypertensive pharmacotherapy in Stage 1 hypertension ($130\text{--}139 / 80\text{--}89\text{ mmHg}$) from $\ge 10\%$ to $\ge 7.5\%$. For a 57-year-old man with Stage 1 hypertension and an $8.4\%$ 10-year PREVENT risk ($8.1\%$ legacy PCE risk), superseded guidance ($<10\%$ cutoff) defers medication in favor of lifestyle modification alone, whereas current guidance ($\ge 7.5\%$ cutoff) recommends initiating antihypertensive therapy alongside lifestyle modification.

In the regional specificity example (\cref{fig:targeted_regional_example_box}), US guidelines (ADA/ACC/AHA) recommend universal moderate-intensity statin therapy for adults aged 40--75 years with Type 2 diabetes regardless of calculated 10-year risk, whereas the paired International monograph (incorporating UK NICE guidance) requires calculating 10-year CVD risk using QRISK3 and offers routine statin therapy at a 10-year risk of $\ge 10\%$. For an otherwise low-risk 46-year-old woman with Type 2 diabetes and a 10-year QRISK3 score of $4.5\%$, the US edition recommends starting a moderate-intensity statin, whereas the International edition (following UK NICE guidance) withholds routine statin initiation.

\vspace{4pt}

\begin{figure*}[ht!]
\centering
\PcpFitPageBox{%
\begin{tcolorbox}[
  colback=white,
  colframe=SlateDark,
  boxrule=1.1pt,
  arc=3pt,
  left=5pt, right=5pt, top=4.5pt, bottom=1.5pt,
  fonttitle=\PcpTitleFont,
  coltitle=white,
  colbacktitle=SlateDark,
  toptitle=3.5pt, bottomtitle=3.5pt,
  title={\textbf{Temporal Freshness Example: Stage 1 Hypertension Pharmacotherapy Threshold} \hfill \normalfont\PcpBoxFont\textcolor{white!90!black}{Essential Hypertension}}
]
\PcpBoxFont

\PcpTwoColRow{colback=TierMajorBg, colframe=TierMajor, boxrule=0.65pt, arc=2pt, left=5pt, right=5pt, top=3.5pt, bottom=3.5pt, fontupper=\PcpBoxFont\raggedright, title={\PcpBoxFont\textbf{Outdated Guideline (2017 ACC/AHA: PCE, $\ge$10\% Cutoff)}}, colbacktitle=TierMajor, coltitle=white, toptitle=2pt, bottomtitle=2pt}{%
  \textbf{Recommendation:} Estimate 10-year ASCVD risk using the \textbf{2013 Pooled Cohort Equations (PCE)}. In Stage 1 hypertension ($130\text{--}139 / 80\text{--}89\text{ mmHg}$) without clinical CVD, diabetes, or CKD, initiate antihypertensive medication only if 10-year risk is $\ge 10\%$; manage patients with \textbf{$<$10\% risk with lifestyle modification alone for 3--6 months}.\par\vspace{2.5pt}
  \textbf{Guideline Extract:} \textit{``Patients with stage 1 hypertension and assessed as at low risk of CVD ($<$10\% 10-year atherosclerotic CVD risk) may initially be managed with lifestyle modifications and reassessment in 3--6 months.''}
}{colback=TierAdherentBg, colframe=TierAdherent, boxrule=0.65pt, arc=2pt, left=5pt, right=5pt, top=3.5pt, bottom=3.5pt, fontupper=\PcpBoxFont\raggedright, title={\PcpBoxFont\textbf{Updated Guideline (2025 ACC/AHA: AHA PREVENT, $\ge$7.5\% Cutoff)}}, colbacktitle=TierAdherent, coltitle=white, toptitle=2pt, bottomtitle=2pt}{%
  \textbf{Recommendation:} Estimate 10-year total CVD risk using the race-free \textbf{AHA PREVENT calculator}. Initiate antihypertensive pharmacotherapy alongside lifestyle modification when \textbf{10-year PREVENT total CVD risk is $\ge$7.5\%}; reserve lifestyle modification alone strictly for patients with $<$7.5\% risk.\par\vspace{2.5pt}
  \textbf{Guideline Extract:} \textit{``Adults with stage 1 hypertension who have... increased 10-year predicted cardiovascular risk of $\ge$7.5\% defined by PREVENT should receive antihypertensive drugs alongside lifestyle interventions.''}
}

\PcpTwoColRow{colback=BoxGrayBg, colframe=BoxGrayBorder, boxrule=0.6pt, arc=2pt, left=5pt, right=5pt, top=3.5pt, bottom=3.5pt, fontupper=\PcpBoxFont\raggedright, title={\PcpBoxFont\textbf{Clinical Scenario Summary}}, colbacktitle=SlateMuted, coltitle=white, toptitle=2pt, bottomtitle=2pt}{%
  \textbf{Presentation:} A 57-year-old man presents to review his 2-week home blood pressure log and routine labs after an elevated reading at his annual physical. He feels well and asks whether he needs blood pressure medication today or can try diet and exercise alone for 3--6 months first.\par\vspace{2.5pt}
  \textbf{Findings \& Risk Score:} Former smoker (quit 10 years ago) with no history of CVD, diabetes, or CKD. Office BP is \textbf{134/86 mmHg} and 2-week home BP averages \textbf{133/83 mmHg}, with normal renal function, fasting glucose, and lipids.\newline
  $\bullet$ \textbf{Calculated 10-Year CVD Risk:} \textbf{8.4\% via AHA PREVENT} (\textit{8.1\% via legacy PCE})
}{colback=GtBoxBg, colframe=GtBoxBorder, boxrule=0.6pt, arc=2pt, left=5pt, right=5pt, top=3.5pt, bottom=3.5pt, fontupper=\PcpBoxFont\raggedright, title={\PcpBoxFont\textbf{Reference Ground Truth}}, colbacktitle=TierAdherent, coltitle=white, toptitle=2pt, bottomtitle=2pt}{%
  \textbf{Diagnosis:} Stage 1 essential hypertension with elevated 10-year total CVD risk ($\ge 7.5\%$).\par\vspace{2.5pt}
  \textbf{Required Management (Current Guideline):}\newline
  $\bullet$ \textbf{Stratify risk via AHA PREVENT} (8.4\%) rather than the superseded Pooled Cohort Equations (8.1\%).\newline
  $\bullet$ \textbf{Initiate first-line antihypertensive monotherapy today} (thiazide diuretic, CCB, or ACEi/ARB) alongside lifestyle modifications, targeting BP \textbf{$<$130/80 mmHg}.\par\vspace{2.5pt}
  \textbf{Outdated-Practice Pitfall:} Deferring pharmacotherapy for a 3--6 month trial of lifestyle modification alone based on the superseded $<$10\% PCE cutoff.
}

\end{tcolorbox}
}%
\caption{\textbf{Temporal Freshness Example (Essential Hypertension).} Outdated vs.\ updated guideline recommendations, clinical scenario, and reference ground truth.}
\label{fig:targeted_freshness_example_box}
\end{figure*}

\begin{figure*}[ht!]
\centering
\PcpFitPageBox{%
\begin{tcolorbox}[
  colback=white,
  colframe=SlateDark,
  boxrule=1.1pt,
  arc=3pt,
  left=5pt, right=5pt, top=4.5pt, bottom=1.5pt,
  fonttitle=\PcpTitleFont,
  coltitle=white,
  colbacktitle=SlateDark,
  toptitle=3.5pt, bottomtitle=3.5pt,
  title={\textbf{Regional Specificity Example: Primary Prevention Statin Therapy in Type 2 Diabetes} \hfill \normalfont\PcpBoxFont\textcolor{white!90!black}{Type 2 Diabetes in Adults}}
]
\PcpBoxFont

\PcpTwoColRow{colback=RegionUsBg, colframe=RegionUsBorder, boxrule=0.65pt, arc=2pt, left=5pt, right=5pt, top=3.5pt, bottom=3.5pt, fontupper=\PcpBoxFont\raggedright, title={\PcpBoxFont\textbf{US Edition (ADA / ACC / AHA: Universal Ages 40--75)}}, colbacktitle=RegionUsAccent, coltitle=white, toptitle=2pt, bottomtitle=2pt}{%
  \textbf{Recommendation:} Initiate at least \textbf{moderate-intensity statin therapy universally for all adults aged 40--75 years with diabetes} for primary ASCVD prevention, regardless of baseline LDL-C or calculated 10-year CVD risk score.\par\vspace{2.5pt}
  \textbf{Guideline Extract:} \textit{``For primary prevention of CVD in adults with diabetes without established CVD, ADA guidelines recommend: Moderate-intensity statin therapy in all people ages 40--75 years.''}
}{colback=RegionRowBg, colframe=RegionRowBorder, boxrule=0.65pt, arc=2pt, left=5pt, right=5pt, top=3.5pt, bottom=3.5pt, fontupper=\PcpBoxFont\raggedright, title={\PcpBoxFont\textbf{International Edition (UK NICE: QRISK3 $\ge$10\% Risk Cutoff)}}, colbacktitle=RegionRowAccent, coltitle=white, toptitle=2pt, bottomtitle=2pt}{%
  \textbf{Recommendation:} Do not prescribe statins universally in Type 2 diabetes. Calculate 10-year CVD risk using \textbf{QRISK3} and offer routine atorvastatin $20\text{ mg}$ daily at a \textbf{10-year QRISK3 score $\ge$10\%}.\par\vspace{2.5pt}
  \textbf{Guideline Extract:} \textit{``For patients with type 2 diabetes without established ASCVD (i.e., primary prevention), NICE recommends: Offering high-intensity statin therapy (atorvastatin) to those with a $\ge$10\% 10-year risk (estimated using QRISK3) of developing CVD.''}
}

\begin{tcolorbox}[colback=BoxGrayBg, colframe=BoxGrayBorder, boxrule=0.6pt, arc=2pt, left=5pt, right=5pt, top=3.5pt, bottom=3.5pt, before skip=0pt, after skip=3.5pt, fontupper=\PcpBoxFont\raggedright, title={\PcpBoxFont\textbf{Clinical Scenario Summary}}, colbacktitle=SlateMuted, coltitle=white, toptitle=2pt, bottomtitle=2pt]
\textbf{Presentation \& Findings:} A 46-year-old woman with Type 2 diabetes (diagnosed 6 months ago, well controlled on metformin $500\text{ mg}$ twice daily, HbA1c $7.0\%$) presents for routine follow-up and asks whether having diabetes means she should start a statin today. She is a lifelong non-smoker without CVD, CKD, or hypertension (BP \textbf{122/76 mmHg}, LDL-C \textbf{104 mg/dL} [$2.7\text{ mmol/L}$]), with a calculated \textbf{10-year QRISK3 CVD risk of 4.5\%}.
\end{tcolorbox}

\PcpTwoColRow{colback=GtBoxBg, colframe=RegionUsAccent, boxrule=0.65pt, arc=2pt, left=5pt, right=5pt, top=3.5pt, bottom=3.5pt, fontupper=\PcpBoxFont\raggedright, title={\PcpBoxFont\textbf{US Ground Truth}}, colbacktitle=RegionUsAccent, coltitle=white, toptitle=2pt, bottomtitle=2pt}{%
  $\bullet$ \textbf{Initiate moderate-intensity statin therapy today} (e.g., atorvastatin $10\text{--}20\text{ mg}$ or rosuvastatin $5\text{--}10\text{ mg}$ daily) alongside continued metformin and lifestyle measures.\newline
  $\bullet$ \textbf{Rationale:} Under US (ADA/ACC/AHA) guidelines, diabetes in any adult aged 40--75 years is an automatic indication for primary prevention statin therapy regardless of baseline LDL-C or 10-year risk score.
}{colback=GtBoxBg, colframe=RegionRowAccent, boxrule=0.65pt, arc=2pt, left=5pt, right=5pt, top=3.5pt, bottom=3.5pt, fontupper=\PcpBoxFont\raggedright, title={\PcpBoxFont\textbf{International (UK) Ground Truth}}, colbacktitle=RegionRowAccent, coltitle=white, toptitle=2pt, bottomtitle=2pt}{%
  $\bullet$ \textbf{Do not routinely initiate statin therapy}; continue metformin and lifestyle optimization with annual cardiovascular risk reassessment.\newline
  $\bullet$ \textbf{Rationale:} Under UK (NICE) guidelines, primary prevention statin therapy in Type 2 diabetes requires formal \textbf{QRISK3} risk scoring and is gated on a $\ge 10\%$ 10-year risk threshold (patient's QRISK3 is \textbf{4.5\%}).
}

\end{tcolorbox}
}%
\caption{\textbf{Regional Specificity Example (Type 2 Diabetes in Adults).} US vs.\ paired International (UK NICE) monograph recommendations, clinical scenario, and region-specific ground truths.}
\label{fig:targeted_regional_example_box}
\end{figure*}
\clearpage
\definecolor{SlateDark}{HTML}{1E293B}
\definecolor{SlateMuted}{HTML}{475569}
\definecolor{BoxGrayBg}{HTML}{F8FAFC}
\definecolor{BoxGrayBorder}{HTML}{CBD5E1}
\definecolor{GtBoxBg}{HTML}{F0FDF4}
\definecolor{GtBoxBorder}{HTML}{86EFAC}
\definecolor{TierAdherent}{HTML}{059669}
\definecolor{TierAdherentBg}{HTML}{ECFDF5}
\definecolor{TierMinor}{HTML}{D97706}
\definecolor{TierMinorBg}{HTML}{FFFBEB}
\definecolor{TierModerate}{HTML}{EA580C}
\definecolor{TierModerateBg}{HTML}{FFF7ED}
\definecolor{TierMajor}{HTML}{DC2626}
\definecolor{TierMajorBg}{HTML}{FEF2F2}

\providecommand{\PcpBoxFont}{}
\renewcommand{\PcpBoxFont}{\fontsize{7pt}{8.2pt}\selectfont}
\providecommand{\PcpTitleFont}{}
\renewcommand{\PcpTitleFont}{\fontsize{8pt}{9.5pt}\selectfont\bfseries}

\providecommand{\PcpTwoColRow}[4]{}
\renewcommand{\PcpTwoColRow}[4]{%
  \par\noindent%
  \def\PcpColW{0.492\linewidth}%
  \sbox0{\begin{tcolorbox}[#1, width=\PcpColW, nobeforeafter]#2\end{tcolorbox}}%
  \edef\PcpRowH{\the\dimexpr\ht0+\dp0+2pt\relax}%
  \sbox0{\begin{tcolorbox}[#3, width=\PcpColW, nobeforeafter]#4\end{tcolorbox}}%
  \ifdim\dimexpr\ht0+\dp0+2pt\relax>\PcpRowH\relax \edef\PcpRowH{\the\dimexpr\ht0+\dp0+2pt\relax}\fi%
  \begin{tcolorbox}[#1, width=\PcpColW, height=\PcpRowH, nobeforeafter]#2\end{tcolorbox}%
  \hfill%
  \begin{tcolorbox}[#3, width=\PcpColW, height=\PcpRowH, nobeforeafter]#4\end{tcolorbox}%
  \par\vspace{3.5pt}%
}
\providecommand{\PcpThreeColRow}[6]{}
\renewcommand{\PcpThreeColRow}[6]{%
  \par\noindent%
  \def\PcpColW{0.325\linewidth}%
  \sbox0{\begin{tcolorbox}[#1, width=\PcpColW, nobeforeafter]#2\end{tcolorbox}}%
  \edef\PcpRowH{\the\dimexpr\ht0+\dp0+2pt\relax}%
  \sbox0{\begin{tcolorbox}[#3, width=\PcpColW, nobeforeafter]#4\end{tcolorbox}}%
  \ifdim\dimexpr\ht0+\dp0+2pt\relax>\PcpRowH\relax \edef\PcpRowH{\the\dimexpr\ht0+\dp0+2pt\relax}\fi%
  \sbox0{\begin{tcolorbox}[#5, width=\PcpColW, nobeforeafter]#6\end{tcolorbox}}%
  \ifdim\dimexpr\ht0+\dp0+2pt\relax>\PcpRowH\relax \edef\PcpRowH{\the\dimexpr\ht0+\dp0+2pt\relax}\fi%
  \begin{tcolorbox}[#1, width=\PcpColW, height=\PcpRowH, nobeforeafter]#2\end{tcolorbox}%
  \hfill%
  \begin{tcolorbox}[#3, width=\PcpColW, height=\PcpRowH, nobeforeafter]#4\end{tcolorbox}%
  \hfill%
  \begin{tcolorbox}[#5, width=\PcpColW, height=\PcpRowH, nobeforeafter]#6\end{tcolorbox}%
  \par\vspace{3.5pt}%
}
\providecommand{\PcpFitPageBox}[1]{}
\renewcommand{\PcpFitPageBox}[1]{%
  \begingroup%
  \setlength{\parskip}{0pt}%
  \setlength{\parindent}{0pt}%
  \sbox6{\begin{minipage}[b]{7.4in}\setlength{\parskip}{0pt}\setlength{\parindent}{0pt}\PcpBoxFont #1\end{minipage}}%
  \sbox8{\resizebox{\textwidth}{!}{\usebox6}}%
  \noindent\makebox[\textwidth][c]{%
    \ifdim\ht8>0.80\textheight%
      \resizebox{!}{0.80\textheight}{\usebox6}%
    \else%
      \usebox8%
    \fi%
  }%
  \endgroup%
}

\section{Illustrative Primary Care Physician AI-Assistance Examples}
\label{appendix:representative_pcp_examples}

To illustrate how the clinician-calibrated autorater scores primary care physician (PCP) management plans and how AI assistance shapes plan revisions across distinct interaction patterns, we present three case studies below (\cref{fig:pcp_case_box_1_adherent_enriched,fig:pcp_case_box_2_adherent_despite_flawed_ai,fig:pcp_case_box_3_moderate_to_adherent}). We used Gemini 4 Argon in the Antigravity harness to scan the full set of traces and to summarize examples representing diverse revision trajectories:

\begin{enumerate}[leftmargin=1.5em, itemsep=2pt, topsep=3pt]
  \item \textbf{Case 1 (\cref{fig:pcp_case_box_1_adherent_enriched}; Adherent plan enriched by AI, \textit{Adherent} $\rightarrow$ \textit{Adherent}):} In an acute knee osteoarthritis flare, the PCP keeps a sound initial plan and adds practical exam, dosing, and monitoring details from an \textit{Adherent} AI recommendation.
  \item \textbf{Case 2 (\cref{fig:pcp_case_box_2_adherent_despite_flawed_ai}; Selective adoption despite an AI error, \textit{Minor Deviation} $\rightarrow$ \textit{Adherent}):} In suspected primary hypothyroidism, both the initial PCP plan (omitting explicit Free T$_4$) and AI advice (premature levothyroxine) begin as \textit{Minor Deviation}; the PCP catches and rejects the premature prescription while adopting explicit Free T$_4$ and cognitive screening to achieve an \textit{Adherent} revised plan.
  \item \textbf{Case 3 (\cref{fig:pcp_case_box_3_moderate_to_adherent}; Incomplete plan corrected by AI, \textit{Moderate Deviation} $\rightarrow$ \textit{Adherent}):} In suspected pulmonary tuberculosis, an \textit{Adherent} AI recommendation prompts the PCP to elevate tuberculosis to the leading diagnosis and add first-line radiography, sputum testing, and isolation precautions.
\end{enumerate}

\clearpage

\begin{figure*}[p]
\centering
\PcpFitPageBox{%
\begin{tcolorbox}[
  colback=white,
  colframe=SlateDark,
  boxrule=1.1pt,
  arc=3pt,
  left=5pt, right=5pt, top=4pt, bottom=4pt,
  fonttitle=\PcpTitleFont,
  coltitle=white,
  colbacktitle=SlateDark,
  toptitle=3.5pt, bottomtitle=3.5pt,
  title={\textbf{Case 1: Acute Mechanical Flare of Knee Osteoarthritis on Lisinopril} \hfill \normalfont\PcpBoxFont\textcolor{white!92!black}{[Adherent $\rightarrow$ Adherent]}}
]
\PcpBoxFont

\begin{tcolorbox}[colback=BoxGrayBg, colframe=BoxGrayBorder, boxrule=0.5pt, arc=2pt, left=5pt, right=5pt, top=2.5pt, bottom=2.5pt, before skip=0pt, after skip=3.5pt, fontupper=\PcpBoxFont]
\textbf{\textcolor{SlateDark}{Good PCP Plan Enriched by AI Assistance}} \hfill
\textbf{Before AI:} \textcolor{TierAdherent}{\textbf{Adherent}} \,$\rightarrow$\,
\textbf{AI Advice:} \textcolor{TierAdherent}{\textbf{Adherent}} \,$\rightarrow$\,
\textbf{After AI:} \textcolor{TierAdherent}{\textbf{Adherent}}
\end{tcolorbox}

\PcpTwoColRow{colback=BoxGrayBg, colframe=BoxGrayBorder, boxrule=0.6pt, arc=2pt, left=4.5pt, right=4.5pt, top=3.5pt, bottom=3.5pt, fontupper=\PcpBoxFont\raggedright, title={\PcpBoxFont\textbf{1. Clinical Scenario Vignette}}, colbacktitle=SlateMuted, coltitle=white, toptitle=2pt, bottomtitle=2pt}{%
  \textbf{Presenting Complaint:} 68-year-old man (retired accountant) presenting to an outpatient primary care clinic with a severe 2-day flare of right knee pain.\par\vspace{2pt}
  \textbf{History of Present Illness:} ``I've had a `bad knee' for about five years now, usually just a dull ache when I'm out in the garden. But two days ago, after spending the afternoon weeding, the pain shot up to an 8 out of 10. It's much worse than my usual grumble, and I'm struggling to walk to the mailbox. It feels stiff for about 20 minutes when I first wake up, then loosens up a bit. I haven't fallen or hit it recently, it doesn't feel hot, and I haven't had fevers or chills. Is the exercise I'm doing damaging the joint further? Should I use a cane or brace?''\par\vspace{2pt}
  \textbf{Objective Findings \& Vitals:} Temp $36.8^\circ$C, BMI 31 kg/m$^2$. \textbf{Right Knee:} Small, non-inflammatory effusion; \textbf{no palpable warmth or erythema}; crepitus on passive range of motion; pain with flexion beyond $90^\circ$; no joint-line laxity; morning stiffness $\approx 20$ minutes.\par\vspace{2pt}
  \textbf{PMH:} Right knee osteoarthritis (diagnosed clinically 4 years ago); essential hypertension. \textbf{Meds:} Lisinopril 10 mg daily; NKDA. \textbf{SH:} Non-smoker; enjoys gardening and walking; lives in a two-story house.
}{colback=GtBoxBg, colframe=GtBoxBorder, boxrule=0.6pt, arc=2pt, left=4.5pt, right=4.5pt, top=3.5pt, bottom=3.5pt, fontupper=\PcpBoxFont\raggedright, title={\PcpBoxFont\textbf{2. Reference Clinical Guideline Ground Truth}}, colbacktitle=TierAdherent, coltitle=white, toptitle=2pt, bottomtitle=2pt}{%
  \textbf{Probable Diagnosis:} Acute exacerbation of typical Knee Osteoarthritis.\par\vspace{2pt}
  \textbf{Plausible Alternatives:} Crystal arthropathy (gout or calcium pyrophosphate deposition [CPPD/pseudogout]); Degenerative meniscal tear; Septic arthritis.\par\vspace{2pt}
  \textbf{Key Investigations:} Clinical diagnosis of typical knee OA (age $>$45, activity-related joint pain, morning stiffness $\le$30 min, absence of warmth or systemic features); focused gait, range-of-motion, and joint-line exam; bilateral weight-bearing knee radiographs if structural progression or alternative pathology is suspected; reserve joint aspiration strictly for warm/markedly swollen effusions or suspected crystal/septic arthritis.\par\vspace{2pt}
  \textbf{Guideline Management \& Red Flags:} Core nonpharmacologic therapy (reassurance that low-impact exercise does not harm cartilage, PT for quadriceps/hip strengthening, weight reduction, cane in contralateral hand, knee brace); \textbf{topical NSAIDs (diclofenac 1\% gel 4 g QID)} as preferred first-line pharmacotherapy (avoiding or minimizing oral NSAIDs given age, hypertension, and ACE-inhibitor co-therapy); scheduled acetaminophen; consider intra-articular corticosteroid injection for acute flare. \textbf{Red Flags:} Joint warmth, erythema, fever, inability to bear weight.\par\vspace{2pt}
  \textbf{Guideline Justification:} In an older adult on lisinopril with hypertension presenting with an acute mechanical knee OA flare without infectious signs, guidelines prioritize topical diclofenac and core biomechanical offloading/PT over systemic NSAIDs (which carry high nephrotoxic and hypertensive risk alongside ACE inhibitors).
}

\PcpThreeColRow{colback=TierAdherentBg, colframe=TierAdherent, boxrule=0.7pt, arc=2pt, left=4pt, right=4pt, top=3.5pt, bottom=3.5pt, lefttitle=3pt, righttitle=3pt, fontupper=\PcpBoxFont\raggedright, title={\PcpBoxFont\textbf{3A. Before AI: Adherent}}, colbacktitle=TierAdherent, coltitle=white, toptitle=2pt, bottomtitle=2pt}{%
  \textbf{Differential Diagnosis:}\newline Acute mechanical flare of established right-knee osteoarthritis; meniscal tear; gout or CPPD; subchondral insufficiency fracture; pes anserine bursitis; ligament strain; septic arthritis (unlikely).\par\vspace{2.5pt}
  \textbf{Investigations:}\newline Clarify mechanical locking/catching, trauma, and systemic symptoms; examine gait, weight-bearing, active/passive ROM, joint lines, patella, bursae, hip, and neurovascular status; \textbf{weight-bearing knee radiographs} (bilateral AP, lateral, Rosenberg, sunrise); labs (CBC, ESR/CRP, uric acid, creatinine/eGFR) or arthrocentesis only if inflammatory/crystal/infectious signs arise; MRI only for persistent locking or occult fracture.\par\vspace{2.5pt}
  \textbf{Management Plan:}\newline RICE for 48--72 hours, cane/crutch offloading, then graded activity and \textbf{physical therapy} (quadriceps/hip strengthening); \textbf{topical NSAID} and trial of acetaminophen; \textbf{avoid or minimize oral NSAIDs because lisinopril increases renal injury and BP risks} (check renal function if used); consider intra-articular corticosteroid injection after excluding infection; orthopedic referral if refractory.
}{colback=TierAdherentBg, colframe=TierAdherent, boxrule=0.7pt, arc=2pt, left=4pt, right=4pt, top=3.5pt, bottom=3.5pt, lefttitle=3pt, righttitle=3pt, fontupper=\PcpBoxFont\raggedright, title={\PcpBoxFont\textbf{3B. AI Advice: Adherent}}, colbacktitle=TierAdherent, coltitle=white, toptitle=2pt, bottomtitle=2pt}{%
  \textbf{Diagnostic Assessment:}\newline \textbf{Probable:} Acute flare-up of knee osteoarthritis. \textbf{Alternatives:} Crystal-induced arthropathy (CPPD or gout), Degenerative meniscal tear, Pes anserine bursitis, Septic arthritis.\par\vspace{2.5pt}
  \textbf{Recommended Investigations:}\newline Assess gait/mobility, neurovascular exam, \textbf{joint-line tenderness / meniscal maneuvers}, and \textbf{blood pressure and BMI measurement} (critical given hypertension, BMI 31, and potential NSAID exposure); weight-bearing bilateral knee X-rays; \textbf{Basic Metabolic Panel (baseline serum creatinine/eGFR)} given lisinopril therapy; arthrocentesis only if effusion worsens or inflammatory signs develop.\par\vspace{2.5pt}
  \textbf{Recommended Management:}\newline First-line \textbf{topical diclofenac 1\% gel} (safer than oral NSAIDs in hypertensive patients on lisinopril); scheduled \textbf{acetaminophen (up to 3--4 g/day)}; reassure that low-impact exercise protects cartilage; \textbf{5--10\% weight loss target}; PT referral for quadriceps strengthening; contralateral cane; advise against ineffective glucosamine/chondroitin; intra-articular steroid injection if flare persists.
}{colback=TierAdherentBg, colframe=TierAdherent, boxrule=0.7pt, arc=2pt, left=4pt, right=4pt, top=3.5pt, bottom=3.5pt, lefttitle=3pt, righttitle=3pt, fontupper=\PcpBoxFont\raggedright, title={\PcpBoxFont\textbf{3C. After AI: Adherent}}, colbacktitle=TierAdherent, coltitle=white, toptitle=2pt, bottomtitle=2pt}{%
  \textbf{Revised Differential Diagnosis:}\newline Acute mechanical flare of established right-knee osteoarthritis; meniscal tear; gout or CPPD; subchondral insufficiency fracture; pes anserine bursitis; ligament strain; septic arthritis \textit{(unchanged)}.\par\vspace{2.5pt}
  \textbf{Revised Investigations:}\newline Retains full history, joint/hip/neurovascular exam, bilateral weight-bearing knee radiographs, and conditional arthrocentesis/MRI criteria. \textbf{Added from AI:} \textbf{Blood pressure and BMI assessment}, \textbf{McMurray test} for meniscal pathology, and \textbf{baseline serum creatinine} before any NSAID exposure.\par\vspace{2.5pt}
  \textbf{Revised Management Plan:}\newline Retains RICE (48--72h), cane/crutch, PT (quadriceps/hip strengthening), avoidance of oral NSAIDs on lisinopril, and intra-articular steroid option. \textbf{Added from AI:} Specific formulation (\textbf{topical diclofenac gel}), explicit \textbf{acetaminophen dosing (up to 2--4 g/day)}, and \textbf{baseline creatinine check} alongside renal monitoring.
}

\begin{tcolorbox}[colback=BoxGrayBg, colframe=SlateDark, boxrule=0.6pt, arc=2pt, left=4.5pt, right=4.5pt, top=3.5pt, bottom=3.5pt, before skip=0pt, after skip=0pt, fontupper=\PcpBoxFont\raggedright, title={\PcpBoxFont\textbf{4. Clinician-Calibrated Autorater Evaluation \& Impact of AI Assistance}}, colbacktitle=SlateDark, coltitle=white, toptitle=2pt, bottomtitle=2pt]
\textbf{\textcolor{TierAdherent}{Initial PCP Rationale:}} \textbf{Adherent:} Formulates a comprehensive, guideline-concordant primary care plan that orders appropriate weight-bearing radiographs, prioritizes physical therapy, offloading, and topical NSAIDs, and proactively identifies the nephrotoxic and hypertensive interaction between oral NSAIDs and lisinopril.\par\vspace{2pt}
\textbf{\textcolor{TierAdherent}{AI Assistance Rationale:}} \textbf{Adherent:} Accurately targets first-line topical diclofenac gel, explicit acetaminophen dosing, contralateral cane use, weight reduction, and baseline renal function testing (BMP) given concurrent ACE-inhibitor therapy.\par\vspace{2pt}
\textbf{\textcolor{TierAdherent}{Revised PCP Rationale:}} \textbf{Adherent (Enriched with Actionable Clinical Specificity):} While already Adherent pre-AI, the PCP explicitly integrated the AI's bedside checks (\textbf{BP, BMI, and McMurray test}), and added a \textbf{baseline serum creatinine} order.
\end{tcolorbox}

\end{tcolorbox}
}%
\caption{\textbf{Case 1 (Good PCP Plan Enriched by AI Assistance): Acute Mechanical Flare of Knee Osteoarthritis on Lisinopril.} Summarized representations of the clinical vignette, reference clinical guideline ground truth, unassisted PCP plan (Before AI: \textit{Adherent}), AI assistance (\textit{Adherent}), revised PCP plan (After AI: \textit{Adherent}), and clinician-calibrated autorater rationales.}
\label{fig:pcp_case_box_1_adherent_enriched}
\end{figure*}

\clearpage

\begin{figure*}[p]
\centering
\PcpFitPageBox{%
\begin{tcolorbox}[
  colback=white,
  colframe=SlateDark,
  boxrule=1.1pt,
  arc=3pt,
  left=5pt, right=5pt, top=4pt, bottom=4pt,
  fonttitle=\PcpTitleFont,
  coltitle=white,
  colbacktitle=SlateDark,
  toptitle=3.5pt, bottomtitle=3.5pt,
  title={\textbf{Case 2: Suspected Primary Hypothyroidism (Hashimoto's Thyroiditis)} \hfill \normalfont\PcpBoxFont\textcolor{white!92!black}{[Minor Deviation $\rightarrow$ Adherent]}}
]
\PcpBoxFont

\begin{tcolorbox}[colback=BoxGrayBg, colframe=BoxGrayBorder, boxrule=0.5pt, arc=2pt, left=5pt, right=5pt, top=2.5pt, bottom=2.5pt, before skip=0pt, after skip=3.5pt, fontupper=\PcpBoxFont]
\textbf{\textcolor{SlateDark}{PCP Adopts Missing Workup \& Rejects Premature AI Treatment}} \hfill
\textbf{Before AI:} \textcolor{TierMinor}{\textbf{Minor Deviation}} \,$\rightarrow$\,
\textbf{AI Advice:} \textcolor{TierMinor}{\textbf{Minor Deviation}} \,$\rightarrow$\,
\textbf{After AI:} \textcolor{TierAdherent}{\textbf{Adherent}}
\end{tcolorbox}

\PcpTwoColRow{colback=BoxGrayBg, colframe=BoxGrayBorder, boxrule=0.6pt, arc=2pt, left=4.5pt, right=4.5pt, top=3.5pt, bottom=3.5pt, fontupper=\PcpBoxFont\raggedright, title={\PcpBoxFont\textbf{1. Clinical Scenario Vignette}}, colbacktitle=SlateMuted, coltitle=white, toptitle=2pt, bottomtitle=2pt}{%
  \textbf{Presenting Complaint:} 45-year-old woman (librarian) presenting to an outpatient primary care clinic with 4 months of progressive fatigue and memory lapses.\par\vspace{2pt}
  \textbf{History of Present Illness:} ``I've been feeling extremely tired for about four months now, like I'm moving through molasses. My memory isn't what it used to be either; last week I forgot where I parked my car twice and missed a staff meeting. I'm much more sensitive to the cold, my skin is dry and flaky, my hair is brittle, I've gained 6 kg in 3 months, my bowels move only every 3--4 days, and my periods have been much heavier. I have no headaches, visual field changes, or galactorrhea. Could this be early-onset dementia?''\par\vspace{2pt}
  \textbf{Objective Findings \& Vitals:} \textbf{HR 56 bpm (regular bradycardia)}, BP 138/88 mmHg, Temp $36.1^\circ$C ($97.0^\circ$F), BMI 28.2 kg/m$^2$. Cool, dry, coarse skin; thinning of outer third of eyebrows. \textbf{Neck:} Thyroid palpably enlarged, firm, and non-tender without nodules. \textbf{Neuro:} Delayed relaxation phase of Achilles deep tendon reflexes (\textbf{Woltman sign}). \textbf{No lab results are available yet at this index visit.}\par\vspace{2pt}
  \textbf{PMH:} None; no prior neck radiation. \textbf{Meds:} None; NKDA. \textbf{FH:} Mother had ``thyroid trouble'' treated with a daily pill. \textbf{SH:} Librarian for 20 years; non-smoker.
}{colback=GtBoxBg, colframe=GtBoxBorder, boxrule=0.6pt, arc=2pt, left=4.5pt, right=4.5pt, top=3.5pt, bottom=3.5pt, fontupper=\PcpBoxFont\raggedright, title={\PcpBoxFont\textbf{2. Reference Clinical Guideline Ground Truth}}, colbacktitle=TierAdherent, coltitle=white, toptitle=2pt, bottomtitle=2pt}{%
  \textbf{Probable Diagnosis:} Primary Hypothyroidism (Hashimoto's Thyroiditis).\par\vspace{2pt}
  \textbf{Plausible Alternatives:} Central hypothyroidism (pituitary/hypothalamic); Chronic fatigue syndrome; Major depressive disorder; Early-onset Alzheimer's disease.\par\vspace{2pt}
  \textbf{Key Investigations:} Order serum \textbf{Thyroid-Stimulating Hormone (TSH)}, serum \textbf{Free Thyroxine (Free T$_4$)} (required alongside TSH to confirm overt hypothyroidism and exclude central hypothyroidism), and serum \textbf{anti-thyroid peroxidase (anti-TPO) antibodies}.\par\vspace{2pt}
  \textbf{Guideline Management \& Red Flags:} \textbf{Follow up in 1 week to review laboratory results and initiate oral levothyroxine ONLY after hypothyroidism is biochemically confirmed}; reassure patient that cognitive slowing (``brain fog'') is a reversible manifestation of hypothyroidism. \textbf{Red Flags:} Confusion or extreme lethargy (myxedema coma), severe chest pain, or worsening dyspnea.\par\vspace{2pt}
  \textbf{Guideline Justification:} Although fatigue, cold intolerance, weight gain, menorrhagia, goiter, bradycardia, and Woltman sign strongly point to primary hypothyroidism, lifelong thyroid hormone replacement must not be started empirically before biochemical confirmation (elevated TSH and low Free T$_4$) and exclusion of central/adrenal causes.
}

\PcpThreeColRow{colback=TierMinorBg, colframe=TierMinor, boxrule=0.7pt, arc=2pt, left=4pt, right=4pt, top=3.5pt, bottom=3.5pt, lefttitle=3pt, righttitle=3pt, fontupper=\PcpBoxFont\raggedright, title={\PcpBoxFont\textbf{3A. Before AI: Minor Deviation}}, colbacktitle=TierMinor, coltitle=white, toptitle=2pt, bottomtitle=2pt}{%
  \textbf{Differential Diagnosis:}\newline 1.~Hypothyroidism; 2.~Symptomatic bradycardia; 3.~Hashimoto's thyroiditis; 4.~Anemia.\par\vspace{2.5pt}
  \textbf{Investigations:}\newline 12-lead ECG (for HR 56 bpm); \textbf{serum TSH and thyroid studies (anti-TPO and anti-Tg antibodies)}, CBC, BMP, iron studies (for menorrhagia), \textbf{morning cortisol}, and thyroid ultrasound. \textit{(Orders TSH, thyroid studies/antibodies, and morning cortisol; does not list Free T$_4$ separately or include cognitive/mood scales.)}\par\vspace{2.5pt}
  \textbf{Management Plan:}\newline \textbf{``Further management based on results. Can initiate levothyroxine as soon as results are available, with consideration to cortisol levels too.''} Endocrinology and Cardiology referrals as indicated.
}{colback=TierMinorBg, colframe=TierMinor, boxrule=0.7pt, arc=2pt, left=4pt, right=4pt, top=3.5pt, bottom=3.5pt, lefttitle=3pt, righttitle=3pt, fontupper=\PcpBoxFont\raggedright, title={\PcpBoxFont\textbf{3B. AI Advice: Minor Deviation}}, colbacktitle=TierMinor, coltitle=white, toptitle=2pt, bottomtitle=2pt}{%
  \textbf{Diagnostic Assessment:}\newline \textbf{Probable:} Overt Primary Hypothyroidism (Hashimoto's Thyroiditis). \textbf{Alternatives:} Iron deficiency anemia, Perimenopause, Major Depressive Disorder, Iodine deficiency goiter, Early-onset Alzheimer's disease.\par\vspace{2.5pt}
  \textbf{Recommended Investigations:}\newline Confirm weight; \textbf{baseline cognitive assessment (MoCA/MMSE)} to evaluate memory complaints; \textbf{depression screen (PHQ-9)}; repeat BP; serum TSH, \textbf{Free T$_4$}, anti-TPO and anti-Tg antibodies, CBC, \textbf{fasting lipid profile}, and thyroid ultrasound.\par\vspace{2.5pt}
  \textbf{Recommended Management:}\newline \textbf{Premature Initiation (Minor Deviation): ``Initiate hormone replacement therapy with oral Levothyroxine (standard starting dose of 1.6 mcg/kg/day)'' immediately at the index visit before laboratory confirmation}; advise empty-stomach dosing separated from iron/calcium by 4 hours; reassure on reversible ``brain fog''; recheck TSH in 6--8 weeks.
}{colback=TierAdherentBg, colframe=TierAdherent, boxrule=0.7pt, arc=2pt, left=4pt, right=4pt, top=3.5pt, bottom=3.5pt, lefttitle=3pt, righttitle=3pt, fontupper=\PcpBoxFont\raggedright, title={\PcpBoxFont\textbf{3C. After AI: Adherent}}, colbacktitle=TierAdherent, coltitle=white, toptitle=2pt, bottomtitle=2pt}{%
  \textbf{Revised Differential Diagnosis:}\newline 1.~Hypothyroidism; 2.~Hashimoto's thyroiditis; 3.~Symptomatic bradycardia; 4.~Anemia; \textbf{perimenopause, major depressive disorder, early-onset Alzheimer's disease} \textit{(added from AI)}.\par\vspace{2.5pt}
  \textbf{Revised Investigations:}\newline Retains 12-lead ECG, TSH, anti-TPO/anti-Tg antibodies, CBC, BMP, iron studies, cortisol, and thyroid US. \textbf{Adopted from AI:} \textbf{Serum Free T$_4$}, \textbf{lipid profile}, and \textbf{formal cognitive and mood evaluations (MoCA/MMSE and PHQ-9)} (\textit{``I did not consider the memory impairment complaints and it does deserve a full assessment like the one proposed by the AI''}).\par\vspace{2.5pt}
  \textbf{Revised Management Plan:}\newline \textbf{Explicitly rejected AI's premature levothyroxine recommendation:} \textit{``I would only start levothyroxine if the levels are low. So at this point I would not make that recommendation.''} Retains plan to await confirmatory lab results and cortisol check before starting levothyroxine, plus myxedema/cardiopulmonary red-flag counseling.
}

\begin{tcolorbox}[colback=BoxGrayBg, colframe=SlateDark, boxrule=0.6pt, arc=2pt, left=4.5pt, right=4.5pt, top=3.5pt, bottom=3.5pt, before skip=0pt, after skip=0pt, fontupper=\PcpBoxFont\raggedright, title={\PcpBoxFont\textbf{4. Clinician-Calibrated Autorater Evaluation \& Impact of AI Assistance}}, colbacktitle=SlateDark, coltitle=white, toptitle=2pt, bottomtitle=2pt]
\textbf{\textcolor{TierMinor}{Initial PCP Rationale:}} \textbf{Minor Deviation (Omitted Essential Free T$_4$):} Identifies primary hypothyroidism/Hashimoto's thyroiditis, orders TSH, thyroid antibodies, and morning cortisol, and safely defers levothyroxine until laboratory confirmation returns, but omits explicit \textbf{serum Free T$_4$} alongside TSH (required to confirm overt disease and distinguish primary from central hypothyroidism).\par\vspace{2pt}
\textbf{\textcolor{TierMinor}{AI Assistance Rationale:}} \textbf{Minor Deviation (Premature Non-Emergent Maintenance Initiation):} Contributes useful workup additions (explicit Free T$_4$, lipid panel, MoCA/MMSE, PHQ-9), but prescribes full-dose \textbf{oral levothyroxine (1.6 mcg/kg/day)} prior to biochemical confirmation of TSH and Free T$_4$.\par\vspace{2pt}
\textbf{\textcolor{TierAdherent}{Revised PCP Rationale:}} \textbf{Adherent (Synergistic Selective Adoption \& Error Rejection):} Although both unassisted plans had a Minor Deviation, the PCP \textbf{selectively adopted} explicit \textbf{Free T$_4$} (resolving their own initial omission), a \textbf{fasting lipid profile}, and \textbf{cognitive/depression screening (MoCA/MMSE, PHQ-9)} while \textbf{actively catching and rejecting} the AI's premature levothyroxine prescription.
\end{tcolorbox}

\end{tcolorbox}
}%
\caption{\textbf{Case 2 (PCP Adopts Missing Workup \& Rejects Premature AI Treatment): Suspected Primary Hypothyroidism (Hashimoto's Thyroiditis).} Summarized representations of the clinical vignette, reference clinical guideline ground truth, unassisted PCP plan (Before AI: \textit{Minor Deviation}), AI assistance (\textit{Minor Deviation}), revised PCP plan (After AI: \textit{Adherent}), and clinician-calibrated autorater rationales.}
\label{fig:pcp_case_box_2_adherent_despite_flawed_ai}
\end{figure*}

\clearpage

\begin{figure*}[p]
\centering
\PcpFitPageBox{%
\begin{tcolorbox}[
  colback=white,
  colframe=SlateDark,
  boxrule=1.1pt,
  arc=3pt,
  left=5pt, right=5pt, top=4pt, bottom=4pt,
  fonttitle=\PcpTitleFont,
  coltitle=white,
  colbacktitle=SlateDark,
  toptitle=3.5pt, bottomtitle=3.5pt,
  title={\textbf{Case 3: Suspected Pulmonary Tuberculosis After Non-Response to Amoxicillin} \hfill \normalfont\PcpBoxFont\textcolor{white!92!black}{[Moderate Deviation $\rightarrow$ Adherent]}}
]
\PcpBoxFont

\begin{tcolorbox}[colback=BoxGrayBg, colframe=BoxGrayBorder, boxrule=0.5pt, arc=2pt, left=5pt, right=5pt, top=2.5pt, bottom=2.5pt, before skip=0pt, after skip=3.5pt, fontupper=\PcpBoxFont]
\textbf{\textcolor{SlateDark}{Incomplete PCP Plan Corrected by AI Assistance}} \hfill
\textbf{Before AI:} \textcolor{TierModerate}{\textbf{Moderate Deviation}} \,$\rightarrow$\,
\textbf{AI Advice:} \textcolor{TierAdherent}{\textbf{Adherent}} \,$\rightarrow$\,
\textbf{After AI:} \textcolor{TierAdherent}{\textbf{Adherent}}
\end{tcolorbox}

\PcpTwoColRow{colback=BoxGrayBg, colframe=BoxGrayBorder, boxrule=0.6pt, arc=2pt, left=4.5pt, right=4.5pt, top=3.5pt, bottom=3.5pt, fontupper=\PcpBoxFont\raggedright, title={\PcpBoxFont\textbf{1. Clinical Scenario Vignette}}, colbacktitle=SlateMuted, coltitle=white, toptitle=2pt, bottomtitle=2pt}{%
  \textbf{Presenting Complaint:} 45-year-old man (accountant) presenting to an outpatient urban primary care clinic with 4 weeks of persistent cough, dyspnea, and unintentional weight loss.\par\vspace{2pt}
  \textbf{History of Present Illness:} ``I've been feeling quite short of breath for about a month now (mMRC Grade 2: walking slower than people my age), and I have this nagging cough that won't go away. The cough produces yellowish phlegm, and twice I've noticed thin streaks of blood in it. I've lost about 4 kg in the last month despite eating normally, and every night for the past two weeks I've woken up with my sheets drenched in sweat and feeling warm in the evenings. Two weeks ago, a walk-in clinic gave me amoxicillin 500 mg TID for 7 days, which didn't help at all. Why didn't the antibiotics clear this up?''\par\vspace{2pt}
  \textbf{Objective Findings \& Vitals:} BP 116/74 mmHg, HR 92 bpm, \textbf{Temp $37.9^\circ$C (oral evening pyrexia)}, RR 22/min, SpO$_2$ 94\% on room air. \textbf{Lungs:} Coarse inspiratory crackles localized to the \textbf{right apex and left upper lobe}; no peripheral edema or digital clubbing.\par\vspace{2pt}
  \textbf{PMH/PSH:} Appendectomy at age 12; last HIV test 1 year ago (negative). \textbf{Meds:} Completed amoxicillin 500 mg TID $\times$ 7 days one week ago; NKDA. \textbf{SH:} Lives alone in city apartment; uses public transit daily; non-smoker.
}{colback=GtBoxBg, colframe=GtBoxBorder, boxrule=0.6pt, arc=2pt, left=4.5pt, right=4.5pt, top=3.5pt, bottom=3.5pt, fontupper=\PcpBoxFont\raggedright, title={\PcpBoxFont\textbf{2. Reference Clinical Guideline Ground Truth}}, colbacktitle=TierAdherent, coltitle=white, toptitle=2pt, bottomtitle=2pt}{%
  \textbf{Probable Diagnosis:} Pulmonary Tuberculosis (TB).\par\vspace{2pt}
  \textbf{Plausible Alternatives:} Non-resolving / atypical community-acquired pneumonia (CAP); Bronchogenic carcinoma (lung malignancy); Sarcoidosis.\par\vspace{2pt}
  \textbf{Key Investigations:} Immediate \textbf{Chest X-ray (PA and lateral)}; \textbf{Sputum for Acid-Fast Bacilli (AFB) smear and culture} (3 consecutive specimens) plus \textbf{Sputum Nucleic Acid Amplification Test (NAAT / Xpert MTB/RIF)}; repeat \textbf{HIV antigen/antibody screening}.\par\vspace{2pt}
  \textbf{Guideline Management \& Red Flags:} Institute immediate \textbf{respiratory / airborne precautions} (masking and avoidance of public transit/crowded spaces pending sputum AFB exclusion); arrange rapid follow-up within 48--72 hours to review Chest X-ray and initial sputum smear results. \textbf{Red Flags:} Massive or worsening hemoptysis, severe respiratory distress, high fever $>39^\circ$C.\par\vspace{2pt}
  \textbf{Guideline Justification:} Subacute cough ($\ge$4 weeks) with hemoptysis, 4-kg weight loss, drenching night sweats, evening fever ($37.9^\circ$C), upper-lobe crackles, and failure to respond to amoxicillin is classic for active pulmonary TB. Guidelines mandate first-line Chest X-ray, sputum AFB smear/culture/NAAT, HIV testing, and immediate respiratory precautions.
}

\PcpThreeColRow{colback=TierModerateBg, colframe=TierModerate, boxrule=0.7pt, arc=2pt, left=4pt, right=4pt, top=3.5pt, bottom=3.5pt, lefttitle=3pt, righttitle=3pt, fontupper=\PcpBoxFont\raggedright, title={\PcpBoxFont\textbf{3A. Before AI: Moderate Deviation}}, colbacktitle=TierModerate, coltitle=white, toptitle=2pt, bottomtitle=2pt}{%
  \textbf{Differential Diagnosis:}\newline Lung cancer, lymphoma, HIV, tuberculosis, granulomatosis with polyangiitis (GPA). \textit{(Lists TB 4th behind malignancy, lymphoma, and HIV.)}\par\vspace{2.5pt}
  \textbf{Investigations:}\newline Chest CT, CBC with differential, peripheral blood smear, CMP, TSH, ANCA panel, ESR, and CRP. \textit{(Skips initial Chest X-ray in favor of upfront Chest CT/ANCA/TSH and completely omits sputum AFB smear/culture/NAAT and TB/HIV testing.)}\par\vspace{2.5pt}
  \textbf{Management Plan:}\newline ``Pending results, follow up in 1 week to discuss.'' \textit{(Omits respiratory isolation/masking precautions in a patient using daily public transit.)}
}{colback=TierAdherentBg, colframe=TierAdherent, boxrule=0.7pt, arc=2pt, left=4pt, right=4pt, top=3.5pt, bottom=3.5pt, lefttitle=3pt, righttitle=3pt, fontupper=\PcpBoxFont\raggedright, title={\PcpBoxFont\textbf{3B. AI Advice: Adherent}}, colbacktitle=TierAdherent, coltitle=white, toptitle=2pt, bottomtitle=2pt}{%
  \textbf{Diagnostic Assessment:}\newline \textbf{Probable:} \textbf{Pulmonary Tuberculosis (TB)}. \textbf{Alternatives:} Bronchogenic carcinoma, Non-responsive or atypical pneumonia, Lung abscess, Fungal pulmonary infection, Sarcoidosis.\par\vspace{2.5pt}
  \textbf{Recommended Investigations:}\newline Place surgical mask immediately (initiate airborne precautions); lymph node exam; baseline weight/BMI; repeat Temp and SpO$_2$; \textbf{urgent Chest X-ray (PA and lateral)}; \textbf{sputum for AFB smear, culture, and NAAT (GeneXpert)} (3 consecutive specimens); \textbf{repeat HIV-1/2 Ag/Ab screen}; CBC, CRP, BMP, and baseline LFTs; reserve Chest CT for inconclusive CXR or suspected mass.\par\vspace{2.5pt}
  \textbf{Recommended Management:}\newline \textbf{Immediate respiratory isolation counseling} (avoid public transportation and crowded spaces until sputum AFB results return; strict cough etiquette); instruct on early-morning deep sputum collection; close follow-up within 48--72 hours; urgent ED return precautions for brisk hemoptysis or resting dyspnea.
}{colback=TierAdherentBg, colframe=TierAdherent, boxrule=0.7pt, arc=2pt, left=4pt, right=4pt, top=3.5pt, bottom=3.5pt, lefttitle=3pt, righttitle=3pt, fontupper=\PcpBoxFont\raggedright, title={\PcpBoxFont\textbf{3C. After AI: Adherent}}, colbacktitle=TierAdherent, coltitle=white, toptitle=2pt, bottomtitle=2pt}{%
  \textbf{Revised Differential Diagnosis:}\newline \textbf{1.~Tuberculosis} \textit{(promoted to primary diagnosis)}; 2.~Pneumonia \textit{(added)}; 3.~Lung cancer \textit{(removed lymphoma and GPA)}.\par\vspace{2.5pt}
  \textbf{Revised Investigations:}\newline \textbf{Chest X-ray}, CBC, CMP, ESR, CRP, \textbf{TB and HIV testing}, and \textbf{sputum acid-fast bacilli (AFB) smear} \textit{(added CXR, sputum AFB smear, and TB/HIV testing; dropped premature upfront Chest CT, ANCA, peripheral smear, and TSH)}.\par\vspace{2.5pt}
  \textbf{Revised Management Plan:}\newline \textbf{``Advise on respiratory precautions and recommend close follow-up.''} \textit{(Added infection-control precautions and shortened follow-up interval.)}
}

\begin{tcolorbox}[colback=BoxGrayBg, colframe=SlateDark, boxrule=0.6pt, arc=2pt, left=4.5pt, right=4.5pt, top=3.5pt, bottom=3.5pt, before skip=0pt, after skip=0pt, fontupper=\PcpBoxFont\raggedright, title={\PcpBoxFont\textbf{4. Clinician-Calibrated Autorater Evaluation \& Impact of AI Assistance}}, colbacktitle=SlateDark, coltitle=white, toptitle=2pt, bottomtitle=2pt]
\textbf{\textcolor{TierModerate}{Initial PCP Rationale:}} \textbf{Moderate Deviation:} Lists TB fourth behind malignancy and vasculitis, orders CT/ANCA/TSH workup while \textbf{omitting mandatory sputum AFB smear/culture/NAAT microbiology, upfront Chest X-ray, and HIV testing}, and fails to institute \textbf{respiratory precautions} or rapid 48--72-hour follow-up in a patient commuting by public transit.\par\vspace{2pt}
\textbf{\textcolor{TierAdherent}{AI Assistance Rationale:}} \textbf{Adherent:} Prioritizes active pulmonary TB, orders urgent PA/lateral Chest X-ray, 3-specimen sputum AFB smear/culture/NAAT, and repeat HIV testing, and immediately initiates masking and home respiratory isolation.\par\vspace{2pt}
\textbf{\textcolor{TierAdherent}{Revised PCP Rationale:}} \textbf{Adherent (Upgraded from Moderate Deviation):} After reviewing the AI plan, the PCP promoted \textbf{Tuberculosis} to the primary diagnosis, replaced the CT/ANCA workup with \textbf{Chest X-ray, sputum AFB smear, and TB/HIV testing}, and added \textbf{respiratory precautions and close follow-up}.
\end{tcolorbox}

\end{tcolorbox}
}%
\caption{\textbf{Case 3 (Incomplete PCP Plan Corrected by AI Assistance): Suspected Pulmonary Tuberculosis After Non-Response to Amoxicillin.} Summarized representations of the clinical vignette, reference clinical guideline ground truth, unassisted PCP plan (Before AI: \textit{Moderate Deviation}), AI assistance (\textit{Adherent}), revised PCP plan (After AI: \textit{Adherent}), and clinician-calibrated autorater rationales.}
\label{fig:pcp_case_box_3_moderate_to_adherent}
\end{figure*}

\clearpage

\end{refsegment}

\end{document}